\documentclass{article}

\usepackage{microtype}
\usepackage{graphicx}
\usepackage{subfigure}
\usepackage{booktabs} 

\usepackage[utf8]{inputenc} 
\usepackage[T1]{fontenc}    
\usepackage{hyperref}       
\usepackage{url}            
\usepackage{booktabs}       
\usepackage{amsfonts}       
\usepackage{nicefrac}       
\usepackage{microtype}      
\usepackage{xcolor}         
\usepackage{xspace} 
\usepackage{subfigure}
\usepackage{graphicx}
\usepackage{amsmath}
\usepackage{bbm}
\usepackage{dsfont}
\usepackage{xspace} 
\usepackage{amsmath}
\usepackage{mathtools}
\usepackage{bbm}
\usepackage{dsfont}
\usepackage{xspace} 
\usepackage{enumitem}
\usepackage{amsthm}

\DeclareMathOperator*{\argmin}{arg\,min}
\usepackage[ruled,linesnumbered, noend]{algorithm2e}

\def \bX {\mathbf{X}}
\def \bY {\mathbf{Y}}

\def \cX {\mathcal{X}}
\def \cY {\mathcal{Y}}
\def \cC {\mathcal{C}}

\def \cH {\mathcal{H}}

\newtheorem{exmp}{Example}

\newcommand{\compilehidecomments}{true}
\ifthenelse{ \equal{\compilehidecomments}{true} }{%
\newcommand{\todiscuss}[1]{\noindent{\colorbox{yellow}{TODiscuss: #1}}}
\newcommand{\outline}[1]{\noindent{\colorbox{lightgray}{OUTLINE:#1}}}
\newcommand{\todo}[1]{\noindent{\colorbox{pink}{TODO:#1}}}
}{
\newcommand{\todiscuss}[1]{}
\newcommand{\outline}[1]{}
\newcommand{\todo}[1]{}
}

\newcommand{\Fair}{{\texttt{Fair}}}
\newcommand{\Loss}{{\texttt{Loss}}}

\newcommand{\train}{{\text{train}}}
\newcommand{\val}{{\text{val}}}
\newcommand{\suggest}{{\texttt{suggest}}}
\newcommand{\update}{{\texttt{update}}}

\newcommand{\FairnessManager}{{\texttt{FairSearcher}}\xspace}
\newcommand{\HPSearcher}{{\texttt{HPSearcher}}\xspace}

\newcommand{\UFMitigation}{{\texttt{Mitigate}}\xspace}

\newcommand{\ECF}{{\text{ECF}}\xspace}
\newcommand{\fairFLAML}{\text{FairFLAML}\xspace}

\newcommand{\fairFLAMLg}{{\text{FairFLAML-g}}\xspace}
\newcommand{\fairFLAMLp}{{\text{FairFLAML-p}}\xspace}
\newcommand{\MasHP}{\texttt{MasHP}\xspace}
\newcommand{\MAlways}{\texttt{MAlways}\xspace}
\newcommand{\PostFLAML}{\texttt{MPost}\xspace}

\usepackage{hyperref}

\usepackage[accepted]{mlsys2023}

\mlsystitlerunning{FairAutoML: Embracing Unfairness Mitigation in AutoML}

\begin{document}

\twocolumn[
\mlsystitle{FairAutoML: Embracing Unfairness Mitigation in AutoML}

\begin{mlsysauthorlist}
\mlsysauthor{Qingyun Wu}{to}
\mlsysauthor{Chi Wang}{goo}
\end{mlsysauthorlist}

\mlsysaffiliation{to}{Pennsylvania State University (part of the work is done when the author is at Microsoft Research).}
\mlsysaffiliation{goo}{Microsoft Research}
\mlsyscorrespondingauthor{Qingyun Wu}{qingyun.wu@psu.edu}

\mlsyskeywords{Machine Learning, MLSys}

\vskip 0.3in

\begin{abstract}
In this work, we propose an Automated Machine Learning (AutoML) system to search for models not only with good prediction accuracy but also fair. We first investigate the necessity and impact of unfairness mitigation in the AutoML context. We establish the FairAutoML framework. The framework provides a novel design based on pragmatic abstractions, which makes it convenient to incorporate existing fairness definitions, unfairness mitigation techniques, and hyperparameter search methods into the model search and evaluation process. Following this framework, we develop a fair AutoML system based on an existing AutoML system. The augmented system includes a resource allocation strategy to dynamically decide when and on which models to conduct unfairness mitigation according to the prediction accuracy, fairness, and resource consumption on the fly. Extensive empirical evaluation shows that our system can achieve a good ‘fair accuracy’ and high resource efficiency. 
\end{abstract}
]



\printAffiliationsAndNotice{}  

\section{Introduction}
\label{intro}
Effective automated machine learning (AutoML) systems have been developed to automatically tune hyperparameter configurations and find ML models with a good predictive performance from given training data. 
These systems are increasingly used to manage ML pipelines in various practical scenarios~\cite{AutoML_Market_report}. In many real-world applications where the decisions to be made have a direct impact on the well-being of human beings; however, it is not sufficient to only have good predictive performance. We also expect ML-based decisions to be ethical that do not put certain unprivileged groups or individuals at a systemic disadvantage. Unfortunately, there has been increasing evidence of various unfairness issues associated with machine-made decisions or predictions in many of such applications~\cite{o2016weapons,barocas2016big,2016machinebias}. Considering the increasing adoption of AutoML systems, we find it crucial to study how to improve fairness in AutoML systems.

Considerable research effort has been devoted to machine learning fairness in recent years, including research on fairness definitions~\cite{dwork2012fairness,hardt2016equality,chouldechova2017fair,lahoti2019ifair} and unfairness mitigation~\cite{kamiran2012data, kamishima2011fairness, friedler2014certifying,hardt2016equality,NIPS2017_9a49a25d,woodworth2017learning,zafar2017fairness, agarwal2018reductions, zhang2021omnifair}. Many of existing unfairness mitigation methods are quite effective in mitigating unfairness-related harms of machine learning models. However, they are rarely explored in AutoML systems.  Considering the enormous practical impact of AutoML, we are motivated to explore \emph{whether it is helpful to incorporate unfairness mitigation techniques to AutoML systems to improve the systems' overall fairness, and if so, what is a good way to incorporate them}. 

To answer the first question, we conduct both literature survey and case studies.  Our study suggests that it is beneficial (and sometimes essential) to incorporate unfairness mitigation into AutoML. Toward answering the second question, we propose a novel and pragmatic solution to enable the incorporation of unfairness mitigation into AuotML. 
Specifically, we first investigate the necessity of unfairness mitigation in AutoML and its potential impact on the AutoML system regarding fairness, predictive performance, and computation cost. We then provide a fair AutoML formulation taking unfairness mitigation into consideration. Accompanying the formulation, we develop necessary abstractions to generally characterize the fairness assessment and unfairness mitigation procedures in the AutoML context. We also propose a self-adaptive decision-making strategy for achieving fair AutoML effectively and efficiently. The strategy helps determine when to conduct regular model training, and whether and when to conduct an additional unfairness mitigation procedure in each trial of the AutoML process.  

We summarize key contributions of this work as follows:
\begin{enumerate} [leftmargin=*]
    \vspace{-5mm}
    \setlength\itemsep{-0.2em}
    \item 
    We provide a rigorous formulation for the fair AutoML problem taking unfairness mitigation into consideration, and propose a general framework for solving the fair AutoML problem. (ref. Section~\ref{sec:formulation_ana}). The abstractions in the framework allow convenient and flexible incorporation of most existing fairness assessment methods and unfairness mitigation techniques into AutoML. 
    \item 

   We propose an adaptive strategy for deciding when and whether to perform unfirenss mitigation are novel and of practical importance. (ref. Section~\ref{sec:framework}). With this strategy, computation resources can be allocated efficiently between hyperparameter tuning and unfairness mitigation. 
    \item We develop a working fair AutoML system 
    based on an existing AutoML system. The proposed fair AutoML system is \emph{flexible}, \emph{robust} and \emph{efficient} in terms of building fair models with good predictive performance. Extensive empirical evaluation involving four machine learning fairness benchmark datasets, two different fairness definitions with two different fairness thresholds, three different unfairness mitigation methods, two tuning tasks (tuning tasks on a single learner and multiple learners), and two tuning methods, verify the robust good performance of our system over all the evaluated scenarios. (ref. Section~\ref{sec:exp}). 
\end{enumerate}

\section{Related work} \label{sec:related_work}

\textbf{Fairness definitions and unfairness mitigation.} There are mainly two types of fairness definitions for machine learning: group fairness~\cite{dwork2012fairness,simoiu2017problem,hardt2016equality} and individual fairness~\cite{dwork2012fairness,lahoti2019ifair}. Under the group fairness framing, fairness, generally speaking, requires some aspects of the machine learning system's behavior to be comparable or even equalized across different groups. 
Commonly used group fairness metrics include Demographic Parity (DP)~\cite{dwork2012fairness} and Equalized Odds (EO) \cite{hardt2016equality}.
Individual fairness ensures that individuals who are `similar' with respect to the learning task receive similar outcomes. Unfairness mitigation methods seek to reduce fairness-related harms of a machine learning model regarding a particular fairness metric. From a procedural viewpoint, unfairness mitigation methods can be roughly categorized into pre-processing~\cite{kamiran2012data,NIPS2017_9a49a25d}, in-processing~\cite{kamishima2011fairness,woodworth2017learning,zafar2017fairness, agarwal2018reductions,zhang2021omnifair}, and post-processing approaches~\cite{friedler2014certifying,hardt2016equality}, depending on whether the methods should be applied before, during or after the model training process. 
Pre-processing methods typically transform the training data to try to remove undesired biases. In-processing methods try to reduce fairness-related harms by imposing fairness constraints into the learning mechanism. Post-processing methods directly transform the model outputs to reduce biases.
Software toolkits have been developed for machine learning fairness. Representative ones include the open-source Python libraries AIF360~\cite{aif360-oct-2018} and FairLearn~\cite{bird2020fairlearn}. Both libraries provide implementations for state-of-the-art fairness measurement metrics, unfairness mitigation methods, and collections of datasets commonly used for machine learning fairness research.

\textbf{AutoML.} Many AutoML toolkits and systems have been developed~\cite{feurer2015efficient,Olson2016TPOT,H2O,li2020system,wang2021flaml}.
There is little work in the AutoML field that takes fairness into consideration, despite some preliminary attempt in the broad regime of AutoML recently. \cite{Schmucker2020} utilize a multi-objective HPO method to find a Pareto frontier of configurations which have good fairness and accuracy. FairHO~\cite{cruz2021promoting} tries to make hyperparameter optimization fairness-aware by including the fairness score as an additional objective. It can be used to either explore the Pareto frontier regarding fairness and prediction accuracy or find an optimal balance between accuracy and fairness based on a user-specified weight. Both the aforementioned two methods essentially add fairness as an additional objective to their hyperparameter optimization process.  FairBO~\cite{perrone2021fair} takes fairness as a constraint and proposes to reduce the unfairness of learning outcomes by varying the hyperparameter configurations and solving a constrained Bayesian optimization problem. In this work, we also take fairness as a constraint in AutoML. Our solution is more general than FairBO. 

\section{Fair AutoML formulation}\label{sec:formulation_ana}

In this section, we first introduce notions and notations to be used throughout this paper, and basic concepts of AutoML. We then investigate the necessity and impact of unfairness mitigation in AutoML. At last, we propose a fair AutoML formulation in which unfairness mitigation is incorporated as an organic component of the AutoML process. 

\begin{itemize} [leftmargin=*]
    \vspace{-1mm}
    \setlength\itemsep{-0.2em}
    \item $\bX$ and $\bY$ denote feature vectors and target values in the data feature space $\cX$ and label space $\cY$ respectively. 
    $D = (\bX, \bY)$ denotes a dataset in general, and $D_{\text{train}} = (\bX_{\text{train}}, \bY_{\text{train}})$ and  $D_{\text{val}}  = (\bX_{\text{val}}, \bY_{\text{val}})$ denote a particular training and validation dataset respectively.
    \item  $c$ denotes a hyperparameter configuration in a particular hyperparameter search space \(\cC\), with $c \in \cC$. When only a single ML learner is involved in the AutoML process, $c$ denotes model hyperparameter configuration of that concerned learner, and when multiple learners are involved, $c$ also includes a dimension about the choice of the learner.
    \item  $f$ denotes a machine learning model in general. $f_{c, D_{\text{train}}}$ is the resulting model associated with hyperparameter configuration $c$, trained on dataset $D_{\text{train}}$, and $f_{c, D_{\text{train}}}(\bX_{\text{val}})$ is the prediction outcome of $f_{c, D_{\text{train}}}$ on input data $\bX_{\text{val}}$. In the later discussion, we also add a superscript on $f_{c, D_{\text{train}}}$ to denote how the model is trained. Specifically, we use $f^{(0)}_{c, D_{\text{train}}}$ to denote the variant of the model which is trained (based on  $c$ and $D_{\text{train}}$) without considering fairness constraints, and use $f^{(1)}_{c, D_{\text{train}}}$ to denote the variant which is trained with unfairness mitigation.
    \item $\texttt{Loss}: \cY \times \cY \to \mathbb{R}$ is a loss function. 
\end{itemize}

\subsection{AutoML}
Given a training dataset $D_{\train}$, a validation dataset $D_{\val}$, and a hyperparameter search space $\cC$, the goal of AutoML\footnote{In this work, we focus on this commonly used formulation of AutoML. The scope of AutoML in general can be beyond this.} is to find the best model which has the lowest validation loss:
\begin{align} \label{eq:automl_formulation}
    \min_{c \in \cC} \Loss(f_{c, D_{\train}}(D_{\text{val}}), Y_{\val})
\end{align}

Solutions to the AutoML problem typically decide which hyperparameter configuration to try in iterative trials and adjust their future decisions toward finding better configurations. To facilitate further analysis, we provide an abstraction for such solutions via \HPSearcher. A \HPSearcher is essentially responsible for making a hyperparameter configuration suggestion in the given search space whenever requested. 
We use $\HPSearcher.\suggest$ to denote the procedure for making a hyperparameter configuration suggestion, and $\HPSearcher.\update$ to denote the procedure for updating necessary information needed. This abstraction is compatible with almost all hyperparameter search methods~\cite{bergstra2011algorithms,bergstra2012random, snoek2012practical,li2020system, wu2021frugal,wang2021economical} for AutoML.

Under this AuotML context, it may seem straightforward that to find a fair model with good validation loss, one can simply add a fairness constraint on top of Eq.~\eqref{eq:automl_formulation} and perform constrained hyperparameter optimization, as what is done in a recent work~\cite{perrone2021fair}.
However, one caveat in this approach is that, for an ad-hoc task, one may not be able to find a single model that both satisfies the fairness constraint and has a good loss by simply tuning the hyperparameters $c \in \cC$ in AutoML. This essentially is because there may exist multiple sources of unfairness \cite{dudk2020assessing}, many of which are just impossible to be alleviated by simply varying model hyperparameters. 

\subsection{Unfairness mitigation in AutoML: necessity and impact}

 \begin{figure*}[ht]
        \subfigure[Unfairness mitigation's impact in tuning multiple (7) learners]{
            \includegraphics[width=.5\textwidth]{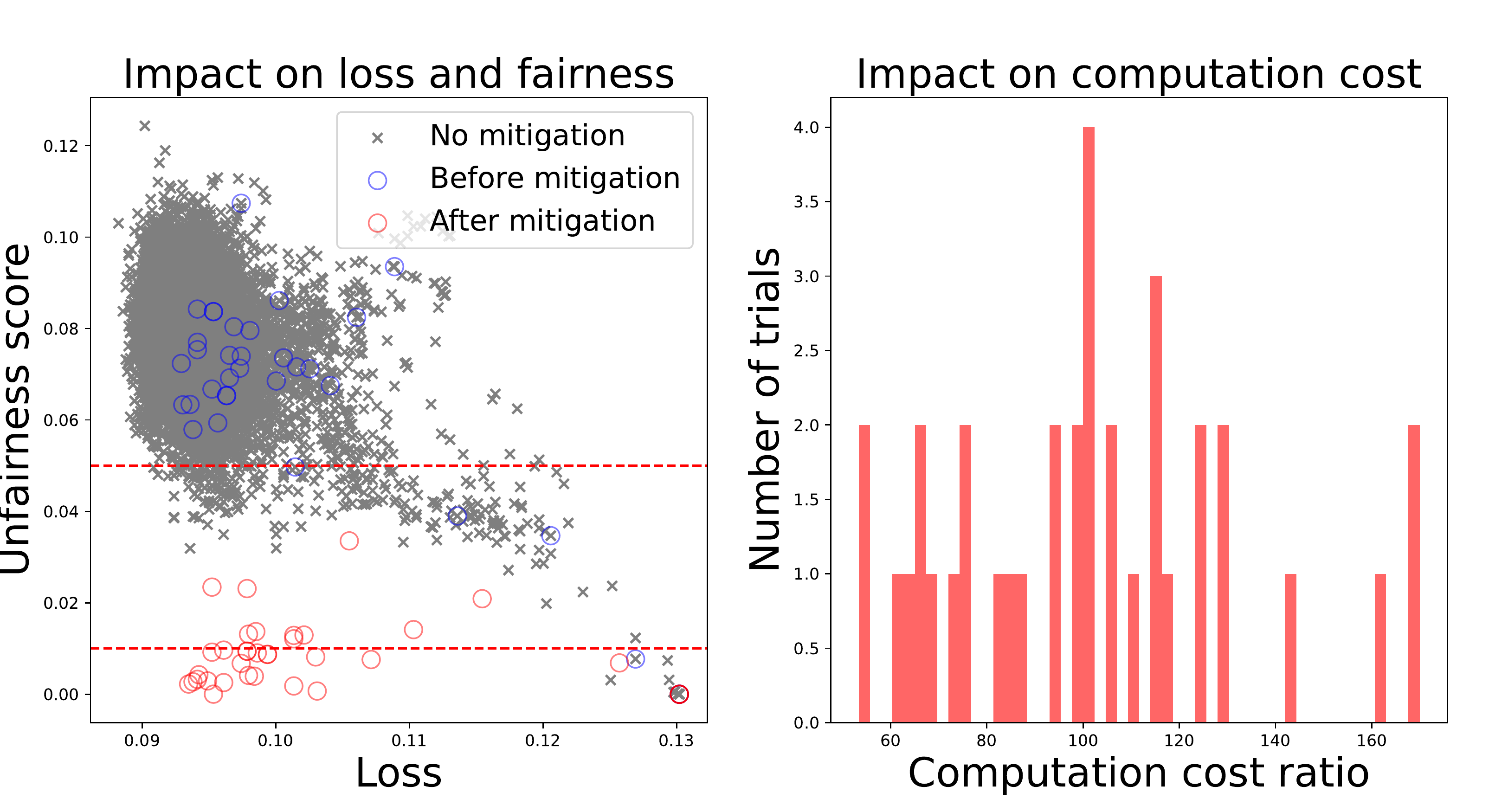}
            }
            \subfigure[Unfairness mitigation's impact in tuning XGBoost]{
            \includegraphics[width=.5\textwidth]{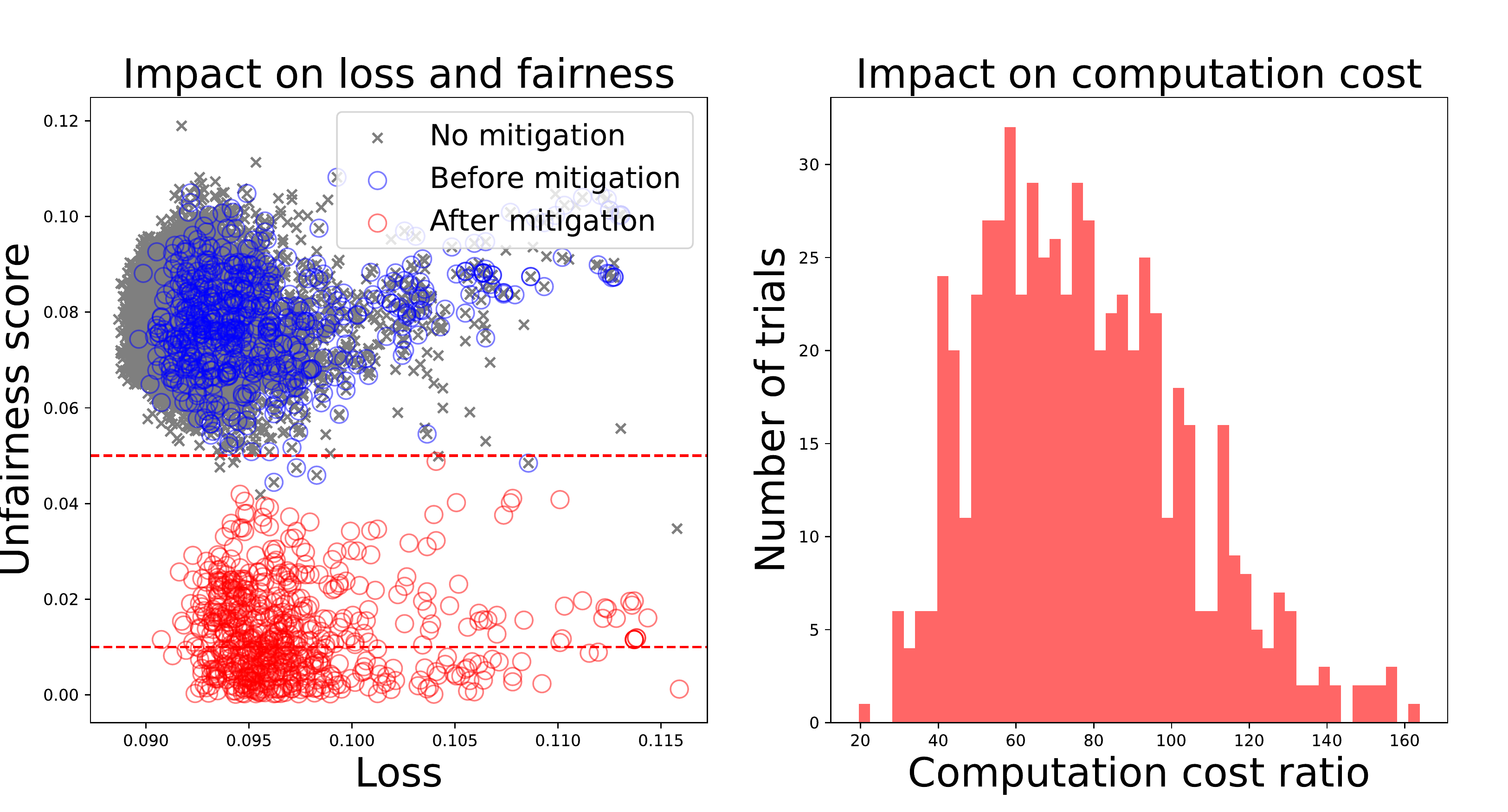}
            }
        \caption{In scatter plots of both (a) and (b), the cross-marked scatters labeled `No mitigation' correspond to models trained in the first experiment, i.e., $\cH = \{ f^{(0)}_{c_i, D_{\train}}\}_{i \in [1, 2, ...]}$; the circle-marked scatters denote models from the second experiments, i.e., $\tilde \cH $. The scatters labeled `Before mitigation' and `After mitigation' correspond to models with the same hyperparameter configuration but trained without unfairness mitigation, i.e., $f^{(0)}_{c_i, D_{\train}}$ and with unfairness mitigation, i.e., $f^{(1)}_{c_i, D_{\train}}$. The horizontal lines correspond to unfairness scores of 0.01 and 0.05 respectively, which are two commonly used thresholds when specifying fairness constraints. The 2nd sub-figure of both (a) and (b) show the distribution of the `computation cost ratio' (the cost of training a model with mitigation, i.e., $f^{(1)}_{c_i, D_{\train}}$ divided by that of training a regular model with the same configuration, i.e., $f^{(0)}_{c_i, D_{\train}}$).}
\label{fig:mitigation_analysis_new}
\end{figure*}

The aforementioned limitations of generic hyperparameter tuning in AutoML in terms of improving fairness motivate us to consider taking dedicated algorithmic unfairness mitigation methods, e.g., the pre-, in-, and post-processing methods mentioned in Sec.~\ref{sec:related_work} into AutoML. 

\textbf{Necessity of unfairness mitigation in AutoML.} 
To verify the necessity of unfairness mitigation in AutoML, we conduct the following two experiments on the \emph{Bank} dataset (details about this dataset are deferred to Section~\ref{sec:exp}). In both experiments, we use the AutoML solution from an open-source library named FLAML~\cite{wang2021flaml}.

\begin{itemize}
\vspace{-1mm}
    \setlength\itemsep{-0.2em}
\item  In the first experiment, we perform a regular AutoML process for 4 hours. Following our previous notations, we get a set of ML models in this experiment denoted by $\cH = \{ f^{(0)}_{c_i, D_{\train}}\}_{i \in [1, 2, ...]}$, in which $c_i$ is the configuration tried at the $i$-th iteration and the largest iteration is determined by the 4-hours time budget.
\item In the second experiment, within the same time budget, i.e., 4 hours, for each hyperparameter configuration $c_i$ proposed by the original AutoML process at iteration $i$, in addition to training a regular ML model $f^{(0)}_{c_i, D_{\train}}$, we also train a second model with unfairness mitigation, i.e., $f^{(1)}_{c_i, D_{\train}}$. We use a state-of-the-art in-processing method named Exponentiated Gradient (EG)~\cite{agarwal2018reductions} as the mitigation method. In this experiment, we get the following set of models $\tilde \cH = \{ f^{(0)}_{c_i, D_{\train}}, f^{(1)}_{c_i, D_{\train}}\}_{i \in [1, 2, \dots]}$. Note that because of the additional unfairness mitigation step, the total number of trials in this experiment is presumably smaller than that in the first one.
\end{itemize}
We perform the two experiments on two tuning tasks: one tune both learner selection and each learner's model hyperparameters, and one tune a single learner XGBoost's\footnote{Note that this single learner setting is common in practical scenarios where due to infrastructure or deployment constraints, the institution or company can only use a certain type of model.} model hyperparameters.  We plot the results from the two tasks in Figure~\ref{fig:mitigation_analysis_new}(a) and (b), respectively. 
In the first sub-figure of Figure~\ref{fig:mitigation_analysis_new}(a) and (b) entitled ``Impact on loss and fairness", we plot the `unfairness score', i.e., Difference of Statsistical Parity (DSP) and loss of the models evaluated in both experiments.
From the scatter plots in Figure~\ref{fig:mitigation_analysis_new}, we have the following observations: (1) We first observe an obvious overall reduction of unfairness scores after mitigation is applied (by comparing the `Before mitigation' and `After mitigation' points) in both tuning tasks. It shows the effectiveness of unfairness mitigation in reducing the unfairness of each single machine learning model for different types of learners. (2) By comparing the `Before mitigation' and `After mitigation' points together with the `No mitigation' points, we can see the necessity of performing unfairness mitigation in AutoML: Although by simply performing learner selection and model hyperparameter tuning, one can find models with different degrees of fairness (as shown by the grey crosses in the scatter plots), the loss of the models is not necessarily good. For example, when 0.01 is used as the threshold in the unfairness constraint, one only gets models with very bad predictive performance in the tuning task in Figure~\ref{fig:mitigation_analysis_new}(a), and one cannot even find a single fair model satisfying the constraint when the learner is restricted to XGBoost. With unfairness mitigation, one can easily produce fair models with a decent loss.
These results confirm the necessity and effectiveness of including unfairness mitigation in AutoML. We now move on to analyze \emph{how} one should incorporate it into AutoML effectively and efficiently. To answer this question, we find it necessary to first examine the impact of unfairness mitigation methods in terms of computation overhead and predictive performance, which are crucial factors that directly affect the final performance of AutoML. 

\textbf{Unfairness mitigation's impact on ML and AutoML.}  
A recent study~\cite{islam2022through} provides a comprehensive investigation on the impact of  unfairness mitigation in machine learning via experimental evaluation on a wide range of unfairness mitigation methods in the fair classification task. The impact can be understood from the following three aspects. (1) \textbf{Fairness:} Most of the state-of-the-art unfairness mitigation methods can effectively mitigate the unfairness. But none of the mitigation methods are guaranteed to make every model fair for every dataset. (2) \textbf{Predictive performance:} In all unfairness mitigation methods, the mitigation of unfairness comes with the degradation of predictive performance, and the trade-off is complex. (3) \textbf{Computation overhead:} Unfairness mitigation methods, especially pre- and in-processing approaches generally incur high computation overhead. We also observe this severe issue on computation overhead in our case study according to the cost histograms in Figure~\ref{fig:mitigation_analysis_new}.

This complex impact makes the incorporation of unfairness mitigation to AutoML highly non-trivial: On the one hand, we want to perform unfairness mitigation to improve the fairness of resulting ML models. On the other hand, performing unfairness mitigation brings in computation overhead and degrades the predictive performance of the models. A subtle balance needs to be made.

\subsection{Fair AutoML with unfairness mitigation} \label{subsec:formulation}

\textbf{Abstractions.} We first develop two fairness-related abstractions. These abstractions are helpful in facilitating the formulation and analysis of the fair AutoML problem. 

Fairness assessment via \Fair: 
We abstract the fairness assessment of a machine learning model $f$, given a validation dataset $D$, as a procedure that calculates a fairness indicator based on the model $f$'s prediction outcome on the dataset $D$ according to a particular quantitative fairness definition. We use $\Fair(f, D)$ to denote this procedure in general. Both group fairness and individual fairness, which are the two dominating types of fairness definitions, fit into this abstraction. This abstraction can also handle the case where the fairness goal is to satisfy multiple fairness constraints. Note that in the context of a particular concrete fairness definition, additional information other than the model prediction and the dataset might be needed. 
\begin{exmp}[Group fairness]
Under the group fairness context, 
a model is considered `fair' if the output of the disparity measurement regarding the sensitive attribute(s) does not exceed a particular threshold. We denote by $A$ the sensitive attribute variable, $\texttt{G}(\cdot ; A)$ a disparity function (e.g., statistical disparity) parameterized by sensitive attribute $A$, and $\delta$ a fairness threshold constant. Given $(A, \texttt{G}, \delta)$, an implementation of the abstraction can be $ \Fair(f, D) = \mathbbm{1}\{\texttt{G}(\bX, \bY, \hat f(D); A) \leq \delta \}$, in which $\mathbbm{1}\{\cdot\}$ is an indicator function.
\end{exmp}

Unfairness mitigation via \UFMitigation: 
We abstract the unfairness mitigation of a machine learning algorithm given a training dataset $D$ as a procedure that intervenes in the model building process (including data transformation, model training, and inference rules) in a way such that the final model's outputs are pushed toward a particular fairness target (specified via \Fair~ function following our abstraction). All the pre-, in-, and post-processing unfairness mitigation introduced earlier fit into this abstraction.  Following notations introduced earlier, $f^{(1)}_{c,D}$ is obtained via this mitigation procedure, i.e., $f^{(1)}_{c, D} \gets \UFMitigation(f, c, D, \Fair)$.

\textbf{Fair AutoML problem formulation.} 
Given datasets $D_{\text{train}}$ and $D_{\text{val}}$,  a loss function $\texttt{Loss}$, a fairness function $\Fair$ and a particular unfairness mitigation method following the \UFMitigation abstraction, we formulate the fair AutoML problem as finding a fair machine learning model minimizing the loss by searching over a set of hyperparameters and deciding whether to do unfairness mitigation, as mathematically expressed below,
\begin{align} \label{eq:fair_aml_formulation}
    \min_{c \in \cC, m \in \{0, 1\}} \Loss(f^{(m)}_{c, D_{\text{train}}}(\bX_{\text{val}}), \bY_{\text{val}}) \\ \nonumber
    s.t.~~ \Fair(f^{(m)}_{c, D_{\text{train}}}, D_{\text{val}}) 
\end{align}
In the rest of the paper we use $\Loss^{(m)}_{c} $ and $\Fair^{(m)}_{c} $ as shorthand for $ \Loss( f^{(m)}_{c, D_{\text{train}}}(\bX_{\text{val}}), \bY_{\text{val}}) $  and $ \Fair(f^m_{c, D_{\text{train}}}, D_{\text{val}})$ respectively without ambiguity. Under this formulation, we can consider fair AutoML as an AutoML problem optimizing for the \emph{fair loss}, i.e., the loss when fairness constraint is satisfied. This \emph{fair loss} notion, denoted as $L_{\text{fair}}$, will be used though this paper to evaluate the effectiveness of fair AutoML solutions.

The proposed formulation has many desirable properties: The unfairness mitigation abstraction decouples the complexity of regular AutoML and unfairness mitigation. On the one hand, this makes it easy to leverage existing research and development efforts on unfairness mitigation. On the other hand, this framing makes it fairly easy to achieve fair AutoML by augmenting an existing AutoML system.

Despite the desirable properties of the proposed formulation, solving the problem is still challenging due to the following reasons:  (1) Solving an AutoML problem itself is notoriously expensive. It typically involves a large number of expensive trials. (2) The large computation overhead compounds this computation challenge by making some of the trials with unfairness mitigation as large as 10x to 100x more expensive.

\section{Fair AutoML framework and system} \label{sec:framework}

We first propose a general fair AutoML framework to solve the fair AutoML problem characterized in Eq.~\eqref{eq:automl_formulation}. The proposed framework is presented in Alg.~\ref{alg:fair_automl_general}, the central component of which is an abstraction named \FairnessManager. 
The main responsibility of \FairnessManager is to suggest a promising hyperparameter configuration $c \in \cC$ and decide whether to perform unfairness mitigation, indicated by $m=0$ or $m=1$ at each trial. This responsibility is realized via a $\suggest\xspace$ function (line 4 of Alg.~\ref{alg:fair_automl_general}).  
The system then builds a machine learning model $f^{(m)}_{c}$ based on the suggested $c$ and $m$ (line 5 of Alg.~\ref{alg:fair_automl_general}), and gets validation loss and fairness on the validation dataset (line 6 of Alg.~\ref{alg:fair_automl_general}). 
Notice that the model training and validation step typically incurs a non-trivial cost, denoted by $\kappa_{c}^{(m)}$. The framework then records the incurred cost $\kappa_{c}^{(m)}$, updates the budget left and records the detailed trial observation, including the suggested hyperparameter configuration $c$, mitigation decision $m$, and the corresponding loss $\Loss_{c}^{(m)}$ , fairness $\Fair^{(m)}_{c}$ and computation cost $\kappa_{c}^{(m)}$,  in $\cH$. Then \FairnessManager will be updated via an $\update\xspace$ function such that the new observation can be used to help with future suggestions.

\subsection{\FairnessManager}


\textbf{Insights.} \FairnessManager shall be designed in a way such that it is able to make suggestions on hyperparameter configuration and unfairness mitigation toward a fair model with good predictive performance. 
Generic hyperparameter tuning methods (following the \HPSearcher abstraction) can be effective in finding models with good predictive performance. And unfairness mitigation methods (following the \UFMitigation abstractions) can be used to further reduce the unfairness of the models. We believe it is beneficial to take the best of both worlds.
Based on this insight, we propose to include  \HPSearcher and \UFMitigation as sub-modules of the \FairnessManager, where \HPSearcher is primarily responsible for proposing hyperparameter configuration $c \in \cC$, and \FairnessManager is responsible for reducing the unfairness of each individual model. We further develop an additional logic to control whether and when to perform unfairness mitigation instead of hyperparameter search.

\begin{algorithm}[t]
\caption{FairAutoML}\label{alg:fair_automl_general}
\begin{algorithmic}[1]
\STATE \textbf{Inputs:} Training and validation data, search space $\cC$, and resource budget $B$ (optional). 
\STATE \textbf{Initialization:} $\cH = []$
\WHILE{$B > 0$}
\STATE $c, m \gets  \FairnessManager.\texttt{suggest}(\cH)$
\STATE Model building: train model $ f^{(m)}_{c}$ with (when $m=1$) or without (when $m=0$) unfairness mitigation 
\STATE  Model validation: Get $\Loss^{(m)}_c$ and $\Fair^{(m)}_c$  
\STATE Audit the incurred cost $\kappa^{(m)}_c$, and update budget $B \gets B - \kappa^{(m)}_{c}$, and historical observations $\cH \gets \cH \cup (c, m, \kappa^{(m)}_c, \Loss^{(m)}_c, \Fair^{(m)}_c)$
\STATE $\FairnessManager.\update(\cH)$.
\ENDWHILE
\end{algorithmic}
\end{algorithm}

\begin{algorithm}[t]
\caption{\FairnessManager.\suggest($\cC$)}\label{alg:fair_suggest}
\begin{algorithmic}[1]
\STATE $\cH^{(0)} = \{ c \in \cH | m_{c} =0\}$
\STATE $c' \gets \argmin_{c \in  \cH^{(0)} \& \neg \Fair_c} \Loss_{c}$
\IF{ $c'$ and $\ECF^{(1)} < \ECF^{(0)}$ and $\bar L_c < L_{fair}^*$ } 
    \STATE  $m \gets 1$, $c \gets c'$
\ELSE
\STATE  $m \gets 0$, $c \gets  \HPSearcher.\suggest \cC)$
\ENDIF
\STATE Return $c, m$
\end{algorithmic}
\end{algorithm}

\begin{algorithm}[t]
\caption{\FairnessManager.\update($\cH$)}\label{alg:fair_update}
\begin{algorithmic}[1]
\STATE $\cH^{(0)} = \{ c \in \cH | m_{c} =0\},  \cH^{(1)} = \cH - \cH^{(0)}$
\STATE Update $\ECF^{(0)}$ based on $\cH^{(0)}$, and update $\ECF^{(1)}$ based on $\cH^{(1)}$ according to their definitions
\STATE  $\HPSearcher.\update(\cH')$
\end{algorithmic}
\end{algorithm}


We present the two major functionalities of the proposed \FairnessManager in Alg.~\ref{alg:fair_suggest} and Alg.~\ref{alg:fair_update} respectively. 
Each time when the $\FairnessManager.\suggest$ function is involved, the \FairnessManager first selects the evaluated but unmitigated hyperparmeter configuration which has the lowest loss and does not satisfy the fairness constraint, as the \emph{candidate} hyperparmeter configuration to perform unfairness mitigation (line 2 of  Alg.~\ref{alg:fair_suggest}).  \FairnessManager then finalizes the decision after checking several conditions (line 3-6 of Alg.~\ref{alg:fair_suggest}): if the conditions for doing unfairness mitigation are satisfied, the \FairnessManager suggests to do unfairness mitigation, i.e., $m=1$, on the candidate hyperparameter configuration $c'$ (line 4); otherwise it invokes the \HPSearcher to suggest model hyperparmeter and set the mitigation decision to $m=0$ (line 6). 

These two choices have different implications to the system. Regular hyperparameter search provides good candidate models to perform unfairness mitigation and sometimes can even directly find better fair models. But the fairness of the models obtained is not guaranteed. 
For example, when tuning XGBoost, by simply tuning hyperparameters one cannot find a single model with unfairness score smaller than 0.01 according to the scatter plot in Figure~\ref{fig:mitigation_analysis_new}(b).  Further unfairness mitigation is still needed. Unfairness mitigation is in general useful in producing a model that is fair. However it may degrade the accuracy of the resulting model, and brings in potentially high computation overhead. The high computation overhead may make the system unable to evaluate enough models to navigate a model with good accuracy when the computation budget is limited. 
The conditions in line 3 of Alg.~\ref{alg:fair_suggest} are designed to make 
a subtle balance between these two choices: it is an adaptive strategy based on online estimations of the utility of regular hyperparameter search and unfairness mitigation.

\textbf{Insights for line 3 of $\FairnessManager.\suggest~$in Alg.~\ref{alg:fair_suggest}:} Considering the potential loss degradation and high computation overhead of unfairness mitigation, the algorithm should not do unfairness mitigation on the configuration $c'$ unless both the following two conditions hold: (1) When doing unfairness mitigation is more `efficient' than doing regular hyperparameter search via \HPSearcher in terms of improving fair loss. In general, there may exist cases where hyperparameter search is already good enough to find fair models with good loss, for example, when the fairness constraint is easy to satisfy. In this case, unfairness mitigation is not needed unless it is more efficient in improving the fair loss than hyperparameter search. (2) When the \FairnessManager can anticipate a fair loss improvement by doing mitigation on the candidate configuration $c$. Ideally, we want to spend resources in unfairness mitigation on the configurations that can yield better fair loss than the current best fair loss. Although such information, i.e., the fair loss of a configuration after mitigation, is not available \textit{a priori}, the original loss before mitigation is usually a meaningful observable indicator of it.

To realize the first condition, we use $\ECF^{(m)}$, which is analogous to the $\emph{CostImp}$ function proposed in ~\cite{wang2021economical}, to characterize the fair loss improvement efficiency of both regular hyperparameter search and unfairness mitigation. It essentially approximates the \textbf{E}stimated \textbf{C}ost needed for producing a model with a better \textbf{F}air loss by doing unfairness mitigation ($\ECF^{(1)}$), or hyperparameter tuning with the \HPSearcher (corresponds to $\ECF^{(0)}$). Unfairness mitigation should not be conducted unless $\ECF^{(1)} < \ECF^{(0)}$.  One nice property of this strategy is that it can achieve a self-adaptive balance between mitigation and hyperparameter search: Once one choice becomes less efficient we turn to the other choice. We include the  definition and detailed explanation of $\ECF^{(m)}$ in Appendix~\ref{appendix:detail}. 

To realize the second condition, we propose to make a projection on the fair loss that can be achieved and the resource needed for the unfairness mitigation before actually performing it. For a particular configuration $c$, we denote by $\bar L_c$ the projected loss after mitigation. Specifically, we estimate $\bar L_c $ by $\bar L_c = \Loss^{(0)}_c + \zeta_{\cH_t, B}$ where $\zeta_{\cH_t, B}$ is an online estimation of the loss degradation of unfairness mitigation based on historical observations in $\cH_t$ and budget $B$. The algorithm performs unfairness mitigation only when $\bar L_c < L_{\text{fair}}^*$, in which $L_{\text{fair}}^*$ is the best fair loss achieved so far by the system. Intuitively speaking, the algorithm tries to skip unfairness mitigation on some of the models if doing is unlikely to yield better fair loss.
We provide the detailed formula of $\bar L_c$ in Appendix~\ref{appendix:detail}.



Considering both conditions, we use $\{ \ECF^{(1)} < \ECF^{(0)} ~~ \& ~~\bar L_c < L_{\text{fair}}^*\}$ (line 3 of Alg.~\ref{alg:fair_suggest}) as the decision rule to determine whether to perform unfairness mitigation.

Once the decisions on which hyperparameter configuration to try and whether to perform unfairness mitigation is made (line 4 of Alg.~\ref{alg:fair_automl_general} via \FairnessManager.\suggest), the algorithm proceeds to build a model and performs model validation accordingly (line 5 and line 6 of Alg.~\ref{alg:fair_automl_general}). The algorithm also records the computation cost incurred during the model building and validation step, the resulting model's predictive loss and fairness, and updates them as part of the historical observations in $\cH$. The remaining budgets will also be updated accordingly.
At last, the algorithm updates the \FairnessManager with the historical observations $\cH$ via \FairnessManager.\update~ (line 8 of Alg.~\ref{alg:fair_automl_general}). \FairnessManager.\update~ is presented in Alg.~\ref{alg:fair_update}. It updates the statistics needed for the \FairnessManager to make suggestions on unfairness mitigation, and also invokes \HPSearcher.\update~ to update the \HPSearcher.

\begin{figure*}
  \centering
\subfigure[Fair loss on \emph{Adult}]{
\includegraphics[width=0.25\textwidth]{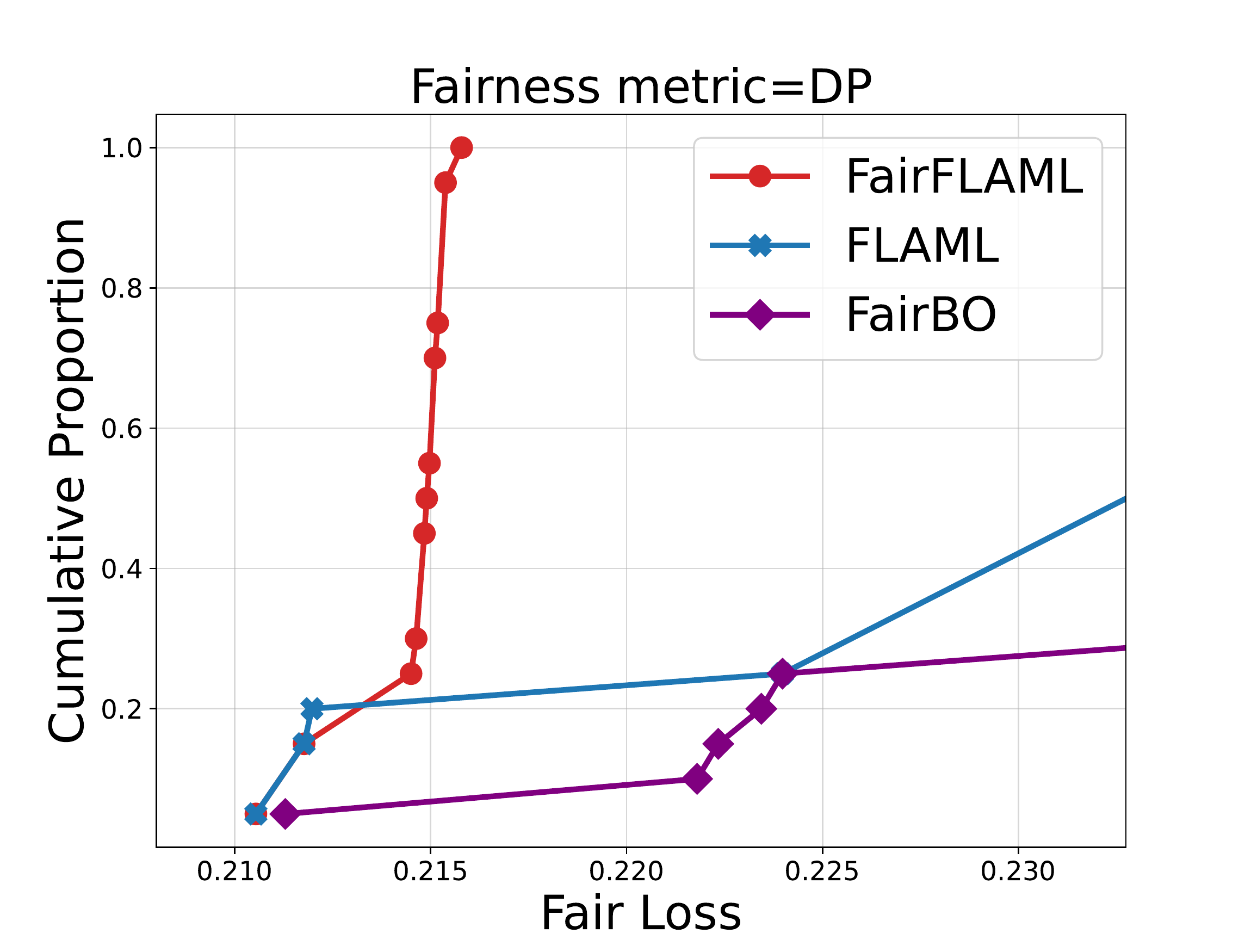}%
\includegraphics[width=0.25\textwidth]{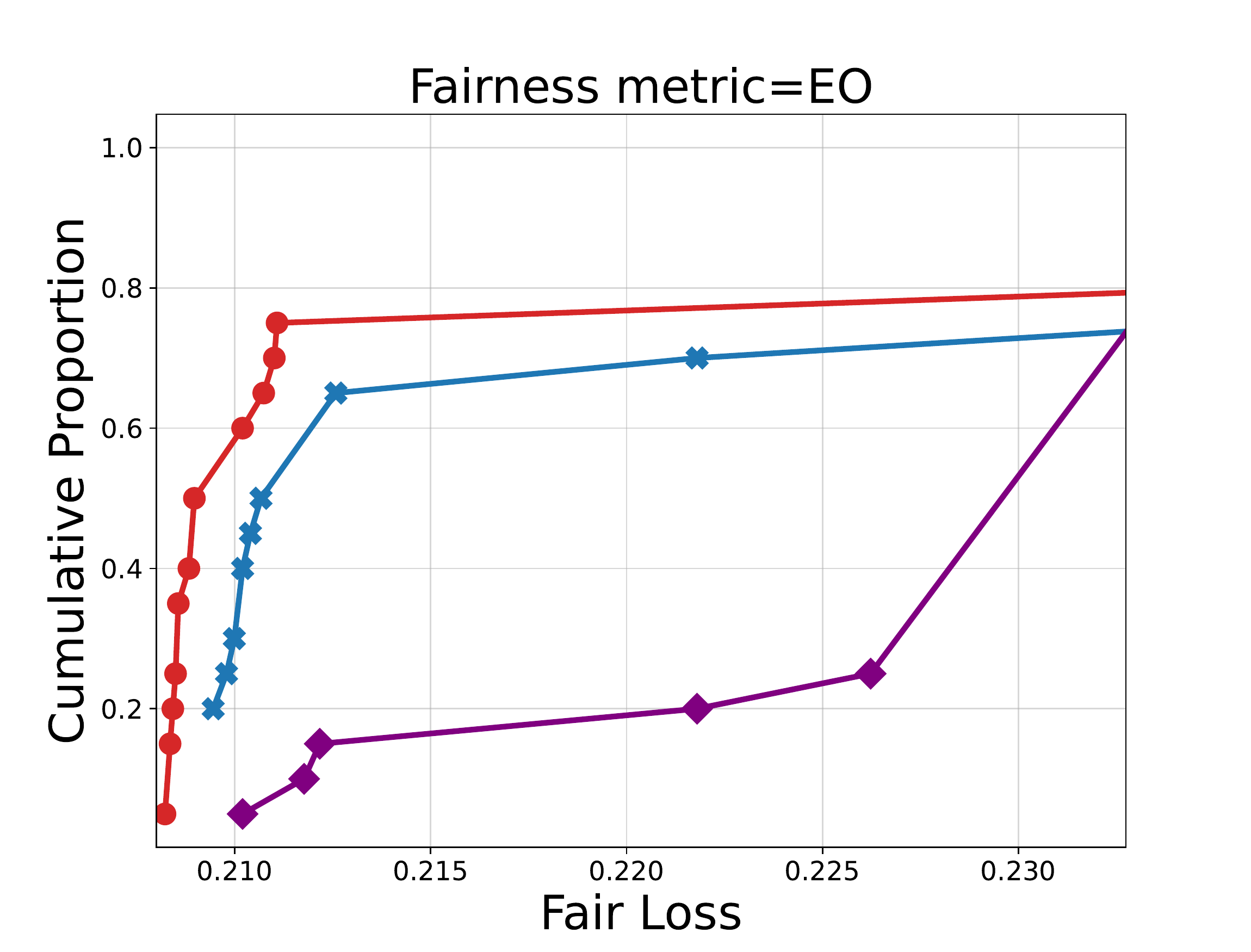}%
\includegraphics[width=0.25\textwidth]{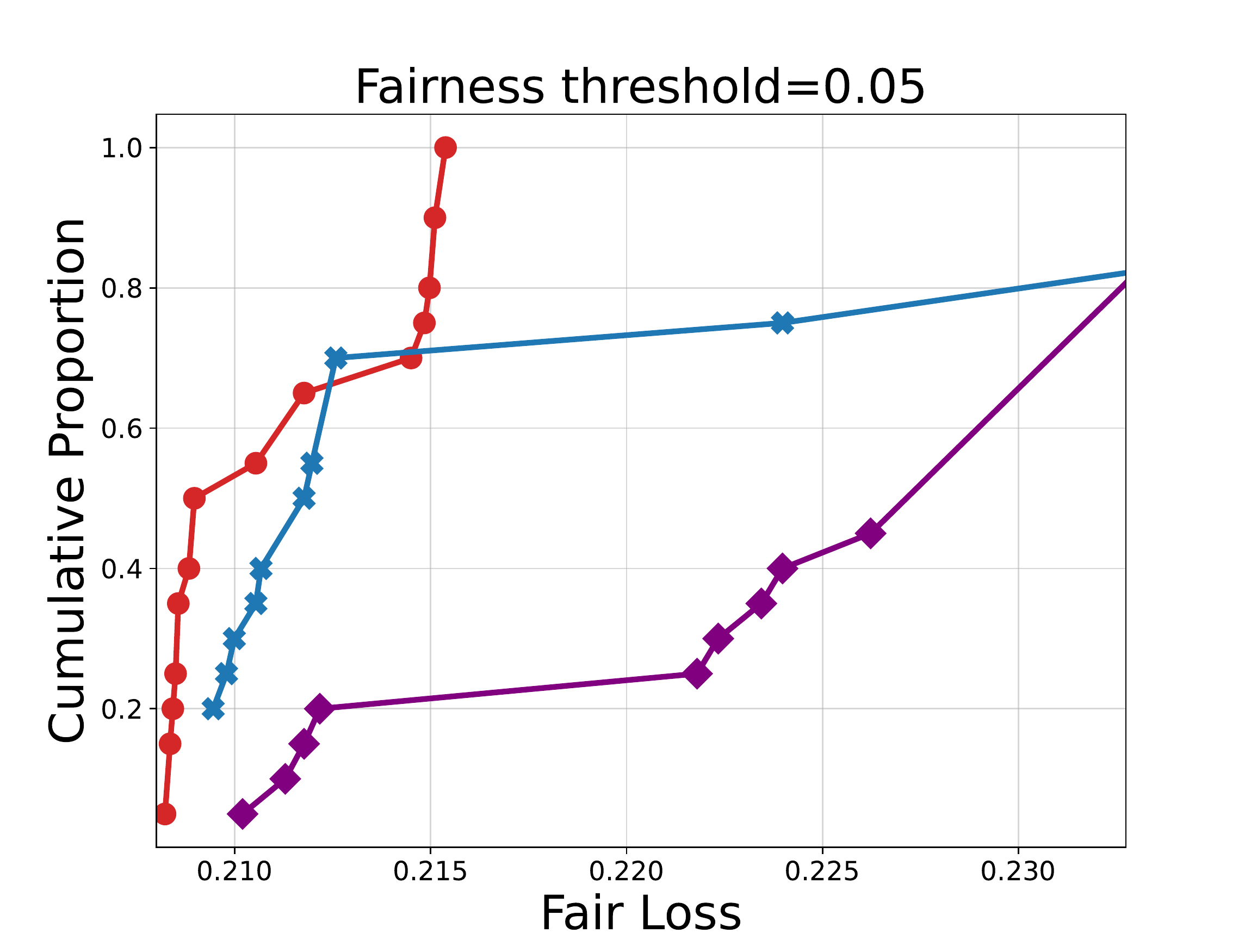}%
\includegraphics[width=0.25\textwidth]{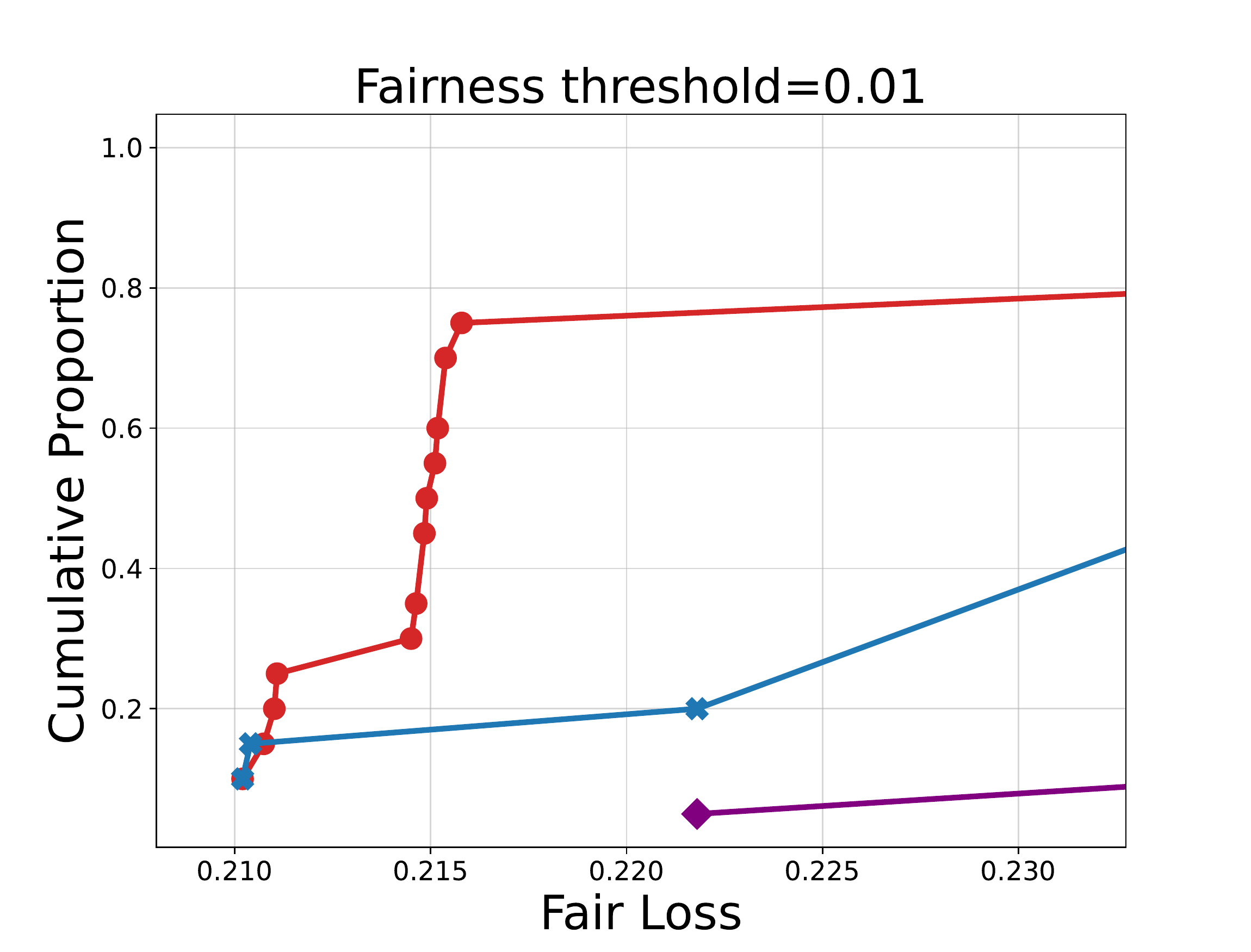}%
}
\\
\subfigure[Fair loss on \emph{Bank}]{
\includegraphics[width=0.25\textwidth]{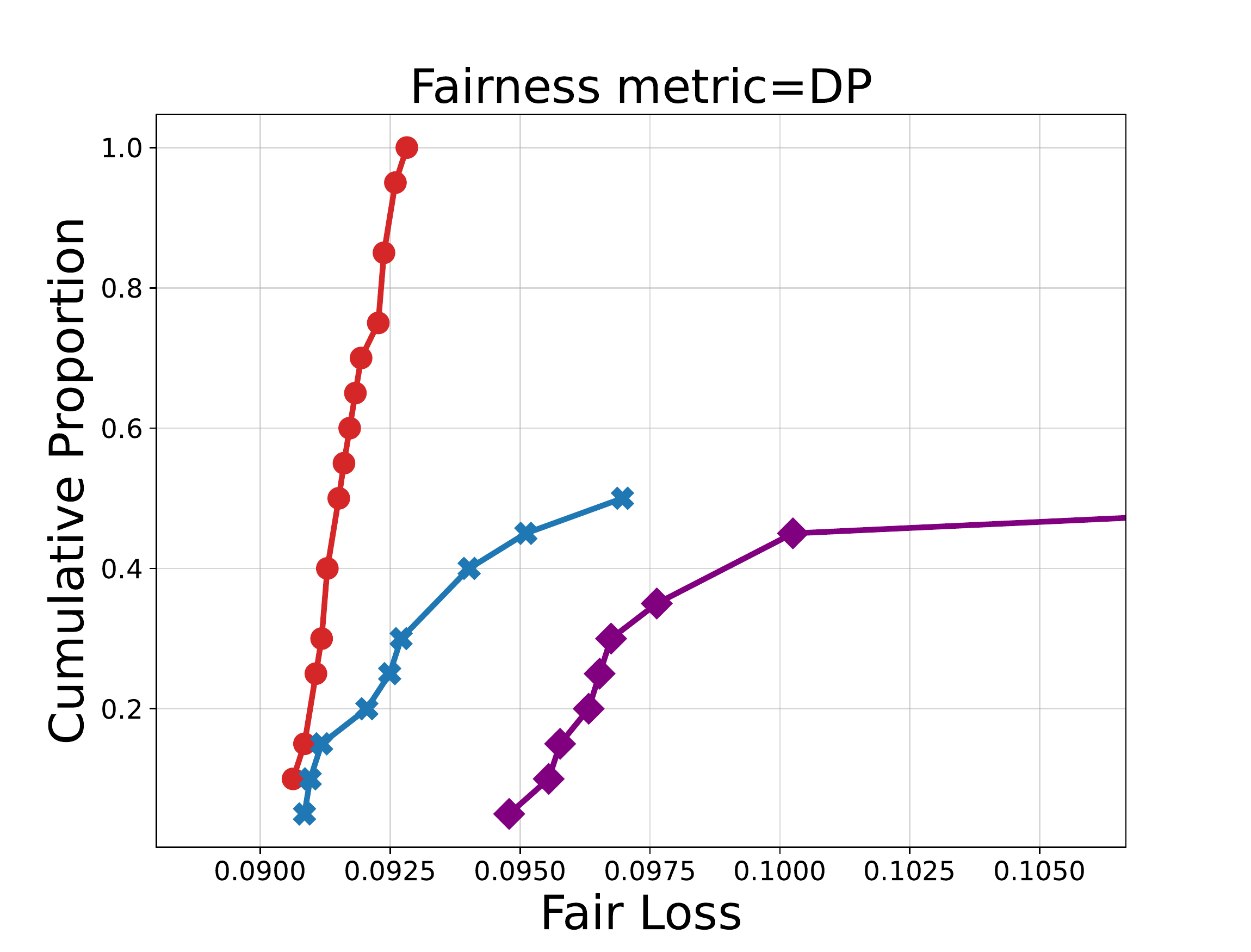}%
\includegraphics[width=0.25\textwidth]{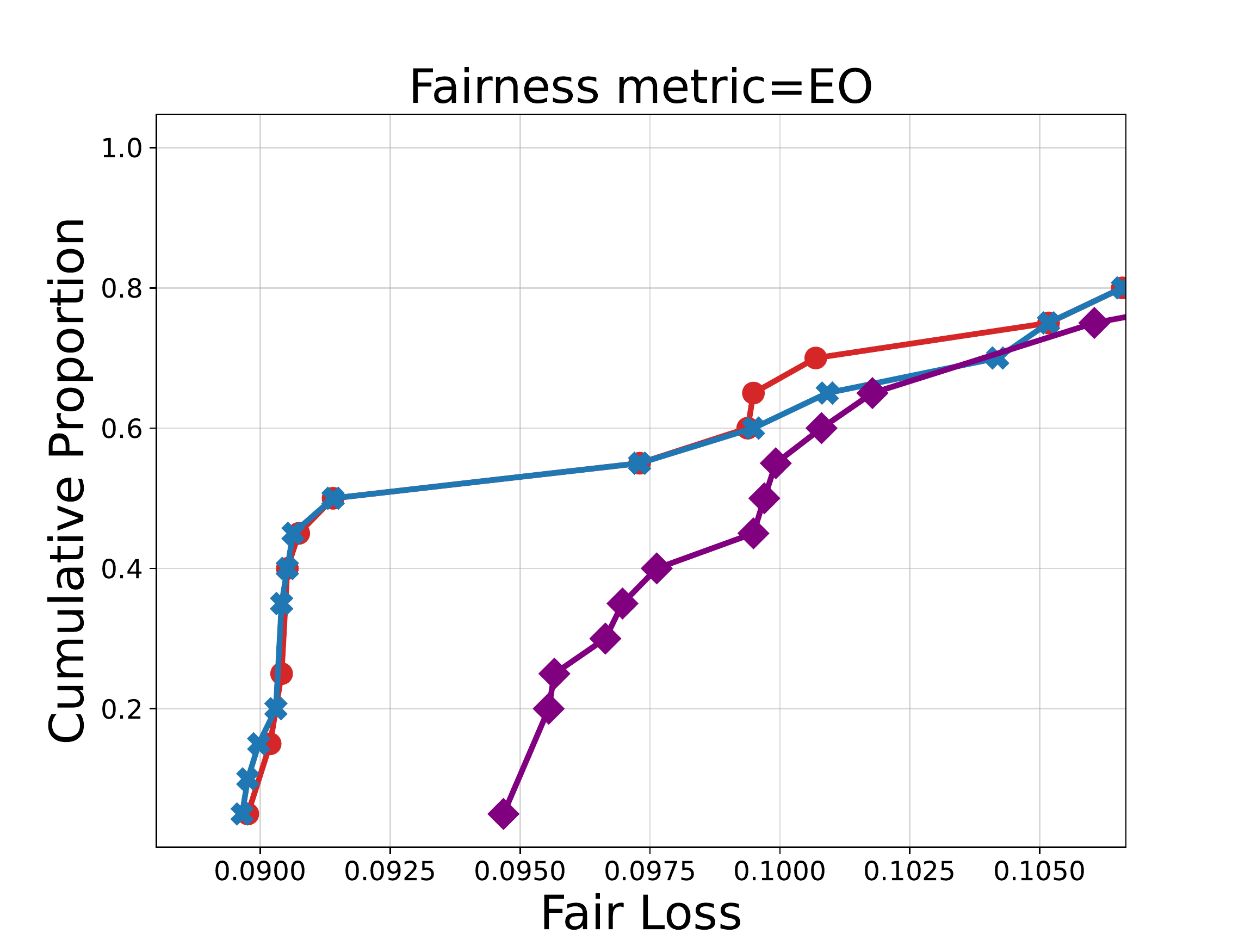}%
\includegraphics[width=0.25\textwidth]{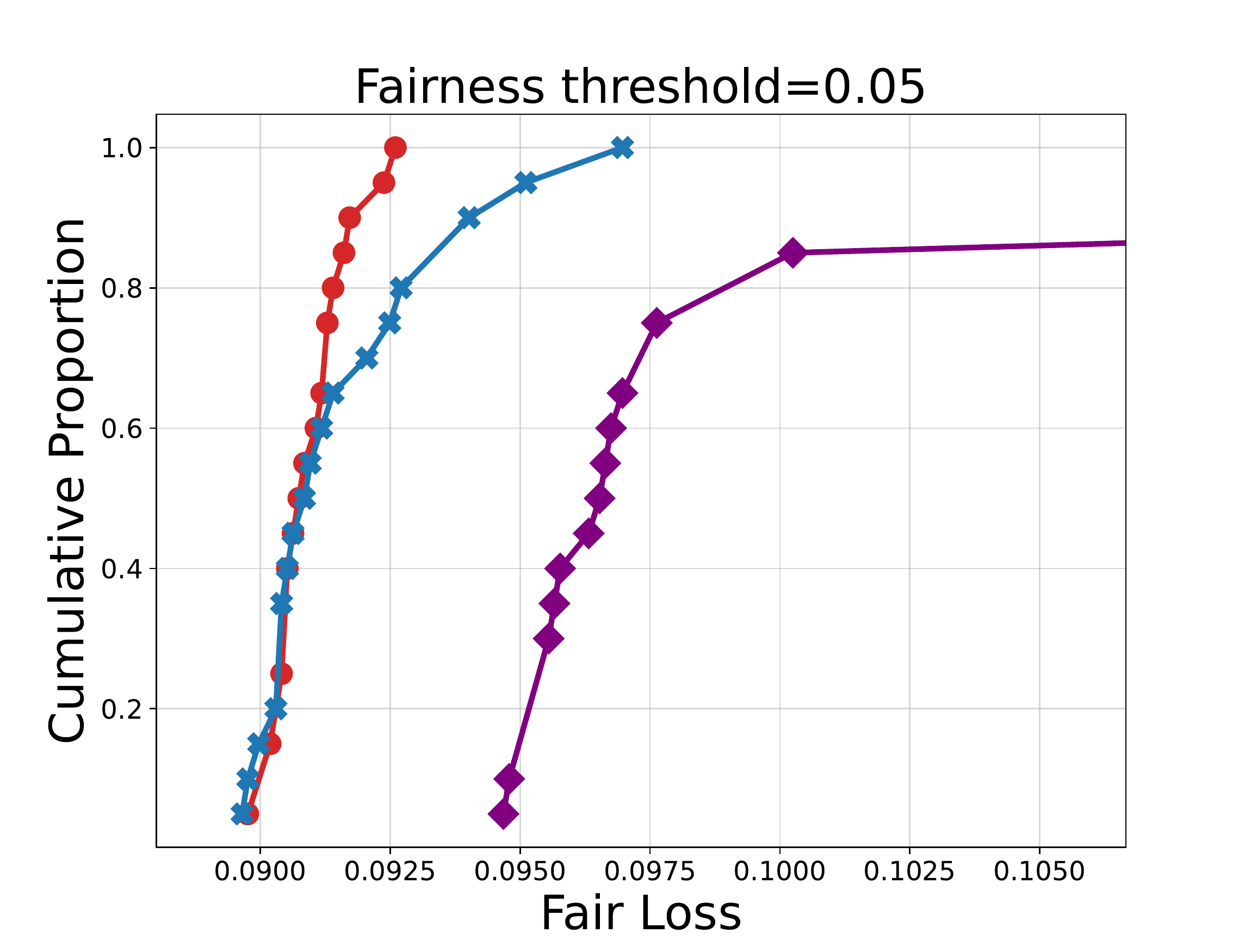}%
\includegraphics[width=0.25\textwidth]{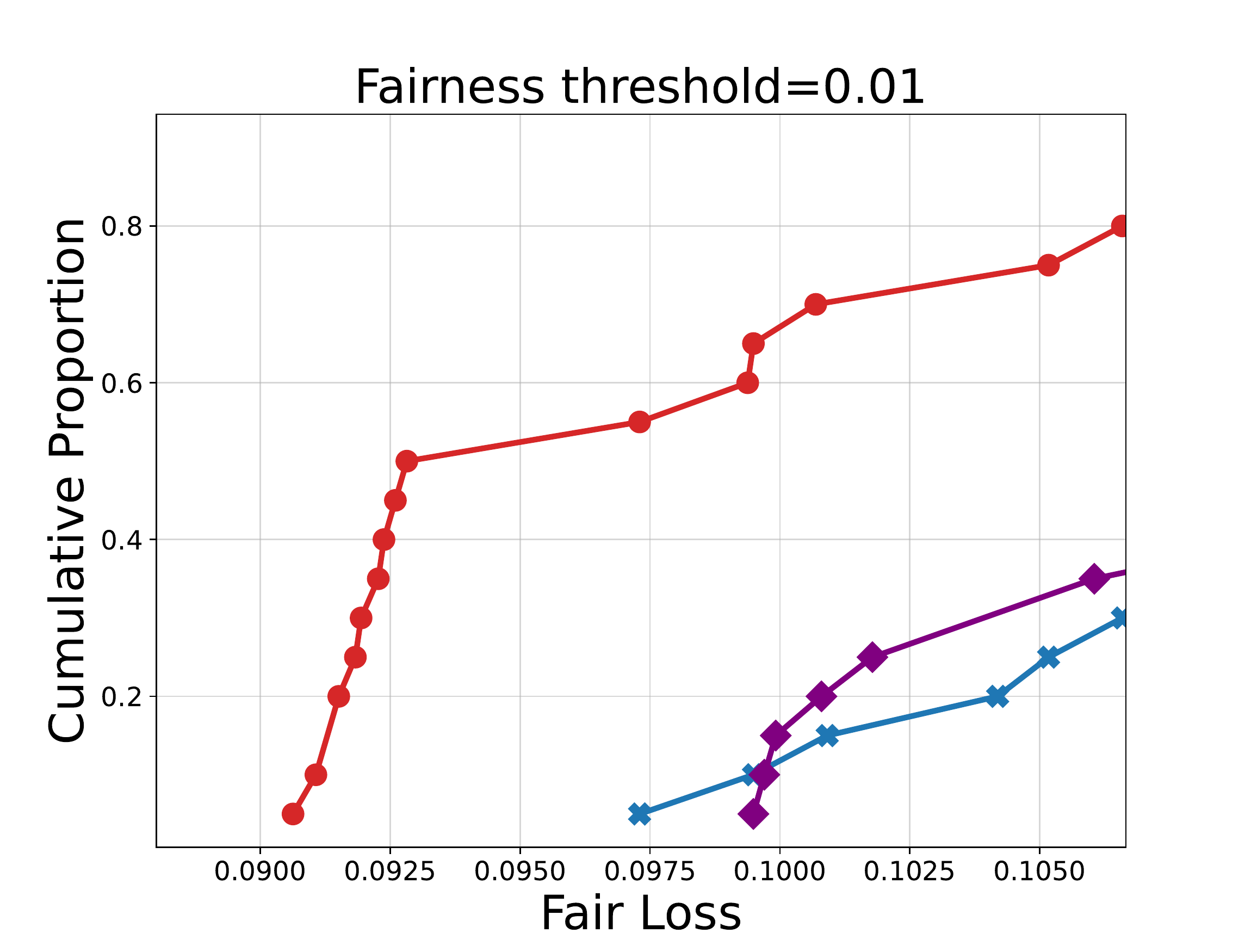}%
}
\\
\subfigure[Fair loss on \emph{Compas}]{
\includegraphics[width=0.25\textwidth]{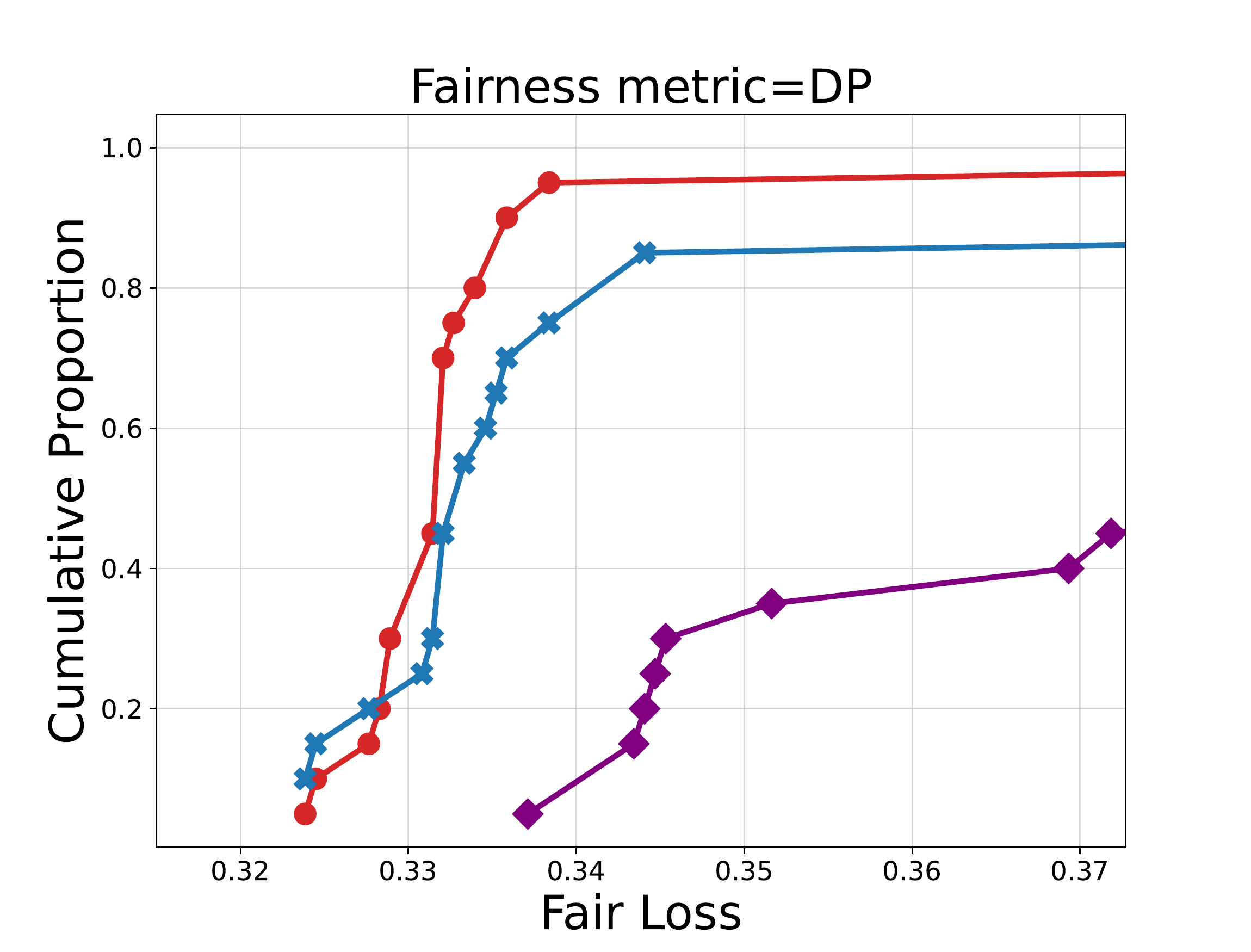}%
\includegraphics[width=0.25\textwidth]{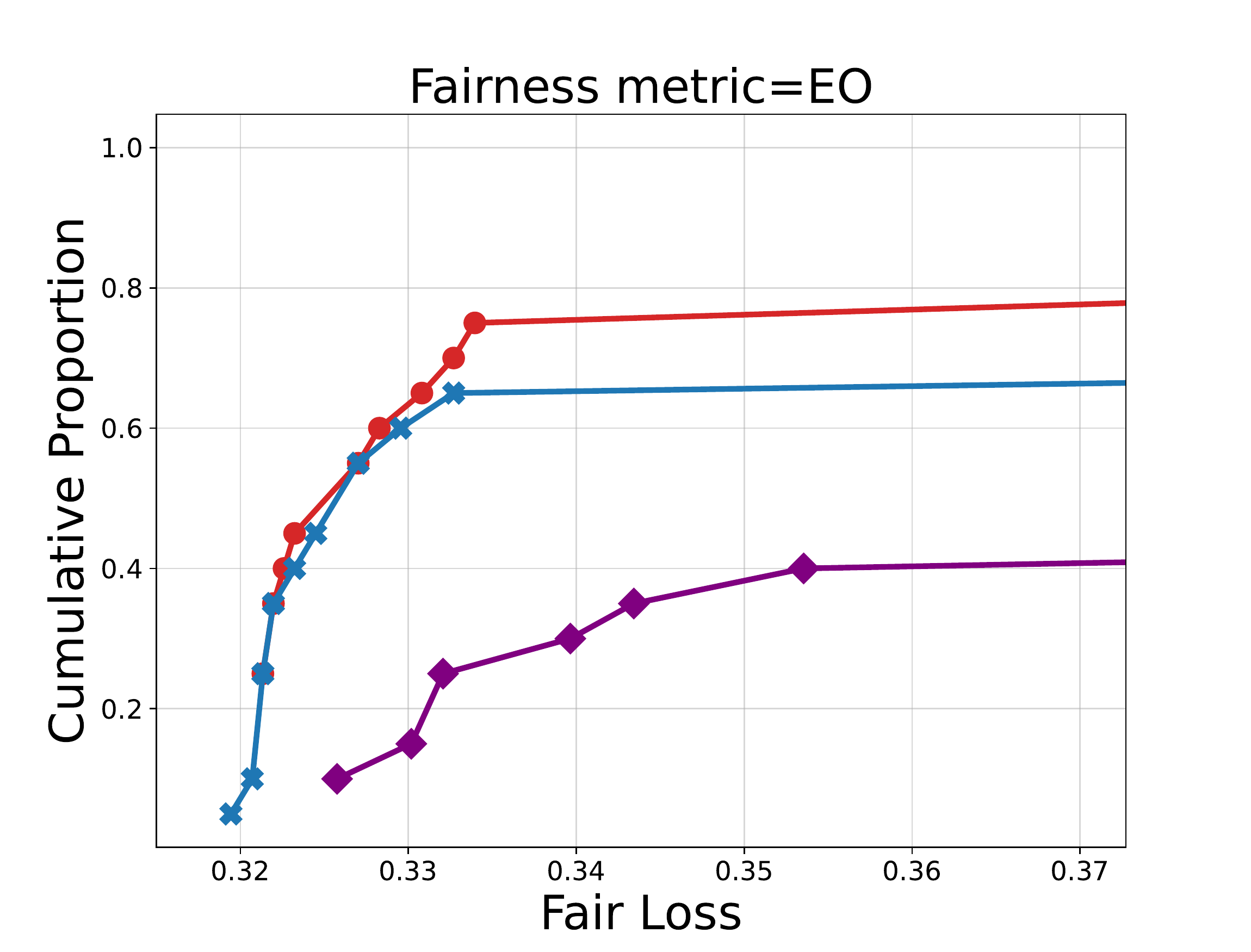}%
\includegraphics[width=0.25\textwidth]{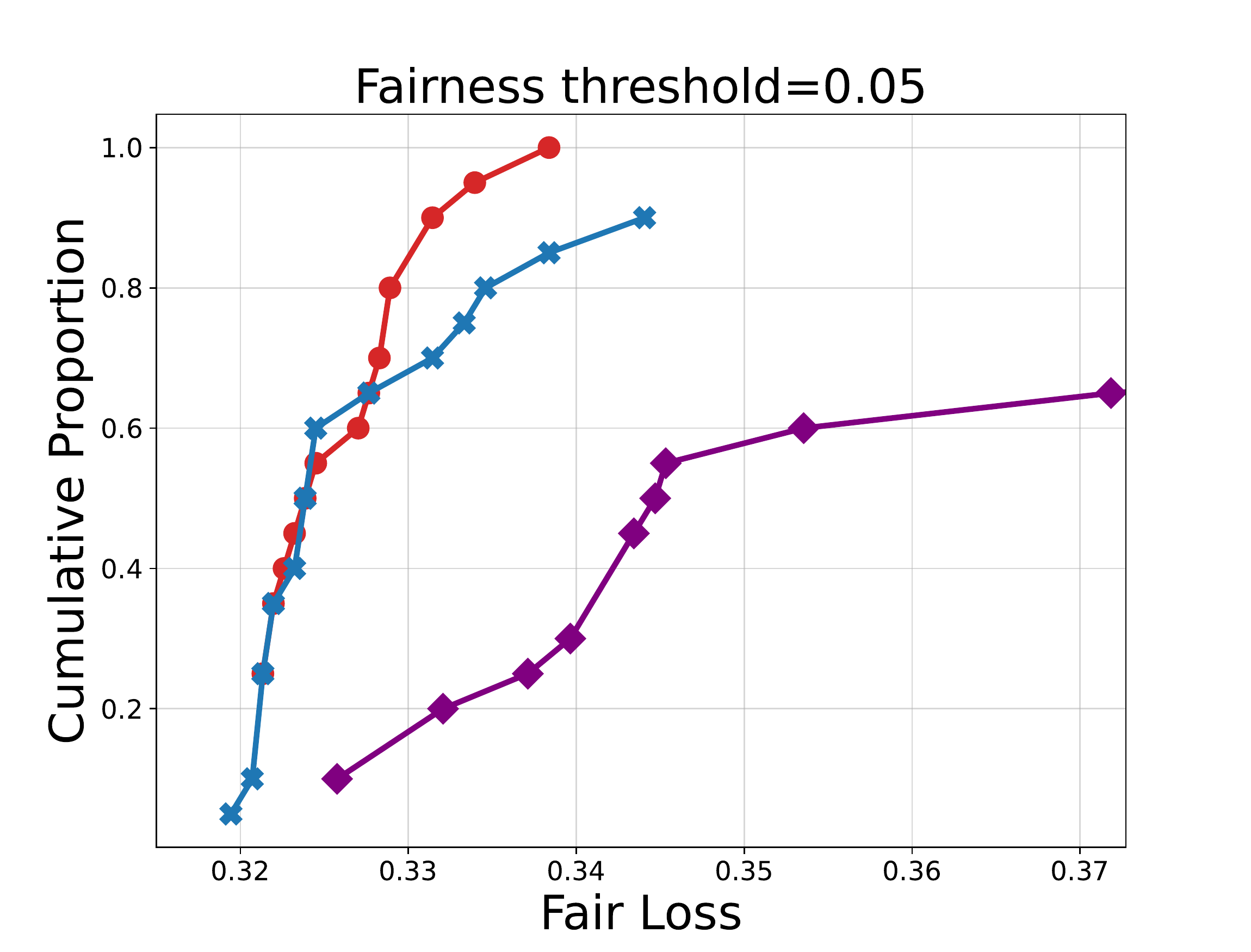}%
\includegraphics[width=0.25\textwidth]{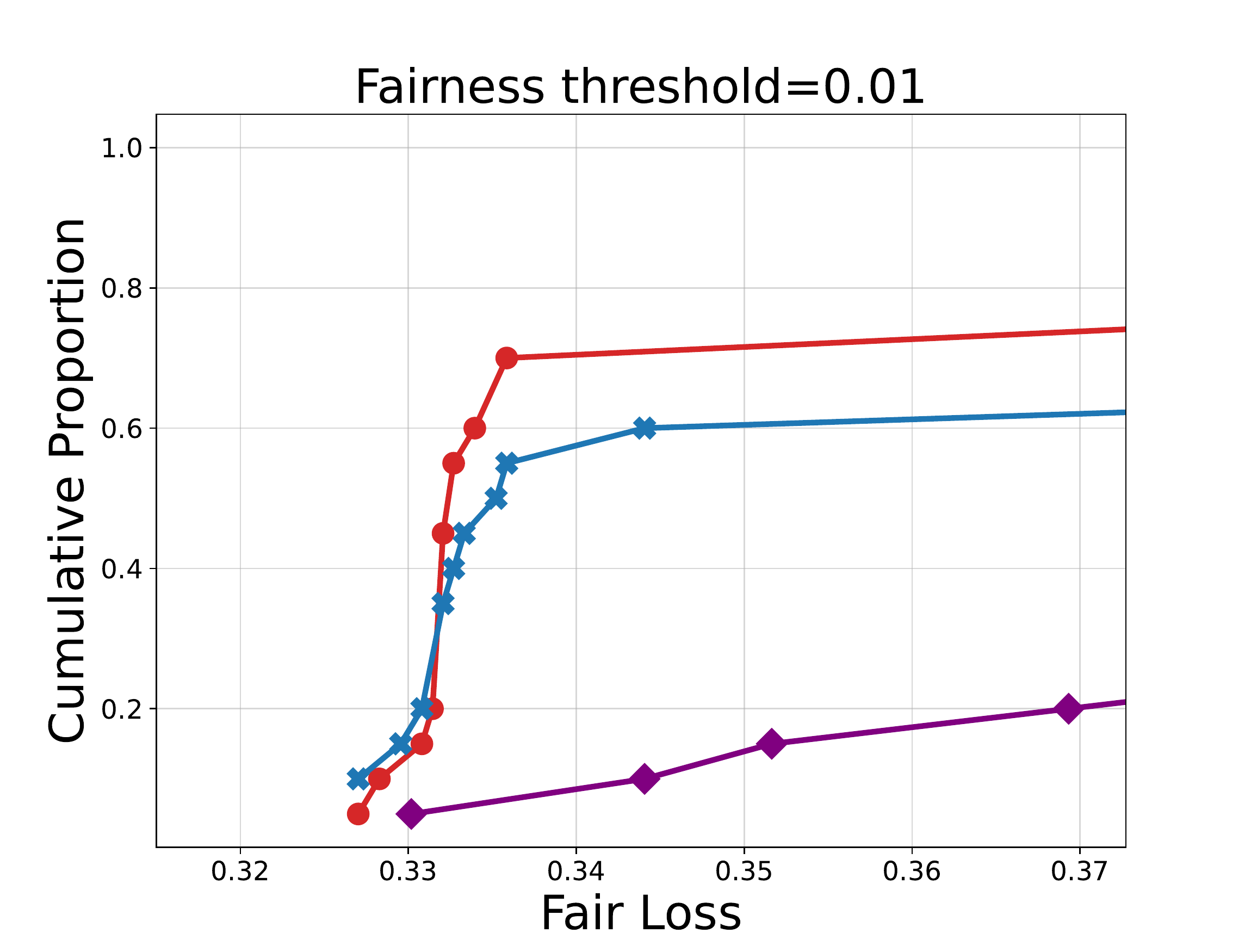}%
}
\\
\subfigure[Fair loss on \emph{MEPS}]{
\includegraphics[width=0.25\textwidth]{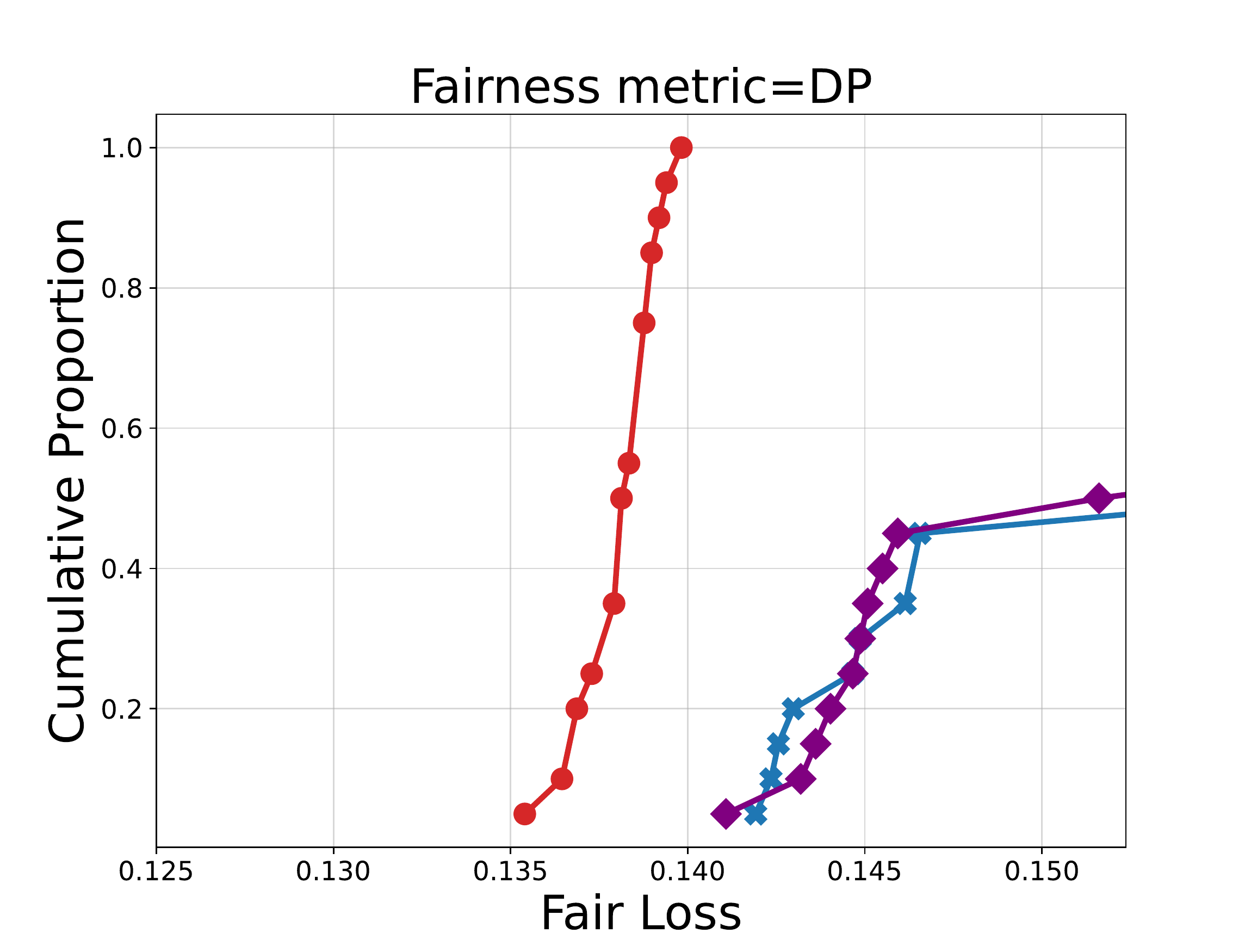}%
\includegraphics[width=0.25\textwidth]{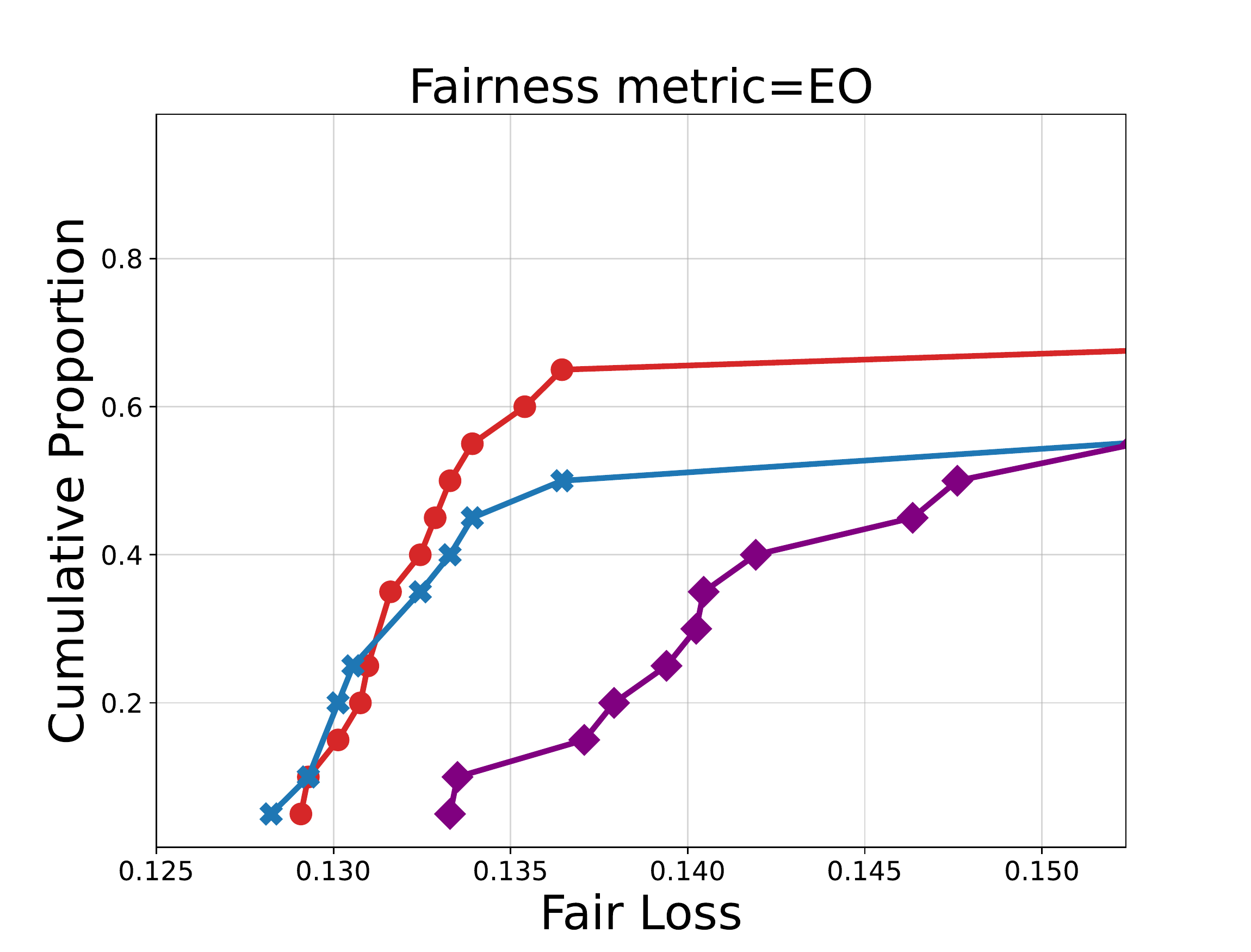}%
\includegraphics[width=0.25\textwidth]{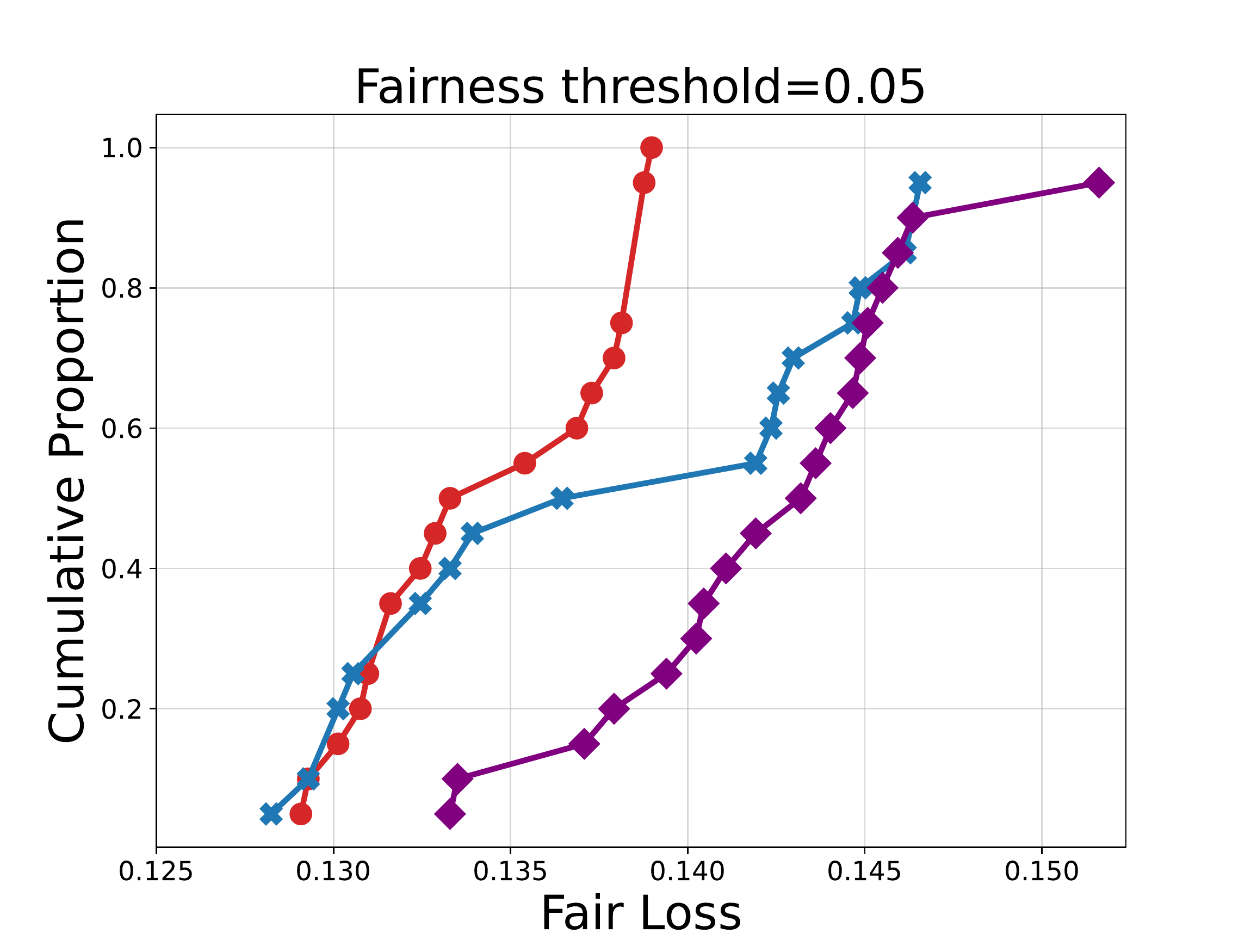}%
\includegraphics[width=0.25\textwidth]{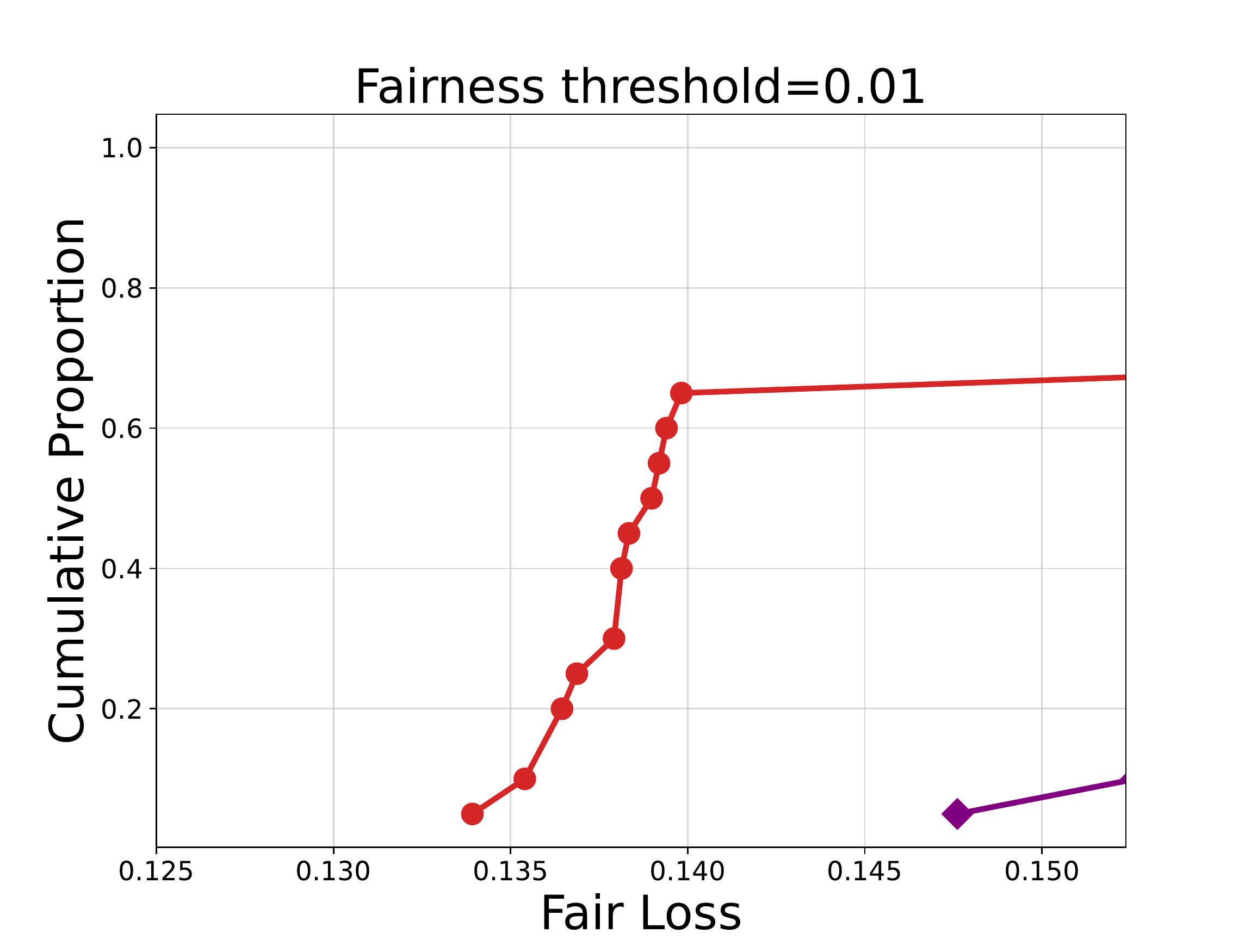}%
}
\\
\caption{Fair loss under different fairness settings when tuning XGBoost on different datasets. Each sub-figure shows the fair losses' cumulative proportion in 20 experiments under a particular fairness metric or threshold indicated by subfigure titles: (2 fairness settings) $\times$ (10 random seeds). We use $1.2 \times $ (best loss) as the x-axis (showing the fair loss) upper bound to improve visibility of the curves.}
 \label{fig:exp_res_cdf}
\end{figure*}

    \begin{figure*}
 \centering
  \includegraphics[width=0.25\textwidth]{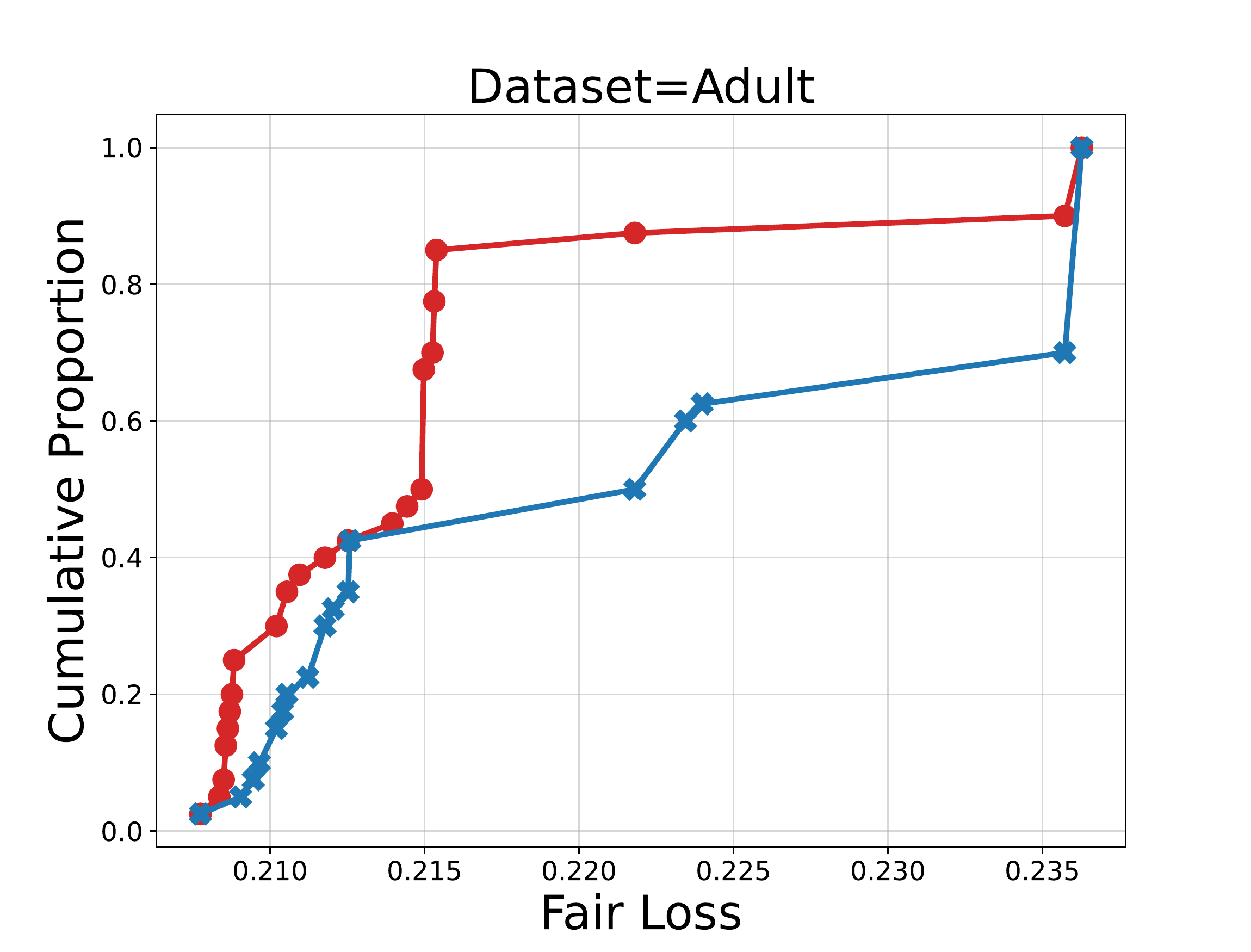}%
  \includegraphics[width=0.25\textwidth]{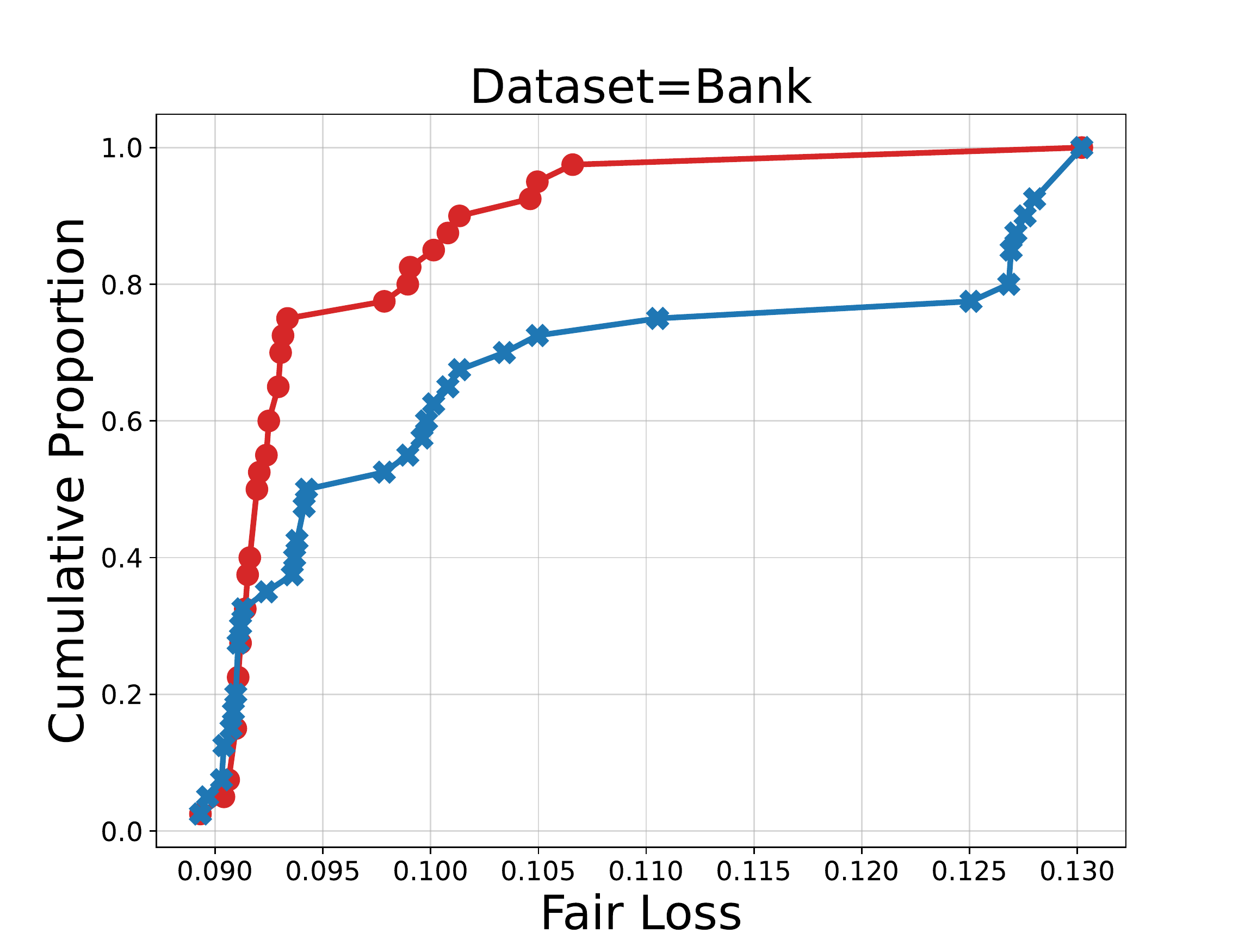}%
  \includegraphics[width=0.25\textwidth]{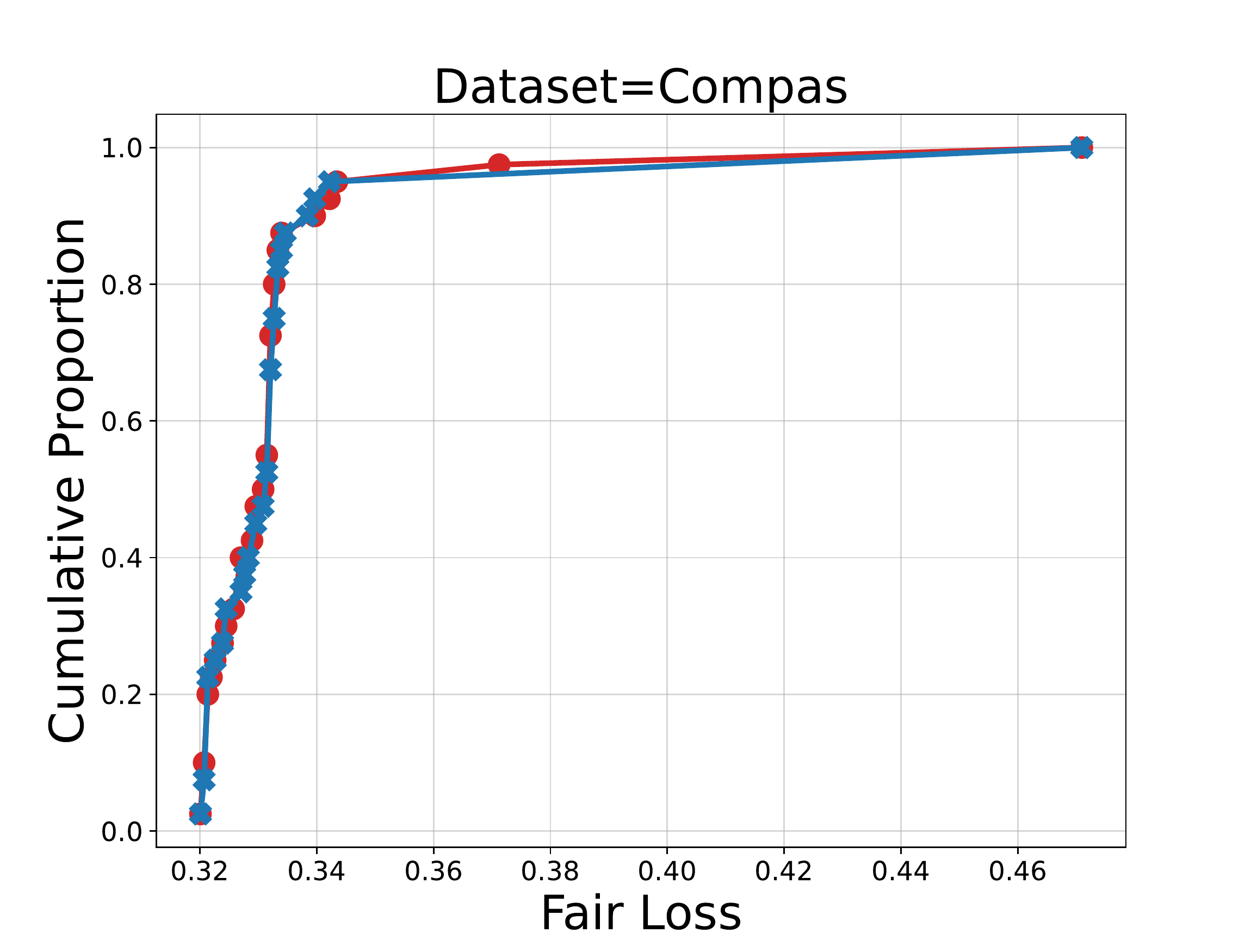}%
  \includegraphics[width=0.25\textwidth]{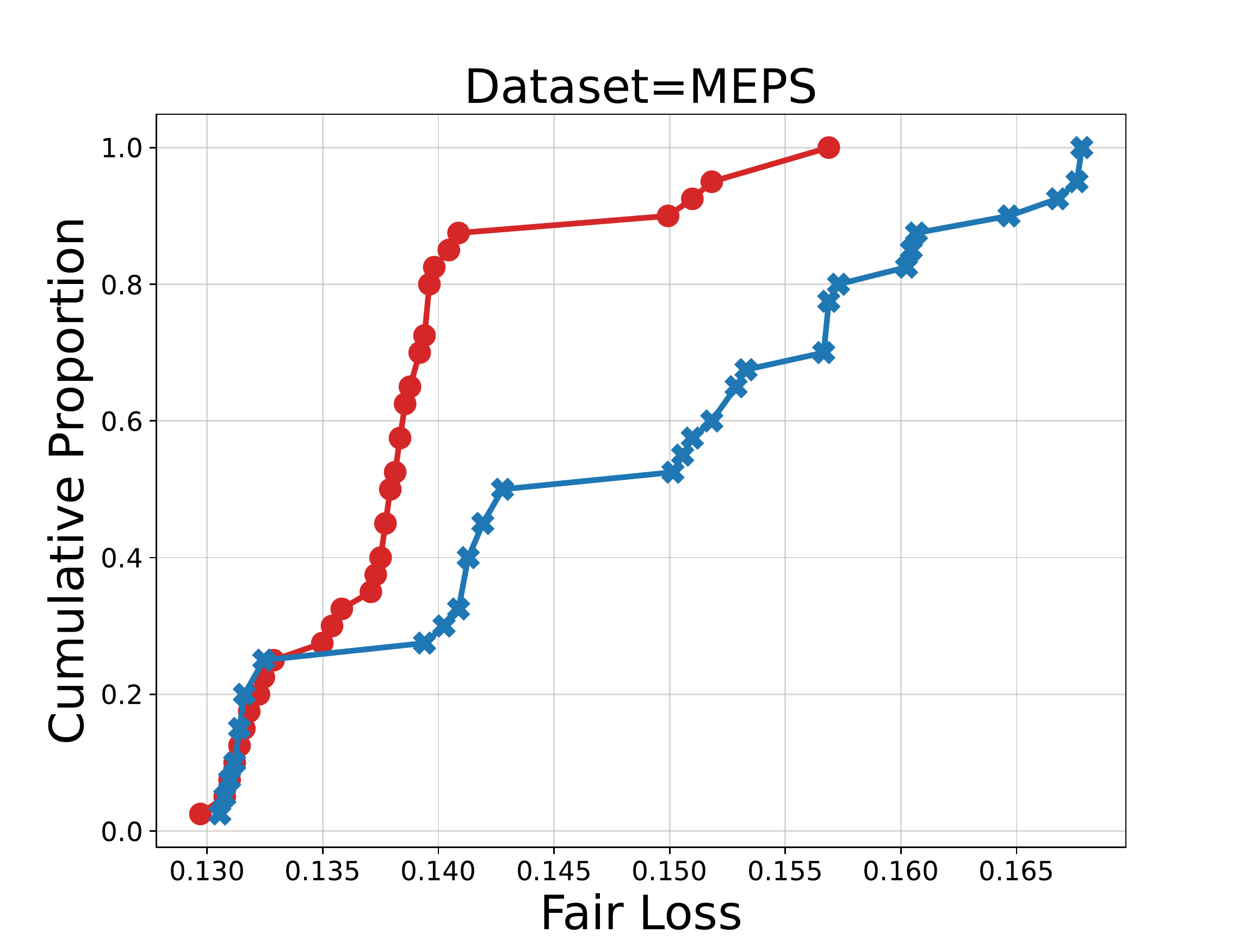}%
  \caption{Fair loss of different methods when tuning multiple learners on different datasts. Each sub-figure aggregates results from 40 experiments: (4 combinations of fairness settings) $\times$ (10 random seeds).}
  \label{fig:exp_res_cdf_aml}
  \end{figure*}
  
\begin{figure*}[ht]
    \centering
    \subfigure[Fair loss, resource consumption breakdown, and loss on the \emph{Bank} dataset]{
    \includegraphics[width=0.3\textwidth]{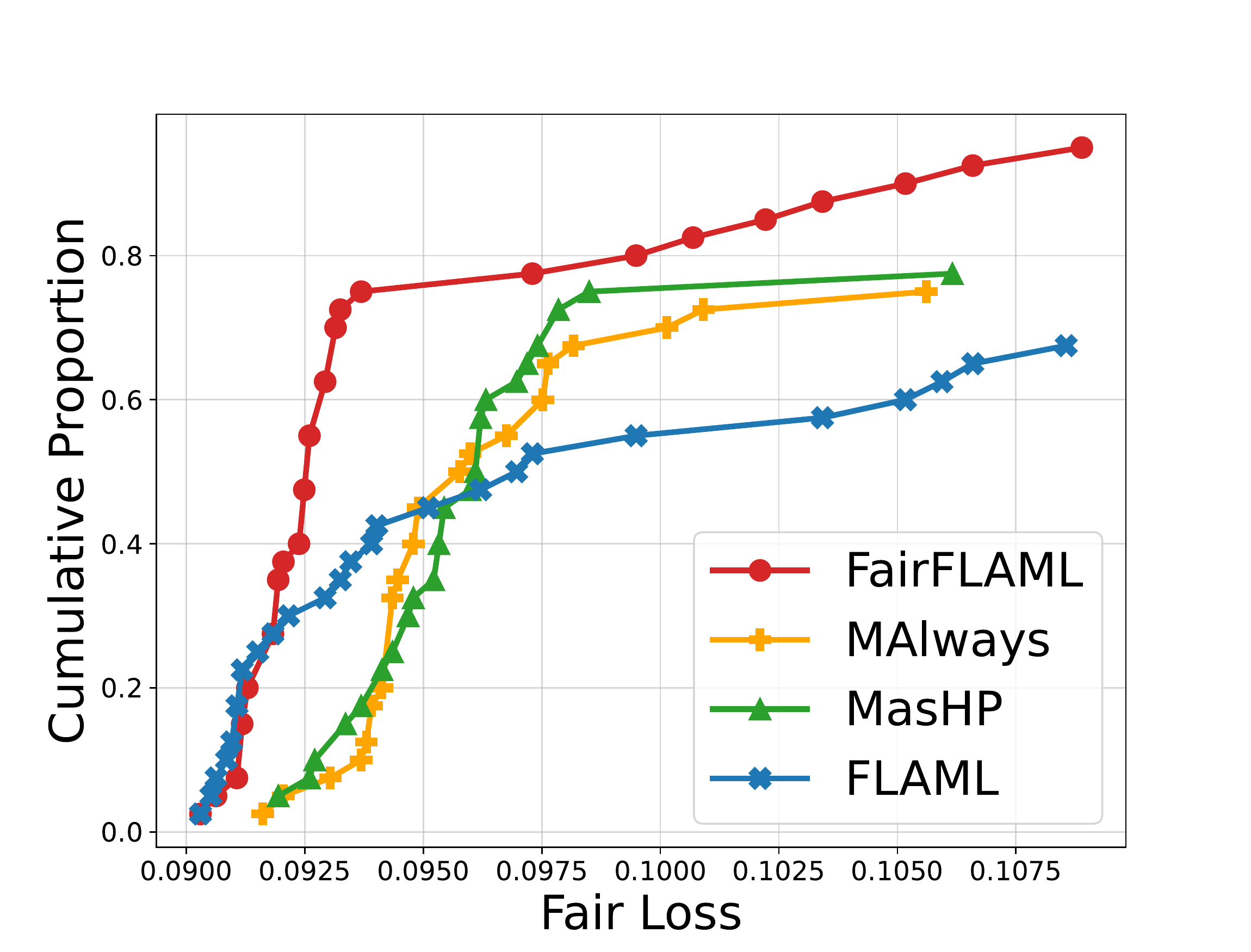}%
    \includegraphics[width=0.3\textwidth]{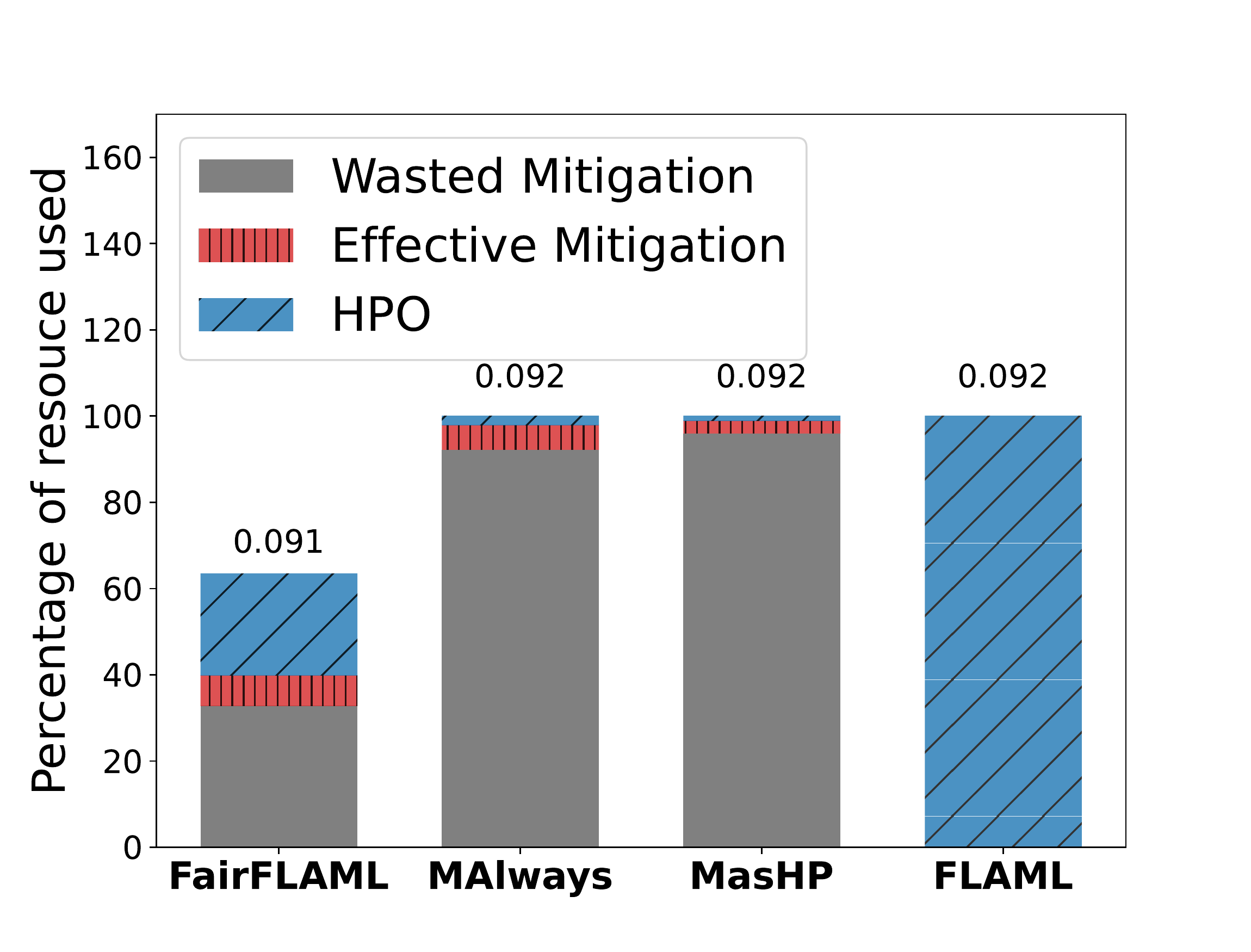}%
    \includegraphics[width=0.3\textwidth]{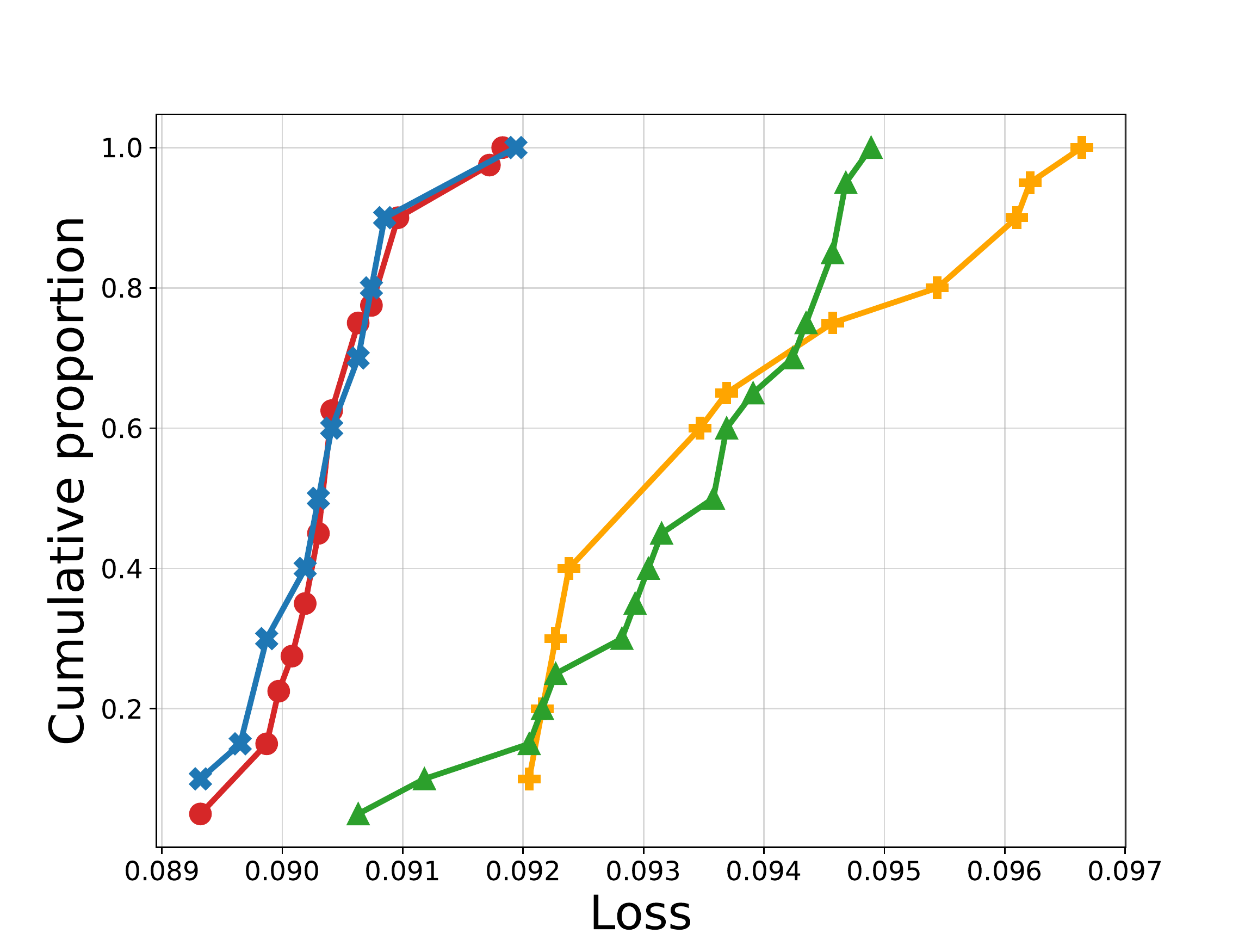}%
    } 
    \\
    \subfigure[Fair loss, resource consumption breakdown, and loss on the \emph{Compas} dataset]{
    \includegraphics[width=0.3\textwidth]{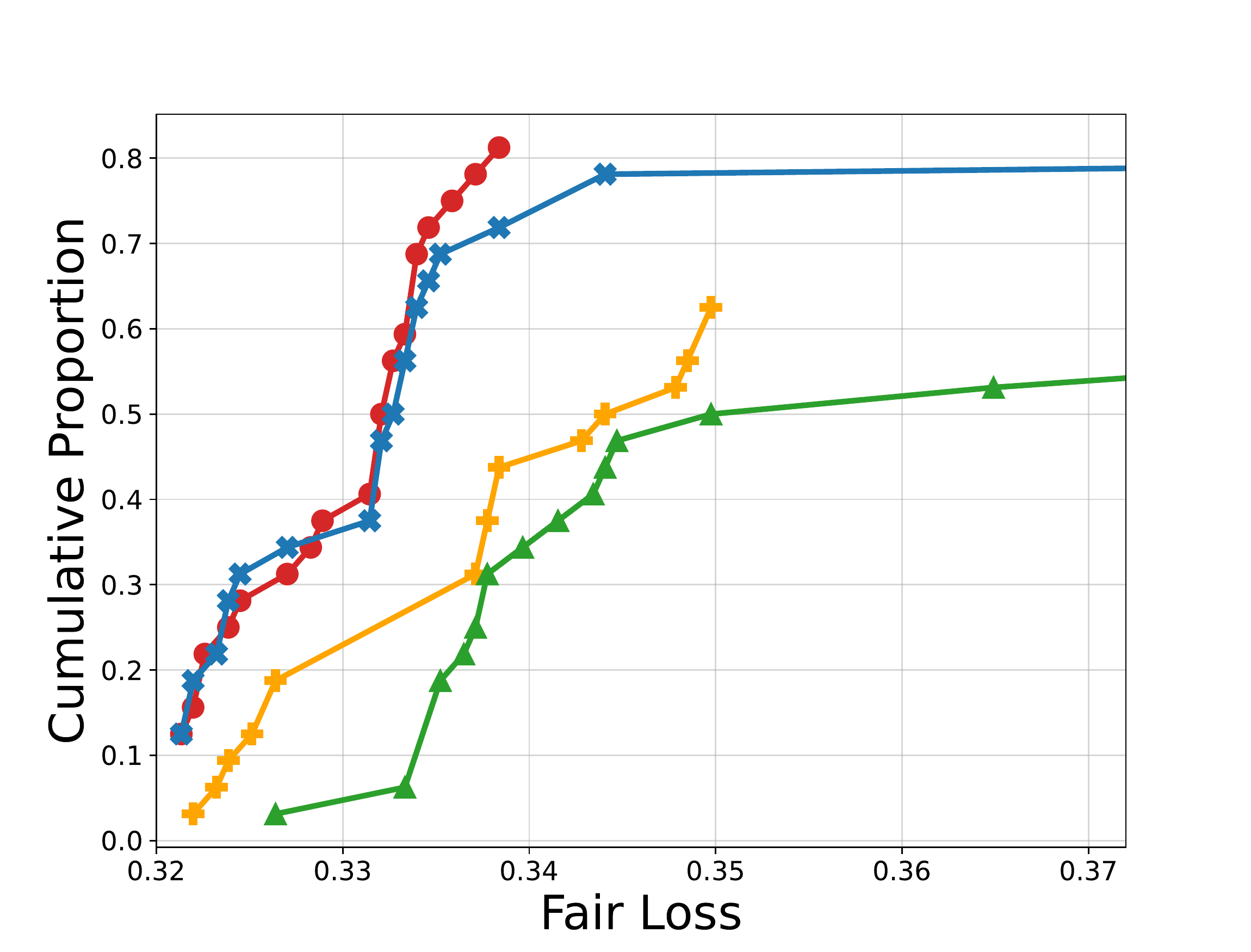}%
    \includegraphics[width=0.3\textwidth]{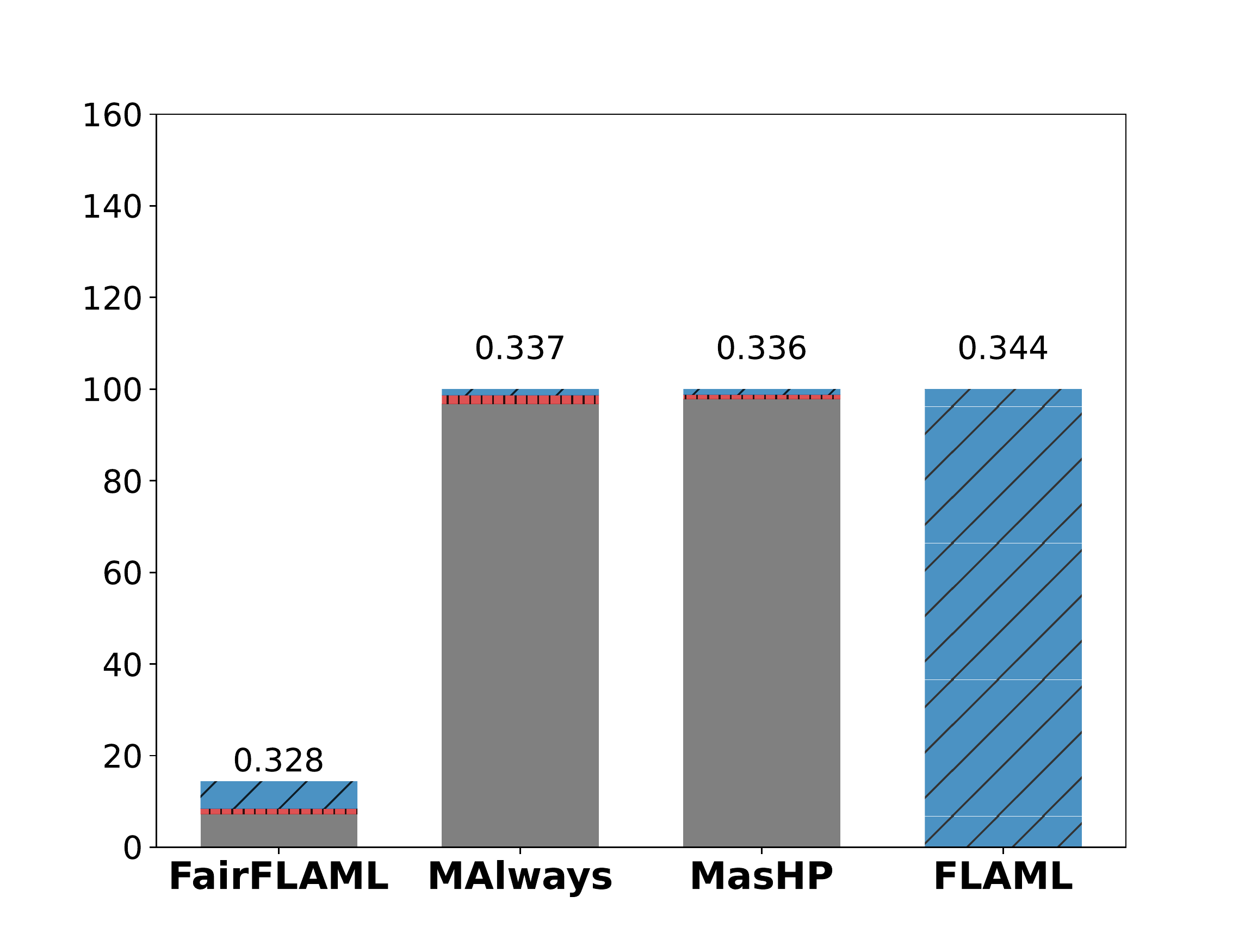}%
    \includegraphics[width=0.3\textwidth]{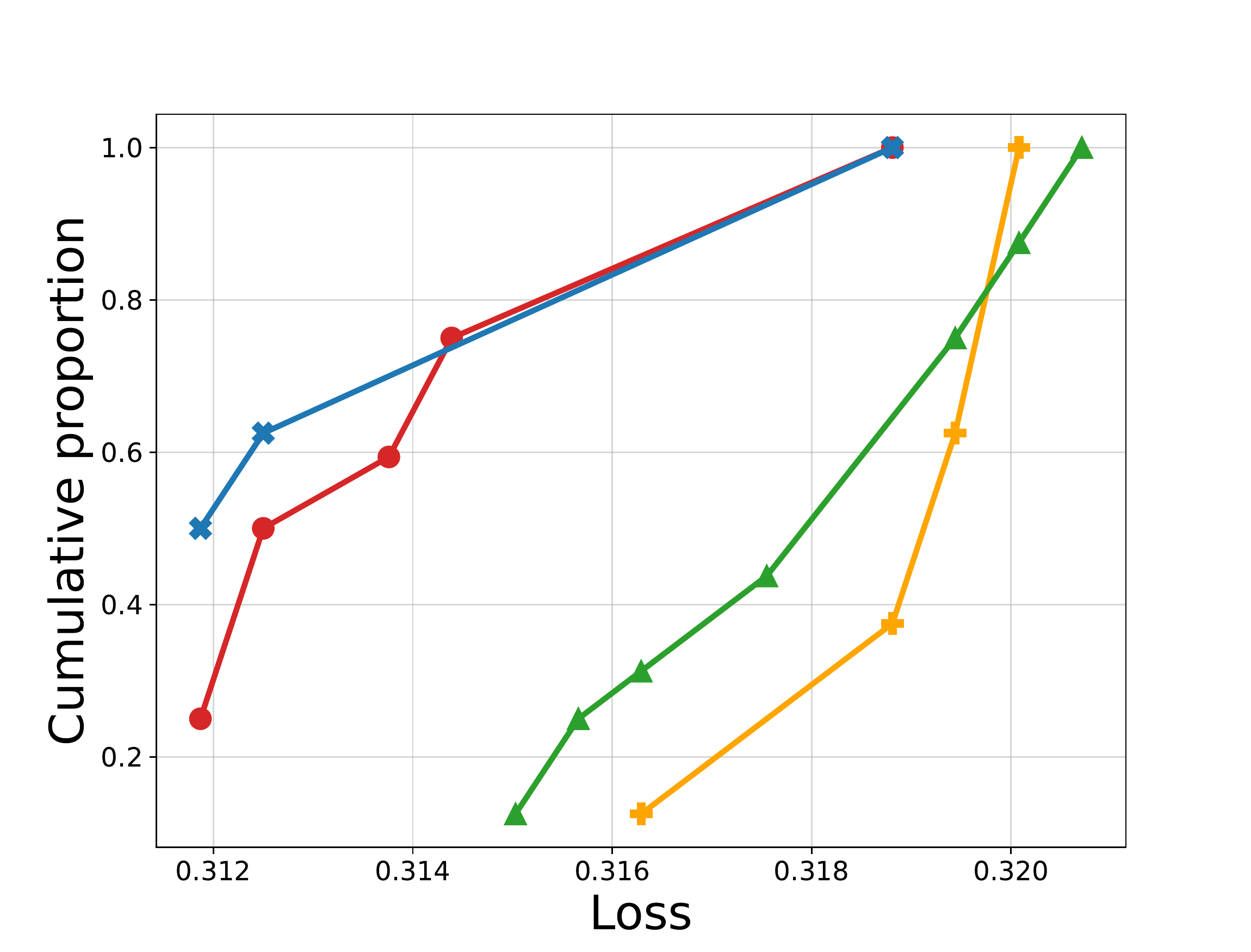}%
    } 
    \\
    \caption{Sub-figures in the 1st column and the 3rd column show fair loss and loss (without considering fairness constraints) on different datasets aggregated over 40 experiments similar to that in Figure~\ref{fig:exp_res_cdf_aml}. Sub-figures in the 2nd column show the detailed resource consumption breakdown. Each bar consists of resource consumption (the height of each bar indicates the resource consumption value) of three possible types of trials: trials where unfairness mitigation is not applied, denoted `HPO'; trials where unfairness mitigation is applied and the resulting model yields better fair loss (at that time point), labeled `Effective Mitigation' and trials where unfairness mitigation does not yield better fair loss, labeled `Wasted Mitigation'.
    The numbers above the bars are the fair losses achieved by the corresponding methods. } 
    \label{fig:exp_ablation_cdf}
    \end{figure*}
    
      \begin{figure*}
 \centering
  \includegraphics[width=0.25\textwidth]{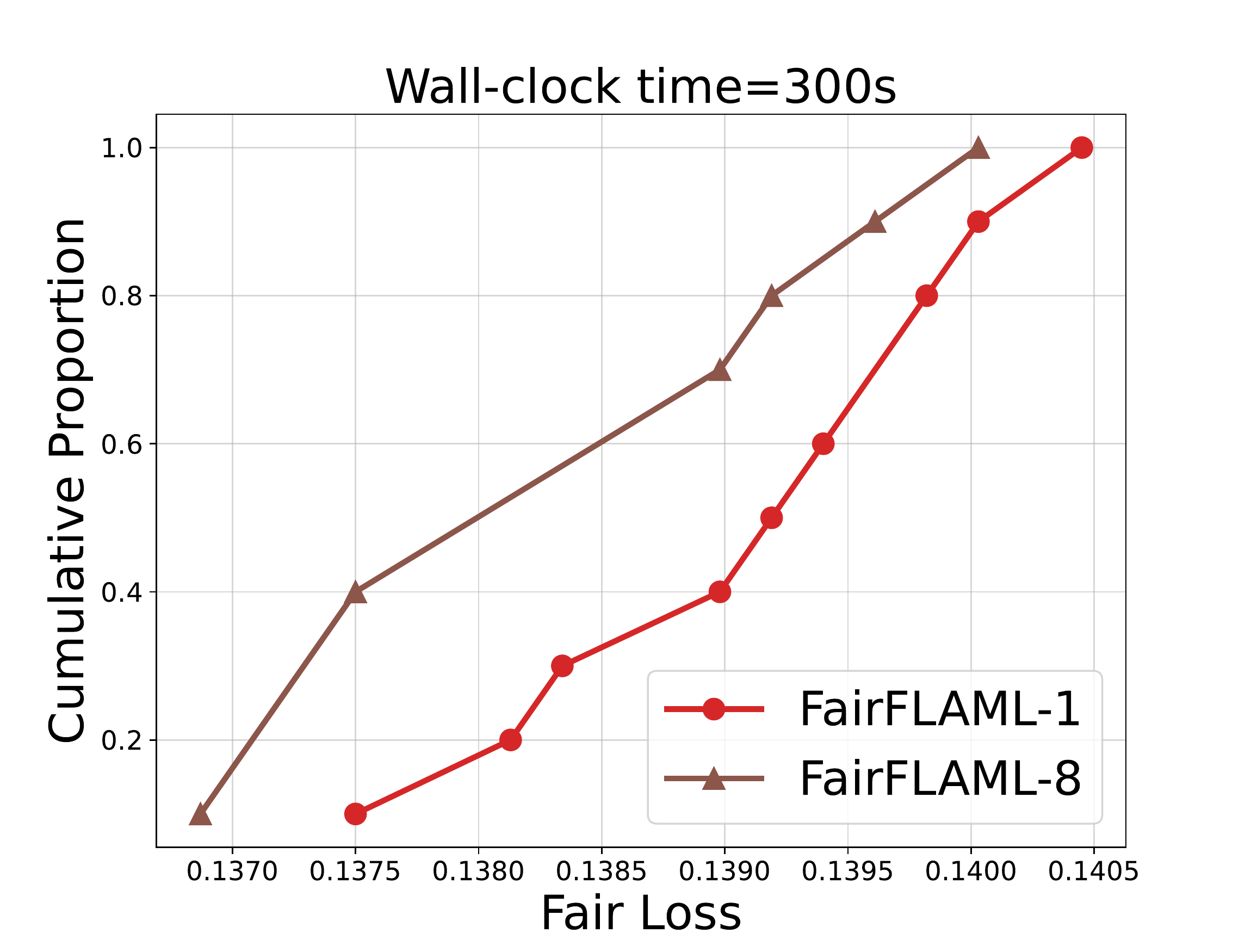}%
  \includegraphics[width=0.25\textwidth]{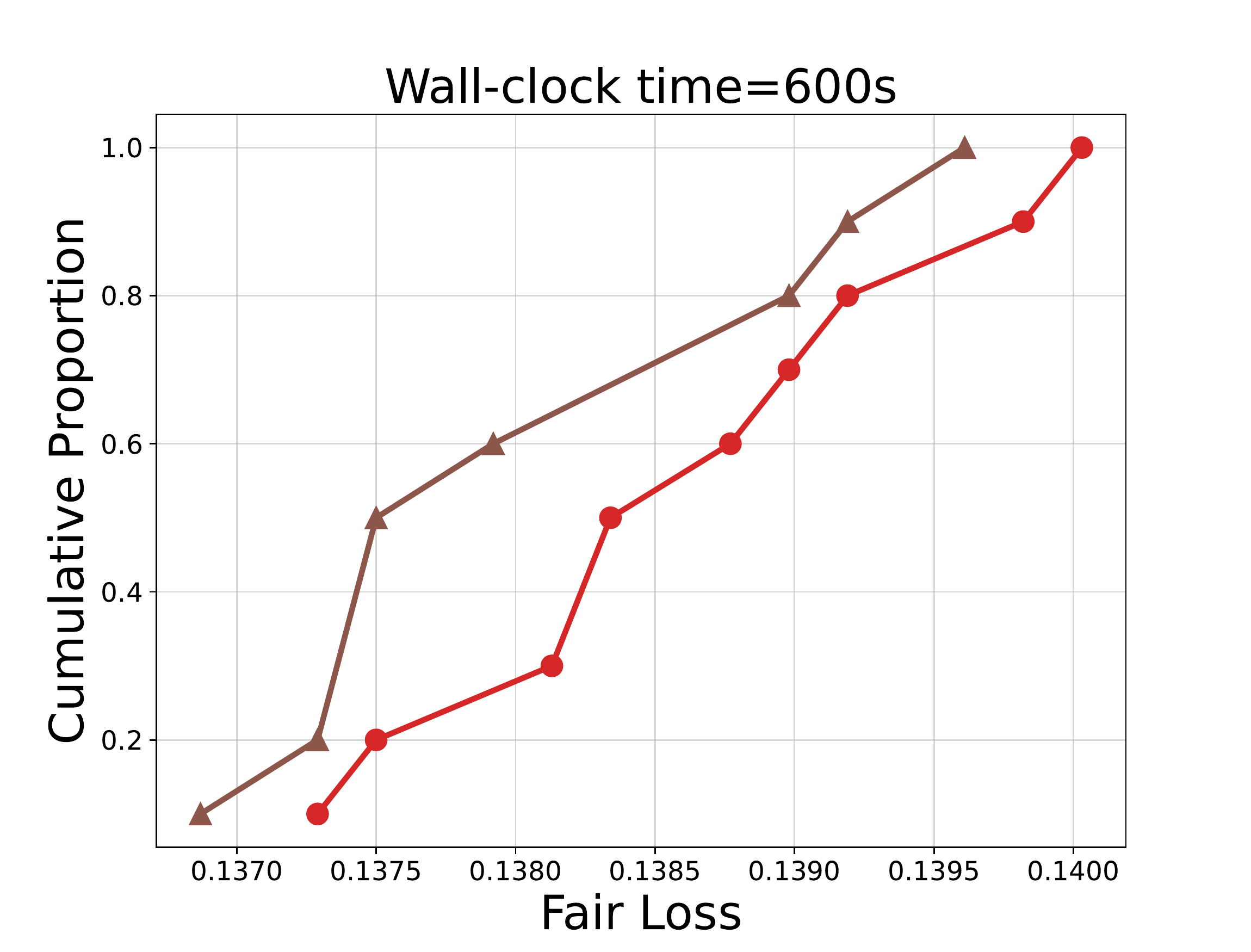}%
  \includegraphics[width=0.25\textwidth]{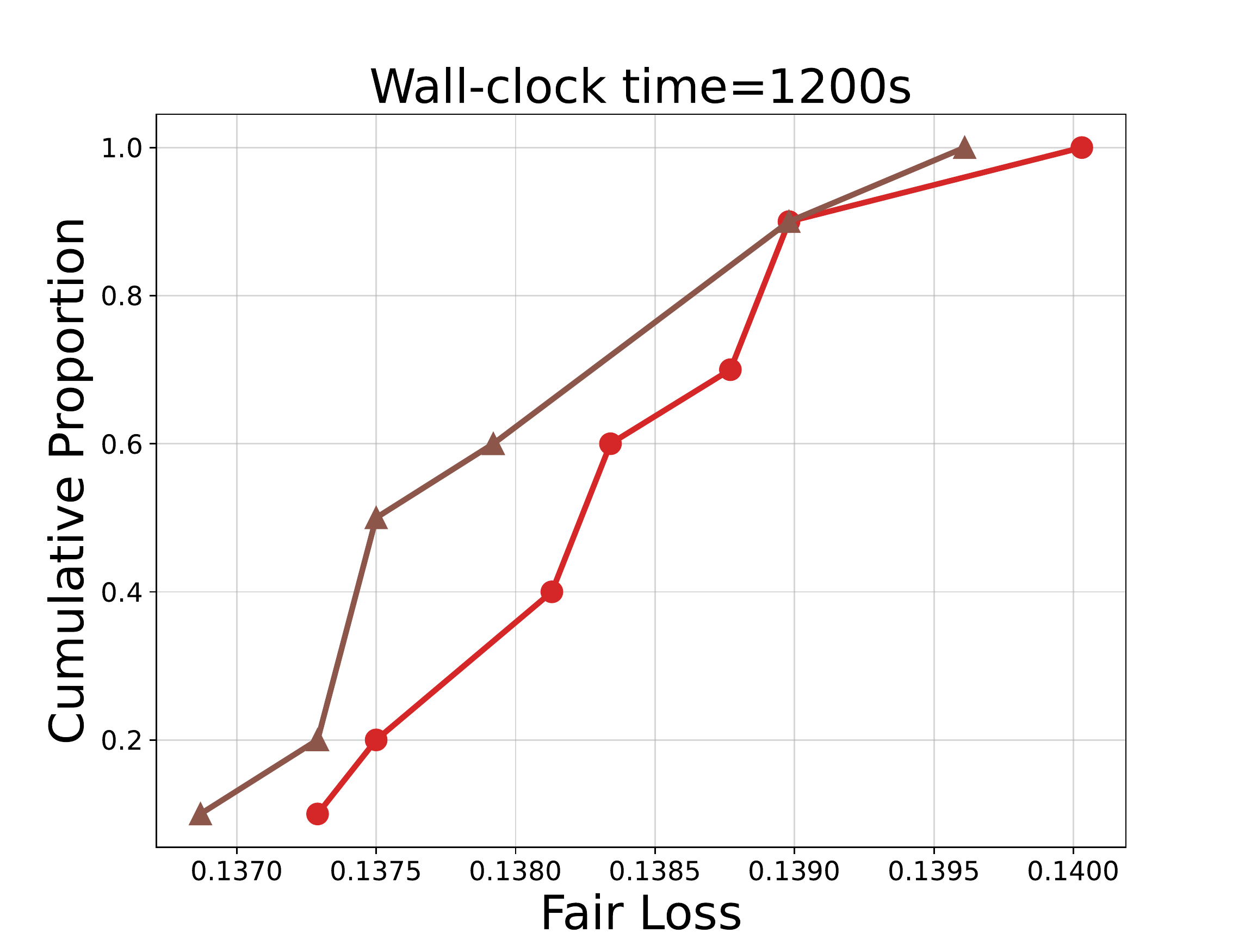}%
  \includegraphics[width=0.25\textwidth]{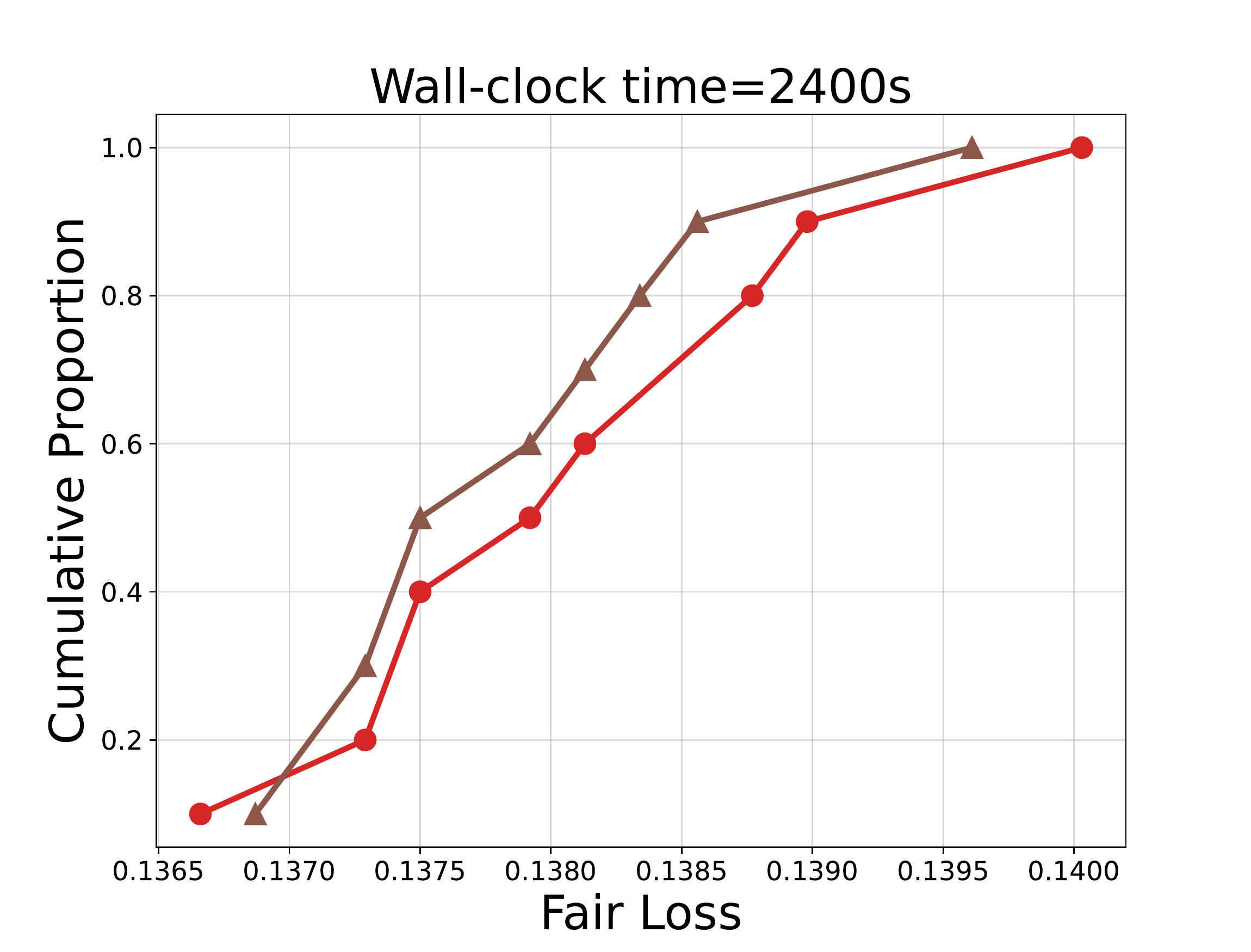}%
  \caption{Parallelization results under wall-clock time budgets on the \emph{MEPS} dataset. \fairFLAML-1 represents the variant of our method ran with a single core and \fairFLAML-8 with 8 cores.}
  \label{fig:exp_para_emps}
  \end{figure*}

\section{Experiments} \label{sec:exp}

\textbf{Datasets, baselines, and evaluation metrics.} We perform empirical evaluation on four machine learning fairness benchmark datasets, including \emph{Adult}~\cite{kohavi1996scaling}, \emph{Bank}~\cite{moro2014data}, \emph{Compas}~\cite{2016machinebias} and a Medical Expenditure Panel Survey dataset (\emph{MEPS} in short). All the datasets are publicly available and we provide more details about them in Appendix~\ref{appendix:detail}. We include FLAML and FairBO, which are reviewed in Section~\ref{sec:related_work}, as the baselines. We use $1 - \emph{validation accuracy}$ as the loss metric and fair loss as the final metric for evaluating the performance of the compared methods.

\textbf{Fair AutoML experiment settings.}
\textbf{(1) Fairness settings.}
We include two commonly used group fairness definitions for classification tasks, including Demographic Parity (DP) and Equalized Odds (EO). Following the group fairness evaluation settings in AIF360, we use `sex' as the sensitive attribute on \emph{Adult} and \emph{Compas}, `age' on \emph{Bank}, and `race' on \emph{MEPS}. We use two commonly used disparity thresholds  $\delta = 0.05$ and $0.01$ to determine whether the disparity is small enough to be fair. These two thresholds correspond to a mild
and a harsh fairness requirement respectively. 
We test two in-processing unfairness methods, including Exponentiated Gradient reduction, Grid Search reduction~\cite{agarwal2018reductions}, and one post-processing method~\cite{hardt2016equality}. We report the results for Exponentiated Gradient in the main paper and the other two in Appendix~\ref{appendix:exp}. We leverage existing implementations of the fairness metrics and unfairness mitigation methods from the open-source library Fairlearn. 
\textbf{(2) AutoML settings.}  
We do evaluations on two tuning tasks: (a) hyperparameter tuning on XGBoost; (b) hyperparameter tuning (including learner selection and model hyperparameter tuning) on multiple machine learners. All the other AutoML-related components, such as data pre-processing, the choice of search space, and the hyperparameter searcher, remain the same as FLAML's default options. Note that two different tuning methods~\cite{wu2021frugal,wang2021economical} are used in XGBoost and multi-learner tuning tasks, respectively, according to the default settings of FLAML. In all the presented results, we name our method \fairFLAML (instead of the general name FairAutoML) to reflect the fact that built based on the AutoML system FLAML. We run all the experiments up to 1 hour wall-clock time with 1 CPU core ten times with different random seeds if not otherwise noted. 
(3) \textbf{Parallelization}. We test the performance of our method under both sequential and parallel computation resources and include the results in the third subsection.

\subsection{Effectiveness}
We first want to verify the effectiveness of our method in terms of producing models with good fair loss. We summarize the results from \fairFLAML and the baselines when tuning XGBoost in Figure~\ref{fig:exp_res_cdf}, and when tuning multiple learners in in Figure~\ref{fig:exp_res_cdf_aml}. In the latter experiment, we do not include FairBO as it does not support tuning multiple learners simultaneously. 

In the sub-figures of Figure~\ref{fig:exp_res_cdf}, we summarize the fair loss of different methods under different fairness settings. In the sub-figures of Figure~\ref{fig:exp_res_cdf_aml}, we show results aggregated over all combinations of fairness settings and random seeds on different datasets. 

We first observe that, on the \emph{Compas} dataset, it is indeed possible to produce ML models with good fair loss by simply tuning regular hyperparameters in certain cases: both FLAML and \fairFLAML achieve the best fair loss when tuning multiple learners as shown in the 3rd sub-figure of Figure~\ref{fig:exp_res_cdf_aml}. However, in all the other cases, \fairFLAML achieves significantly better fair loss than both FLAML and FairBO (when tuning XGBoost), especially when the fairness requirement is harsh. For example, when tuning XGBoost on the \emph{MEPS} dataset (Figure~\ref{fig:exp_ablation_cdf}(d)), when threshold = 0.05, FLAML's performance is close to \fairFLAML. When threshold = 0.01, FLAML's performance can become very bad and sometimes it cannot even find a single model satisfying the fairness constraint. This again verifies the huge benefit of introducing unfairness mitigation to AutoML. Regarding the baselines, we also observe that the performance of FairBO is much worse than both \fairFLAML and FLAML. There are mainly two reasons for its bad performance: (1) Similar to the case of FLAML, FairBO is also tuning regular model hyperparameters to find fair models with good loss. Using a fairness-constrained optimization method does not reconcile the ineffectiveness of hyperparameter tuning in building a model with good fair loss; (2) Unlike FLAML, the Bayesian optimization method used in FairBO does not consider the different computation cost of models with different complexities, and thus may try unnecessarily costly models, which makes the method, in general, more time-consuming than FLAML.   

In summary, the comprehensive evaluations verify the necessity of introducing unfairness mitigation to AutoML and the effectiveness of our method.

\subsection{Ablation study and efficiency}

\textbf{Alternative ways to include unfairness mitigation.} We further compare \fairFLAML with two alternative approaches to incorporate unfairness mitigation under the general fair autoML framework proposed in Alg.~\ref{alg:fair_automl_general}:
(1) As mentioned in Section~\ref{sec:formulation_ana}, one straightforward approach to include unfairness mitigation is to always enforce it during model training after each hyperparameter configuration is proposed.  We name this approach \MAlways. 
(2) Another alternative is to treat unfairness mitigation as an additional categorical `hyperparameter' taking binary values (1/0 for enabling/disabling unfairness mitigation) and apply constrained hyperparameter optimization~\cite{FLAML_constrained}.  We name this method \MasHP. Both these approaches can be realized via the \FairnessManager abstraction and can be considered as an ablation of \fairFLAML.  By comparing with these two alternative approaches, we show the necessity of the self-adaptive mitigation strategy in \fairFLAML.

We provide result comparisons with $\MAlways$ , $\MasHP$, and FLAML in on the \emph{Bank} and \emph{Compas} Figure~\ref{fig:exp_ablation_cdf}. Due to page limit, we include the results on the other two datasets in Figure~\ref{fig:exp_ablation_cdf_append} of the appendix.
\begin{enumerate}
\vspace{-1mm}
    \setlength\itemsep{-0.2em}
    \item In the 1st column of Figure~\ref{fig:exp_ablation_cdf}, we show the fair loss of \fairFLAML and its two alternatives under a small resource budget (300s).  The results demonstrate \fairFLAML's significant advantage over the two alternatives regarding the fair loss. 
    \item  To better understand how resource is used in different methods, in the 2nd column of  Figure~\ref{fig:exp_ablation_cdf}, we visualize the resource breakdown for different methods to achieve the best fair loss under a particular experiment setting with one random seed.  (1) We first observe that in \MAlways and \MasHP,  `Wasted Mitigation' dominates the resource consumption, which is an important source of their inefficiency. 
     (2) FLAML can be efficient in finding fair models with the best fair loss when it is possible to do so via regular hyperparameter search as shown on the \emph{Compas} dataset. However, the effectiveness of this method is highly dataset-dependent. 
     (3) \fairFLAML can achieve on average better performance with much less resource than the two alternatives according to the fair loss values and the heights of the resource consumption bars. \fairFLAML is able to realize the benefit of both hyperparameter search and unfairness mitigation. 
    \item We believe it is also meaningful to investigate the original loss without considering fairness constrains, we show the results in the 3rd column of Figure~\ref{fig:exp_ablation_cdf}. We observe that \fairFLAML preserves FLAML's good performance regarding the original loss while having the best fair loss. This property is especially desirable when the practitioners are in an explorative mode regarding machine learning fairness, which is quite common due to the under-development of this topic in practical scenarios. It alleviates potential hesitations of adopting unfairness mitigation to AutoML due to concerns about a degraded original loss. By knowing the best original loss achievable without considering fairness constraints, practitioners can gain more confidence in their decision-making. \fairFLAML is the only system that can find both the best original loss and the best fair loss.
\end{enumerate}

\subsection{Parallelization}
Our method is easy to parallelize. We include the results of our method run with a single core and multiple cores under different wall-clock time budgets on the \emph{MEPS} dataset, which is the largest dataset considering both the number of data instances and dimensionality in Figure~\ref{fig:exp_para_emps}. The results show that by increasing the parallel computation resources, our method can achieve even better results under the same wall-clock time budget, which is helpful in reducing the turn-around time of the tuning task in time-sensitive tasks. We also observe that when wall-clock time budgets increase, the difference between the parallel variant and sequential variant decreases.

\subsection{Additional evaluation results and extensions}
Due to space limit, we include additional evaluation results in Appendix~\ref{appendix:exp}. (1) \textbf{An additional baseline:} In addition to the two alternative approaches in our ablation study, we compare our method with a third alternative. This alternative is a post hoc approach in which we select the configurations with the best loss to perform unfairness mitigation after a regular AutoML process is finished. We include the detailed descriptions about these three alternatives, and comparisons with the post hoc alternative under different experiment settings in Figure~\ref{fig:postflaml} of Appendix~\ref{appendix:exp}. (2) \textbf{Results on different types of unfairness mitigation methods:} Our method is compatible with most of the state-of-the-art unfairness mitigation methods. We include additional experiments under different types of unfairness mitigation methods in Figure~\ref{fig:exp_res_boxplot_grid}  and Figure~\ref{fig:exp_res_boxplot_post} of Appendix~\ref{appendix:exp}.
(3) \textbf{Extensibility:} Our proposed framework and system are highly extensible and are compatible with a wide range of hyperparameter searchers, fairness definitions, and unfairness mitigation methods. We include instructions and code examples to verify the wide compatibility in Appendix~\ref{appendix:extension}.

\section{Summary and future work}
In this work, we first identify that it is beneficial and sometimes necessary to introduce unfairness mitigation to the AutoML workflow to help improve the fairness of the overall system. We propose a general framework for incorporating unfairness mitigation as an organic part of the AutoML process and present a fair AutoML system that is flexible, robust, and efficient. 


\clearpage
\bibliography{ref_fairaml}
\bibliographystyle{mlsys2023}

\appendix
\clearpage
\section{Omitted details} \label{appendix:detail}

\textbf{Formal definition and desirable properties of \ECF.} 
\begin{align} \label{eq:ecf}
    \ECF^{(m)} \coloneqq \max\{ & R_{(m)} - R_{(m)}^{\text{1st}}, R_{(m)}^{\text{1st}} - R_{(m)}^{\text{2nd}}, \\ \nonumber
    & 2 \frac{L_{\text{fair}, (m)}^{\text{1st}}- L_{\text{fair}}^{*}}{s_{(m)}}\} 
\end{align}
in which the superscript or subscript $(m)$ denotes the variant of the variable associated with unfairness mitigation ($m=1$) or hyperparameter tuning ($m=0$). $R_{(m)}$ is the total resource spent, $R_{(m)}^{\text{1st}}$ is the total resource spent when the model with the best fair loss is first obtained and $L_{\text{fair}, (m)}^{1st}$ is the corresponding best fair loss obtained, $R_F^{\text{2nd}}$ is the total resource spent when the model with the second best fair loss is first obtained, $s_{(m)}$ is the speed of fair loss improvement, and $L_{\text{fair}}^*$ is the overall best fair loss obtained in the system.

We mentioned the nice self-adaptive property of the decision-making strategy based on \ECF. It is worth mentioning that this property holds even if \ECF is not an accurate estimation of its ground-truth counterpart: In the case when $\ECF^{(1)}$ and/or $\ECF^{(0)}$ are not accurate estimation of their ground-truth counterparts and a wrong choice is made, the consequence of this wrong choice will be reflected in the $\ECF$  of the selected choice (it will become larger), while the $\ECF$ of the other choice (the second choice) remains unchanged. Thus we will turn to the second choice. 

\textbf{Formal definition of $ \zeta_{\cH_t, B}$.}
Considering the general loss degradation of unfairness mitigation, and the the error analysis provided in~\cite{agarwal2018reductions}, we approximate the loss degradation with the following formula. 
\begin{align} \label{eq:est_fair_loss}
   \zeta_{\cH_t, B} =
    \begin{cases}
      \eta &  \text{ECI} < \tau_{c}, \text{ECI} + \tau_{c} < B\\
       \eta - b & \text{Otherwise}
    \end{cases}       
\end{align}
in which $\eta$ and $b$ are the average and 95\% confidence radius of loss degradation after doing mitigation according to the historical observations respectively; $\tau_c$ is the projected resource needed to do the mitigation estimated based on historical observations; and $\text{ECI}$ is originally from the employed AutoML system FLAML~\cite{wang2021flaml}, and is the estimated cost for achieving loss improvement. 
We calculated the projected resource needed for performing a successful unfairness mitigation on hyperparameter $c$, i.e., $\tau_c$ according to:
\begin{align}
    \tau_c \coloneqq \frac{\kappa^{(0)}_{c}}{q_{\text{mitigation}}} \cdot \frac{1}{|\cH^{(1)}|} \sum_{c' \in \cH^{(1)}} \frac{\kappa^{(1)}_{c'} }{\kappa^{(0)}_{c'}}
\end{align}
in which $|\cH^{(1)}_F|$ is the length of $\cH^{(1)}$ (i.e., the total number of configurations on which we performed unfairness mitigation), and $q_{\text{mitigation}} \coloneqq \frac{ | \{ c' \in \cH^{(1)} | \Fair_{c'} = 1 \}|  }{|\cH^{(1)}| }$ is the success rate of performing unfairness mitigation. Recall that $\kappa^{(1)}_{c'}$ and $\kappa^{(0)}_{c'}$ are the actual resource used for performing model training w/ and w/o unfairness mitigation based on hyperparameter $c'$ respectively. The factor $\frac{ \kappa^{(1)}_{c'} }{\kappa^{(0)}_{c'}}$ is the `computation cost ratio' visualized in the last column of Figure~\ref{fig:mitigation_analysis_new}, and thus $\kappa^{(0)}_{c} \frac{1}{|\cH^{(1)}|} \sum_{c' \in \cH^{(1)}} \frac{ \kappa^{(1)}_{c'} }{\kappa^{(0)}_{c'}}$ is the expected resource consumption for applying mitigation to configuration $c$. We further penalize it 
by $q_{\text{mitigation}}$ to get an estimation of the resource needed for a successful mitigation.

\textbf{Datasets.}
The four datasets used in this paper are all publicly available and representative for three fairness-sensitive machine learning applications: financial resource allocation, business marketing and criminal sentencing. 
The \emph{Adult} dataset is a census dataset, the original prediction task of which is to determine whether a person makes over 50K a year. 
The \emph{Bank} dataset is a classification dataset, the goal of which is to predict if the client will subscribe a term deposit.
The \emph{Compas} dataset is a classification dataset used to predict whether a criminal defendant will re-offend. We provide detailed statistics about these three datasets in Table~\ref{tab:datasets}.
\begin{table}[th!]
\caption{Dataset statistics.}
\begin{center} \label{tab:datasets}
\small
\begin{tabular}{ |c|c|c|c|c|}  
 \hline
   & Adult & Bank & Compas & MEPS  \\  \hline
\# of instance & 48842 & 45211 &  5278 & 15830 \\ \hline
\# of attributes & 18 & 16 & 10 & 138 \\\hline
 Area & finance & marketing & crime & medical\\ \hline
\end{tabular}
\end{center}
\end{table}

Please refer to the following links to access the three datasets tested.
\begin{itemize}
    \item \emph{Adult}.
    Description in AIF360: \url{https://github.com/Trusted-AI/AIF360/tree/master/aif360/data/raw/adult}
    
    \item  \emph{Bank}.  
    Description in AIF360: \url{https://github.com/Trusted-AI/AIF360/tree/master/aif360/data/raw/bank}
    \item \emph{Compas}. 
    Description in AIF360: \url{https://github.com/Trusted-AI/AIF360/tree/master/aif360/data/raw/compas}
    \item \emph{MEPS}. 
    Description in AIF360: \url{https://github.com/Trusted-AI/AIF360/tree/master/aif360/data/raw/meps}
\end{itemize}

\clearpage
\section{Additional evaluation results}\label{appendix:exp}

\begin{figure*}[ht]
    \centering
    \subfigure[Fair loss under low resource budget, resource consumption breakdown, and loss on the \emph{Adult} dataset]{
    \includegraphics[width=0.33\textwidth]{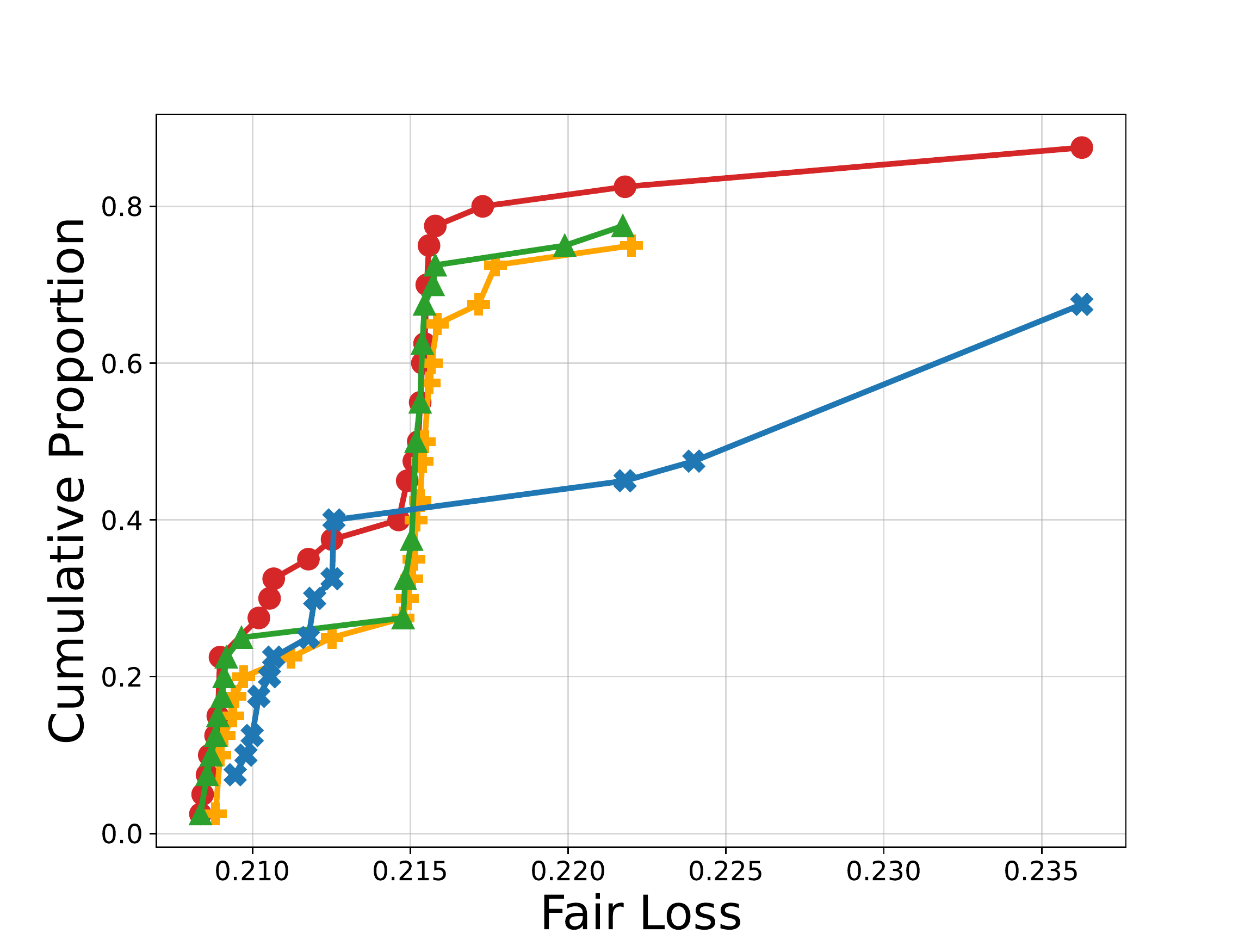}%
    \includegraphics[width=0.33\textwidth]{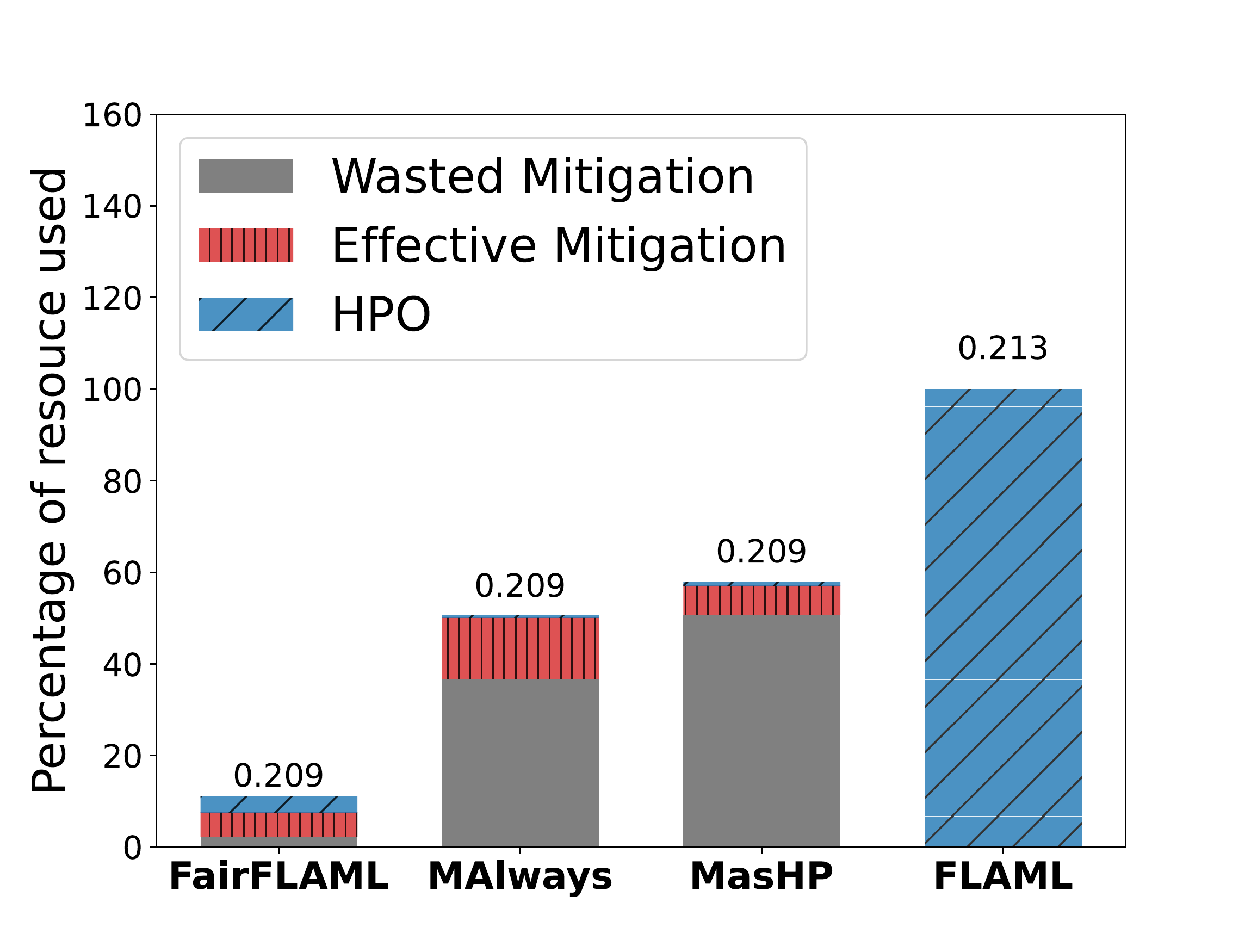}%
     \includegraphics[width=0.33\textwidth]{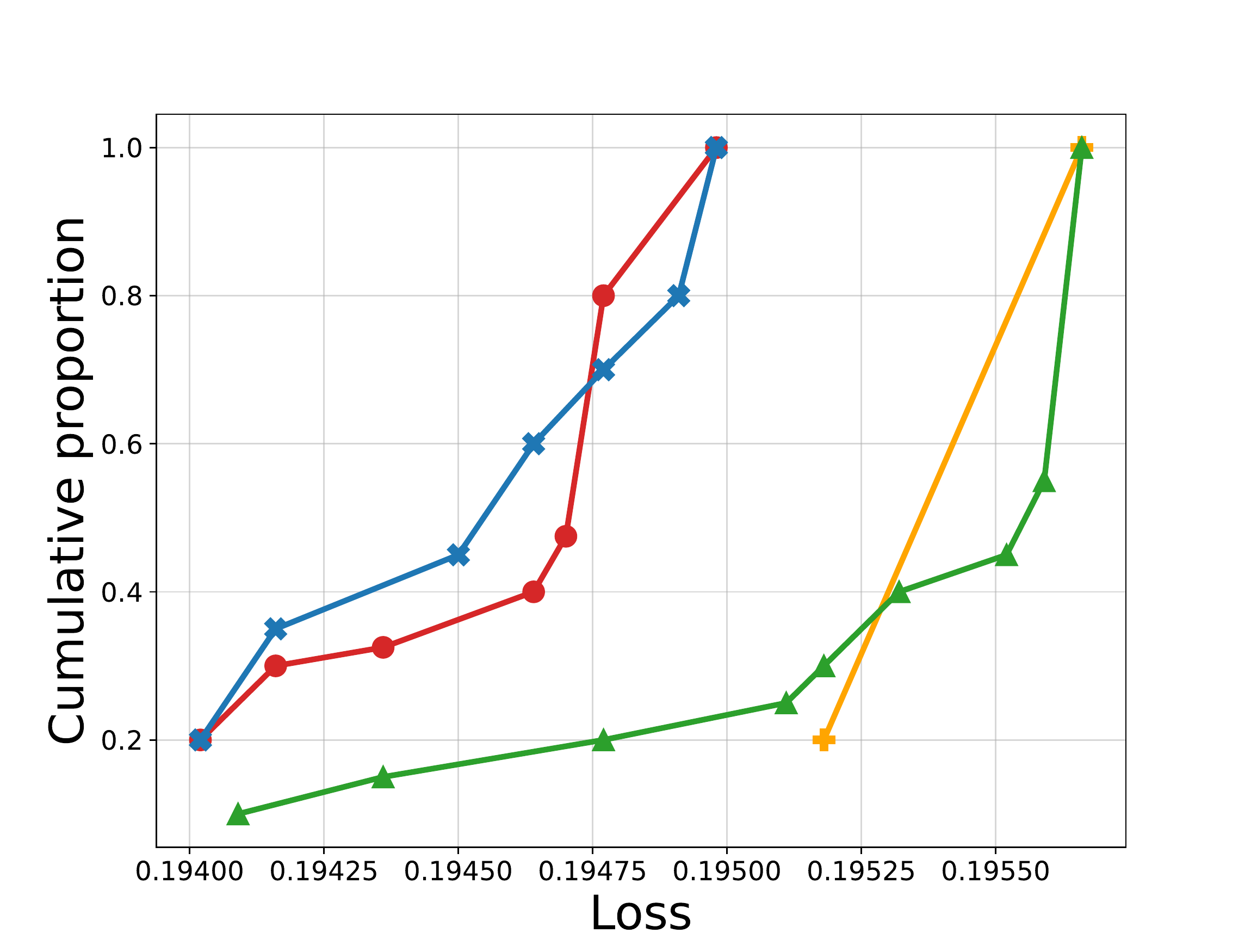}%
        }
    \\
    \subfigure[Fair loss, resource consumption breakdown, and loss on the \emph{MEPS} dataset]{
    \includegraphics[width=0.33\textwidth]{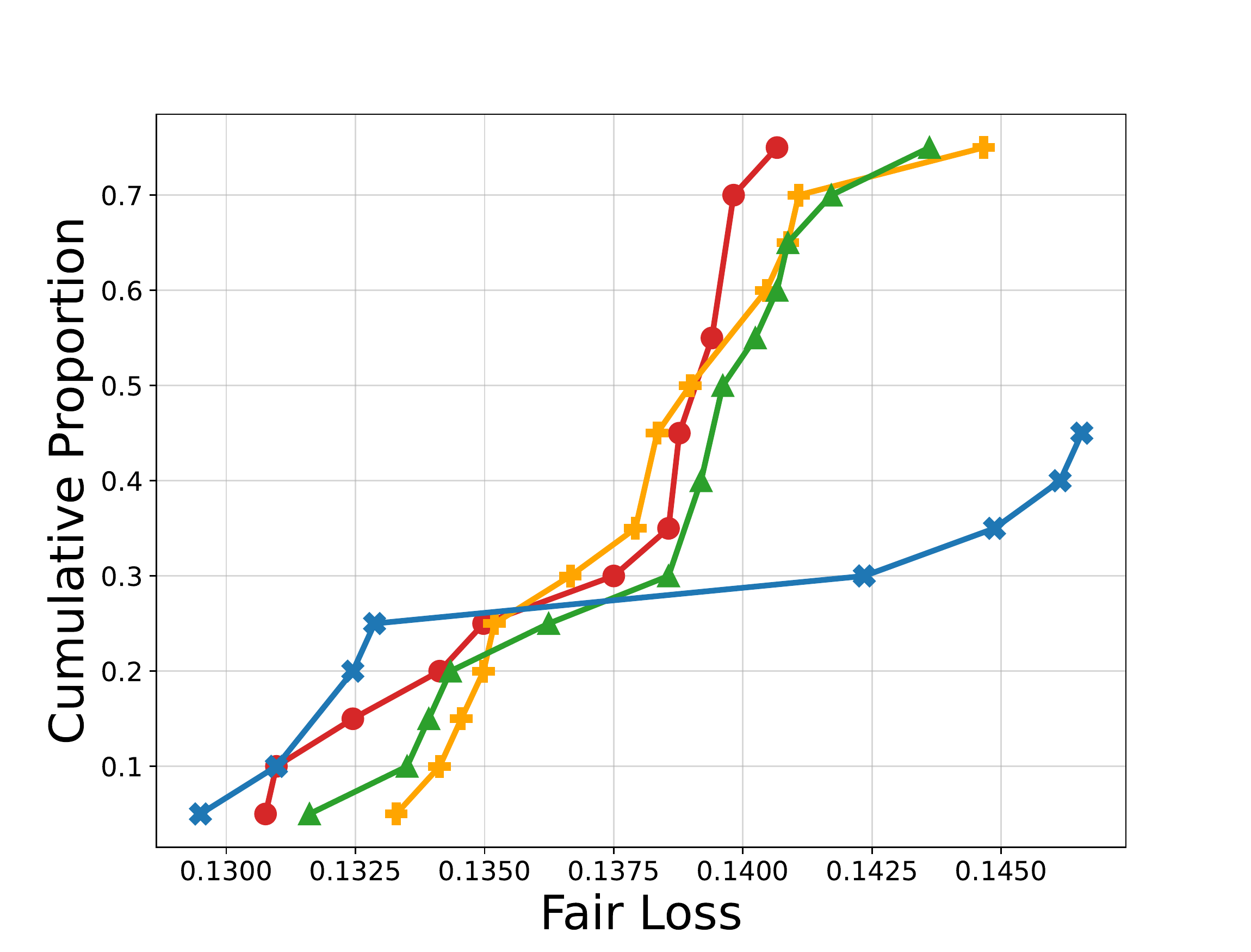}%
    \includegraphics[width=0.33\textwidth]{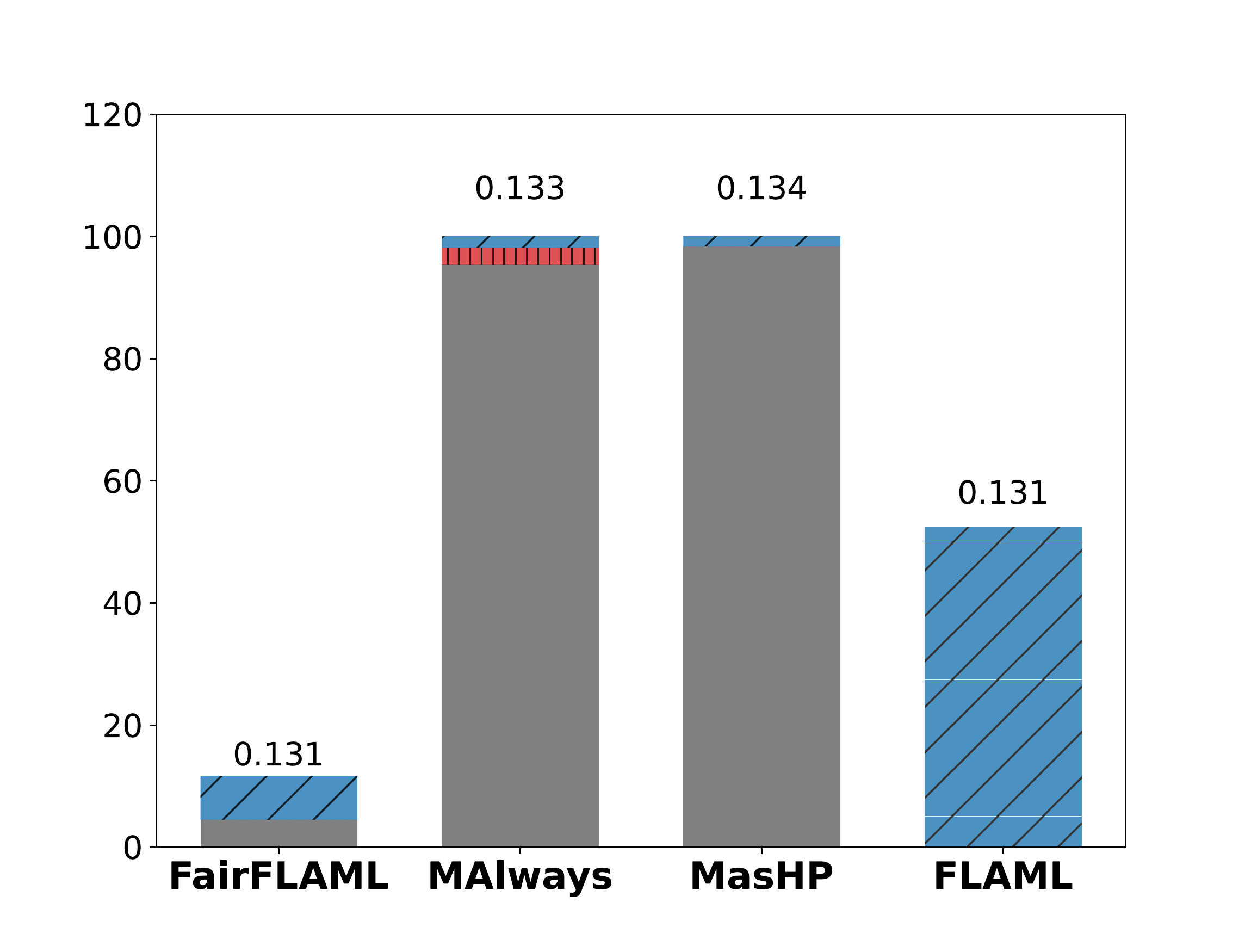}%
    \includegraphics[width=0.33\textwidth]{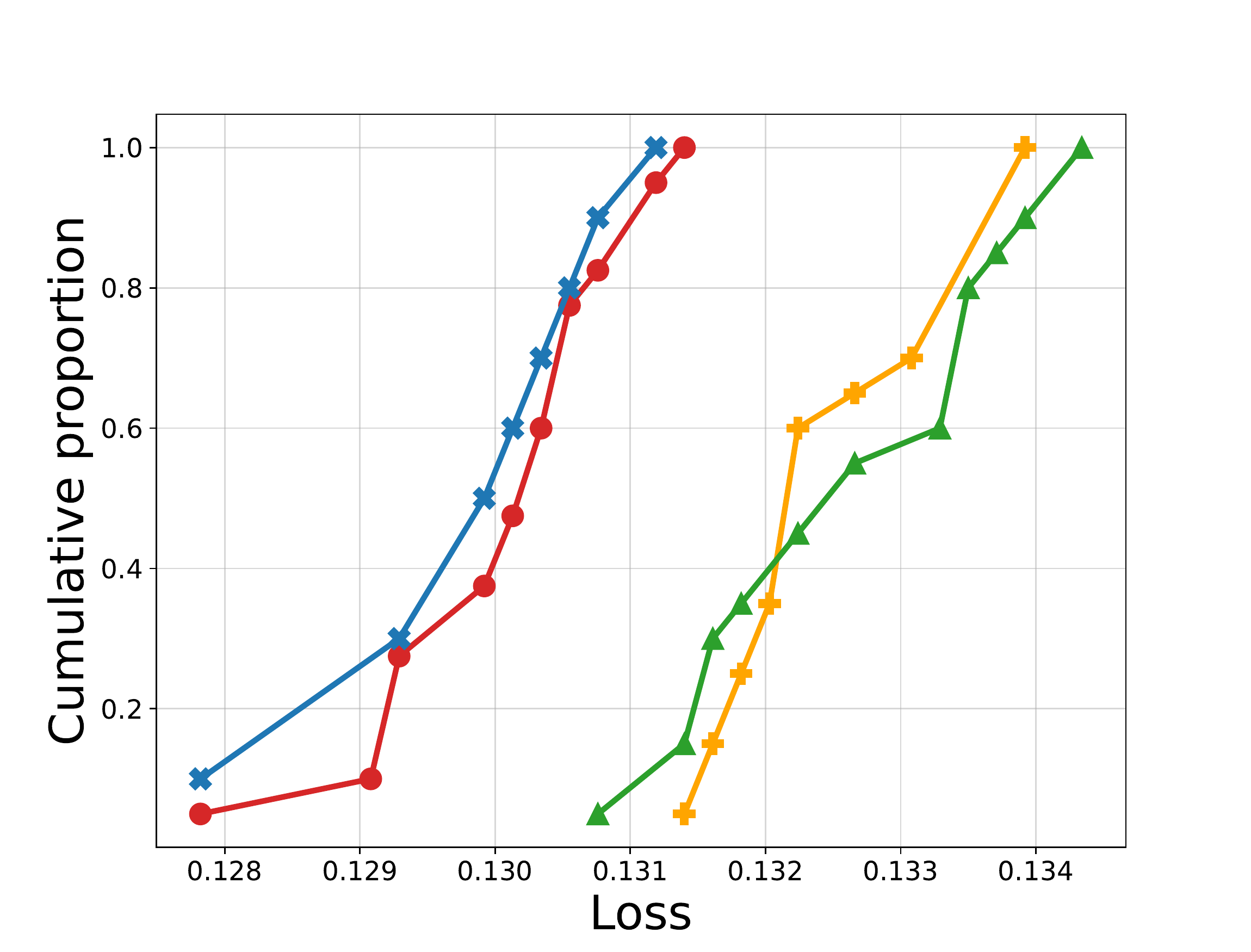}%
    } 
    \caption{Fair loss, resource consumption breakdown, and loss without considering fairness constraints on different datasets. }
    \label{fig:exp_ablation_cdf_append}
    \end{figure*}

\subsection{Comparison with a post-hoc approach}
In this section, we compare \fairFLAML with another straightforward approach which is inspired by the analysis in Section~\ref{sec:formulation_ana}. This approach primarily treats unfairness mitigation as a disjoint component post to AutoML, i.e., one first finishes the AutoML task and then performs unfairness mitigation on the configurations tried during AutoML in ascending order of the loss.

\begin{figure*}[th]
    \centering
    \subfigure[Wall-clock-time = 300s]{
    \includegraphics[width=0.25\textwidth]{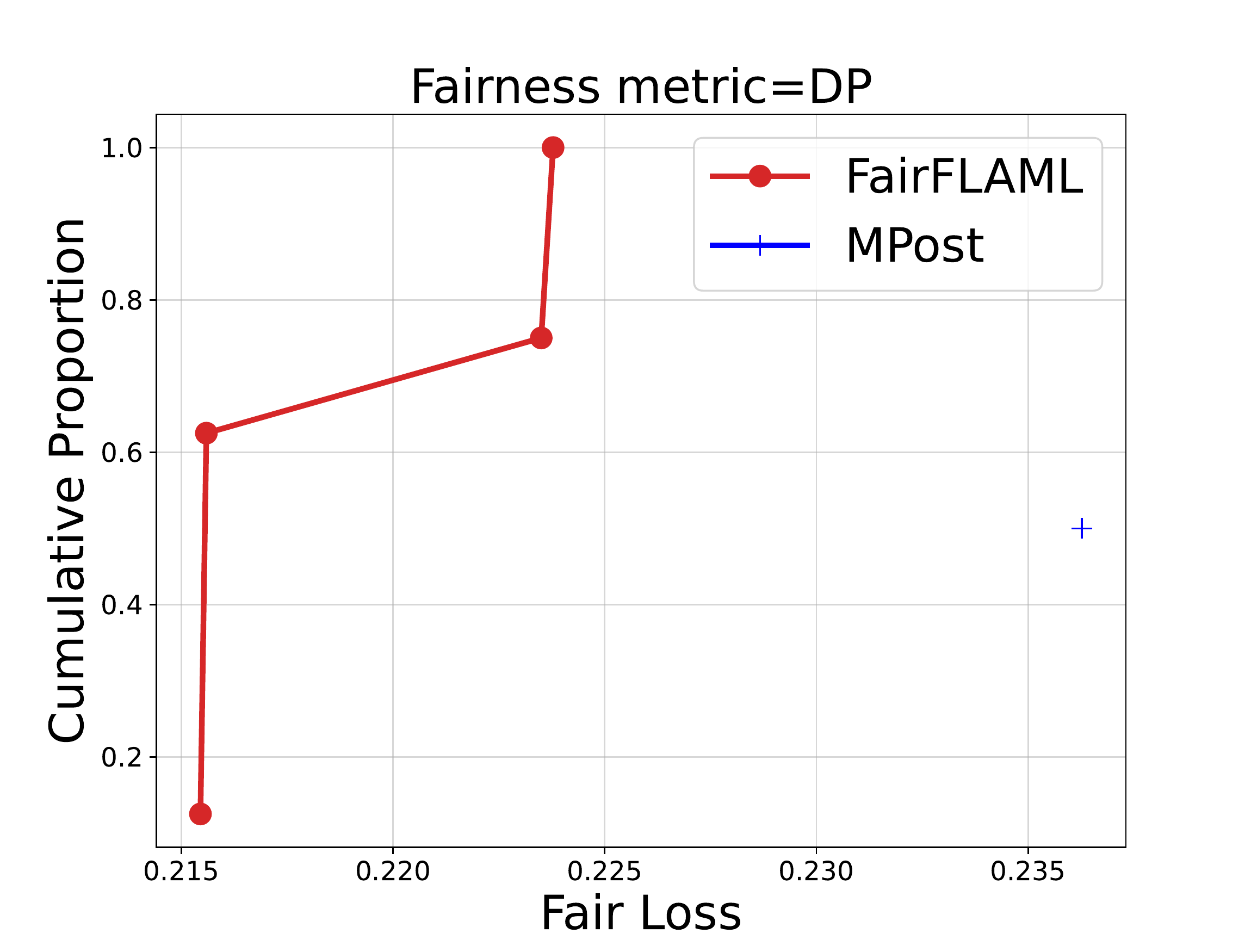}%
    \includegraphics[width=0.25\textwidth]{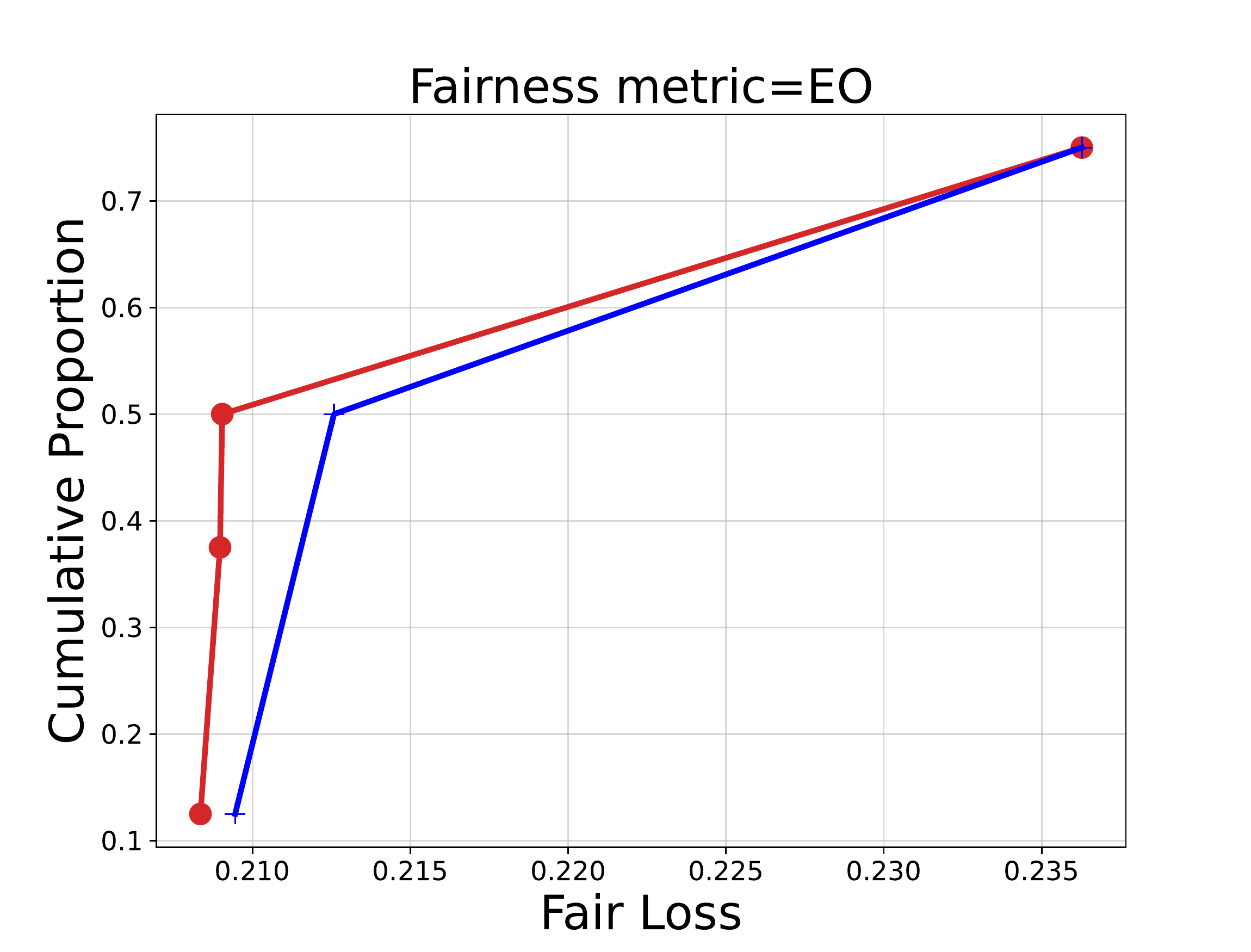}%
    \includegraphics[width=0.25\textwidth]{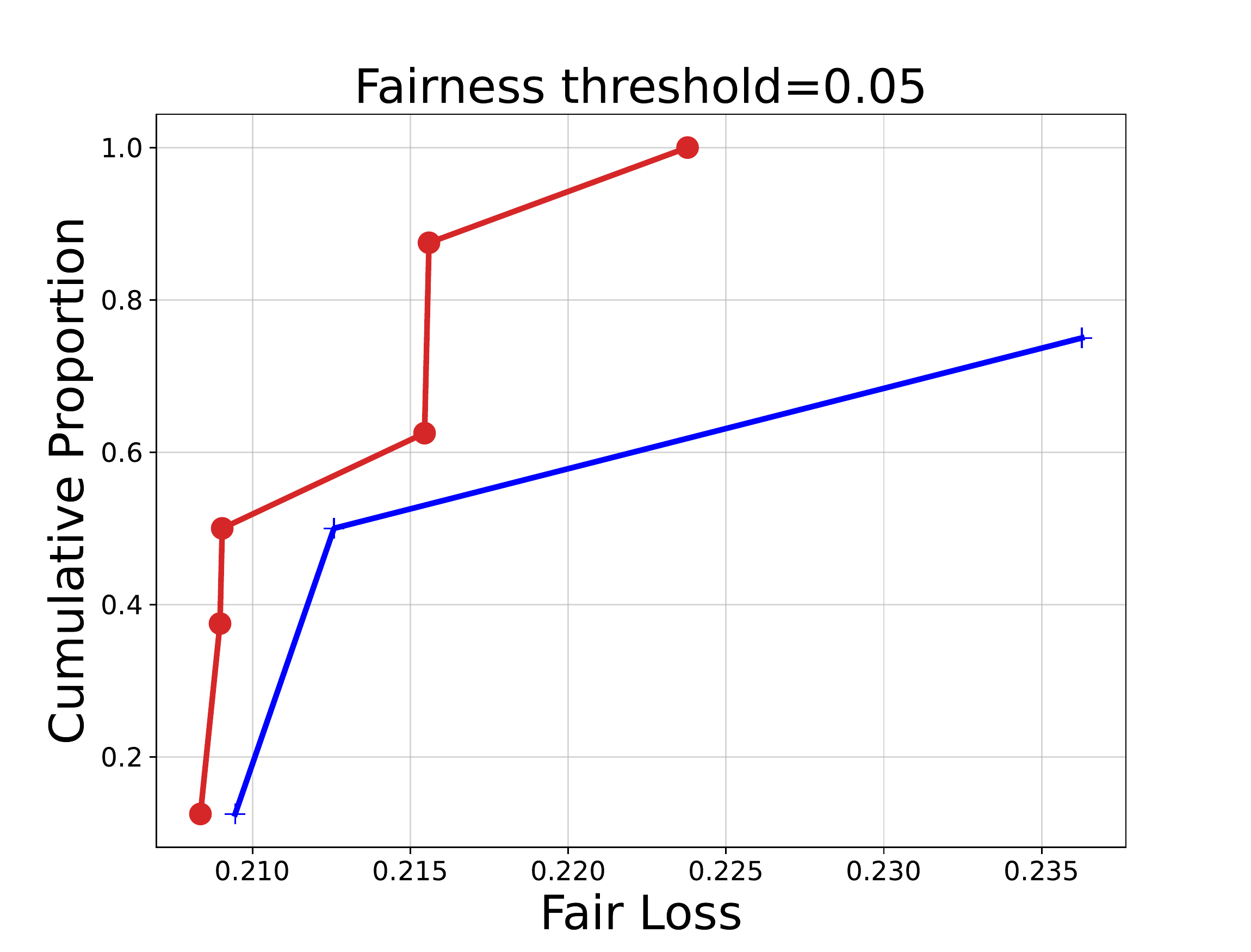}%
    \includegraphics[width=0.25\textwidth]{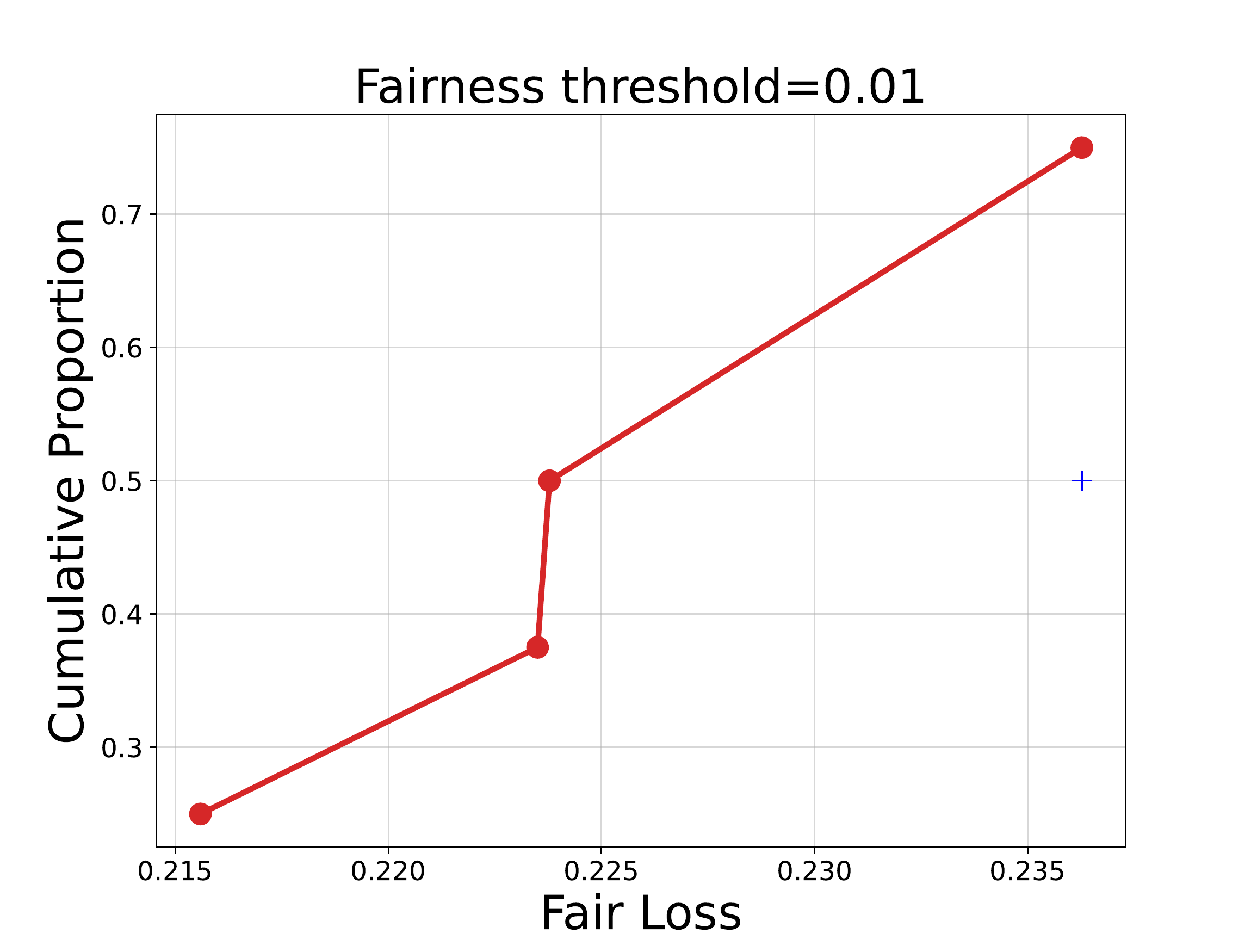}%
        }
        \\
    \subfigure[Wall-clock-time = 600s]{
    \includegraphics[width=0.25\textwidth]{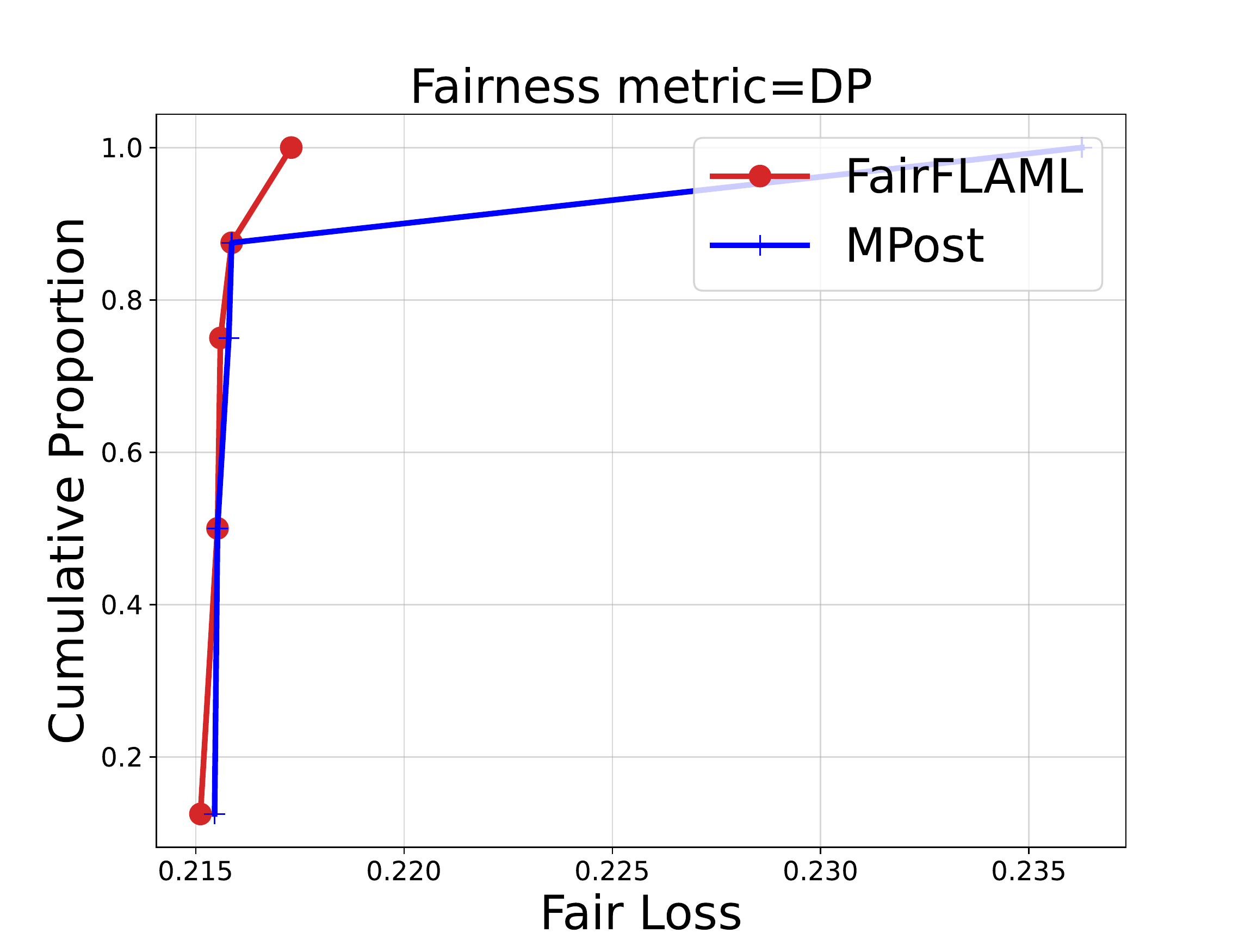}%
    \includegraphics[width=0.25\textwidth]{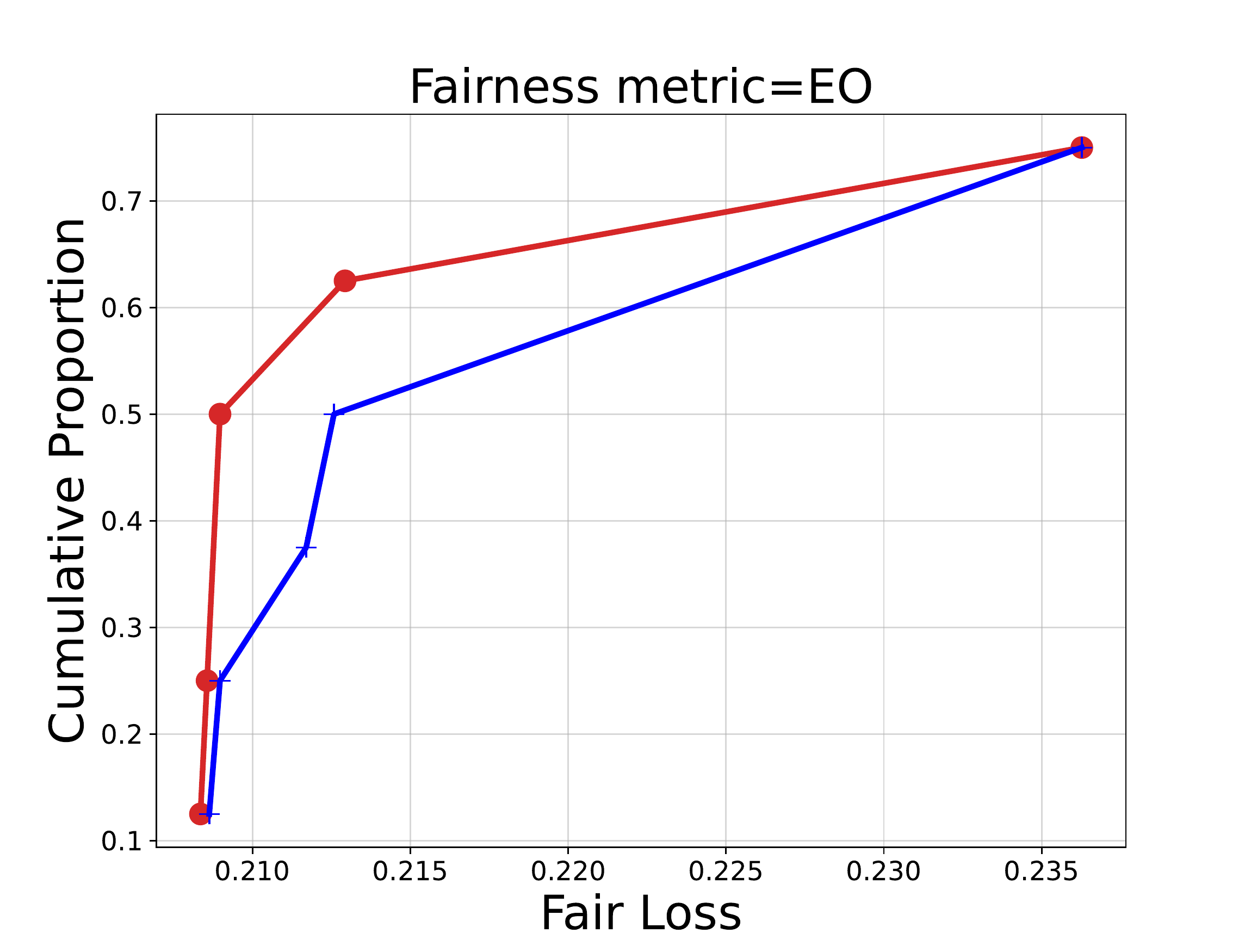}%
    \includegraphics[width=0.25\textwidth]{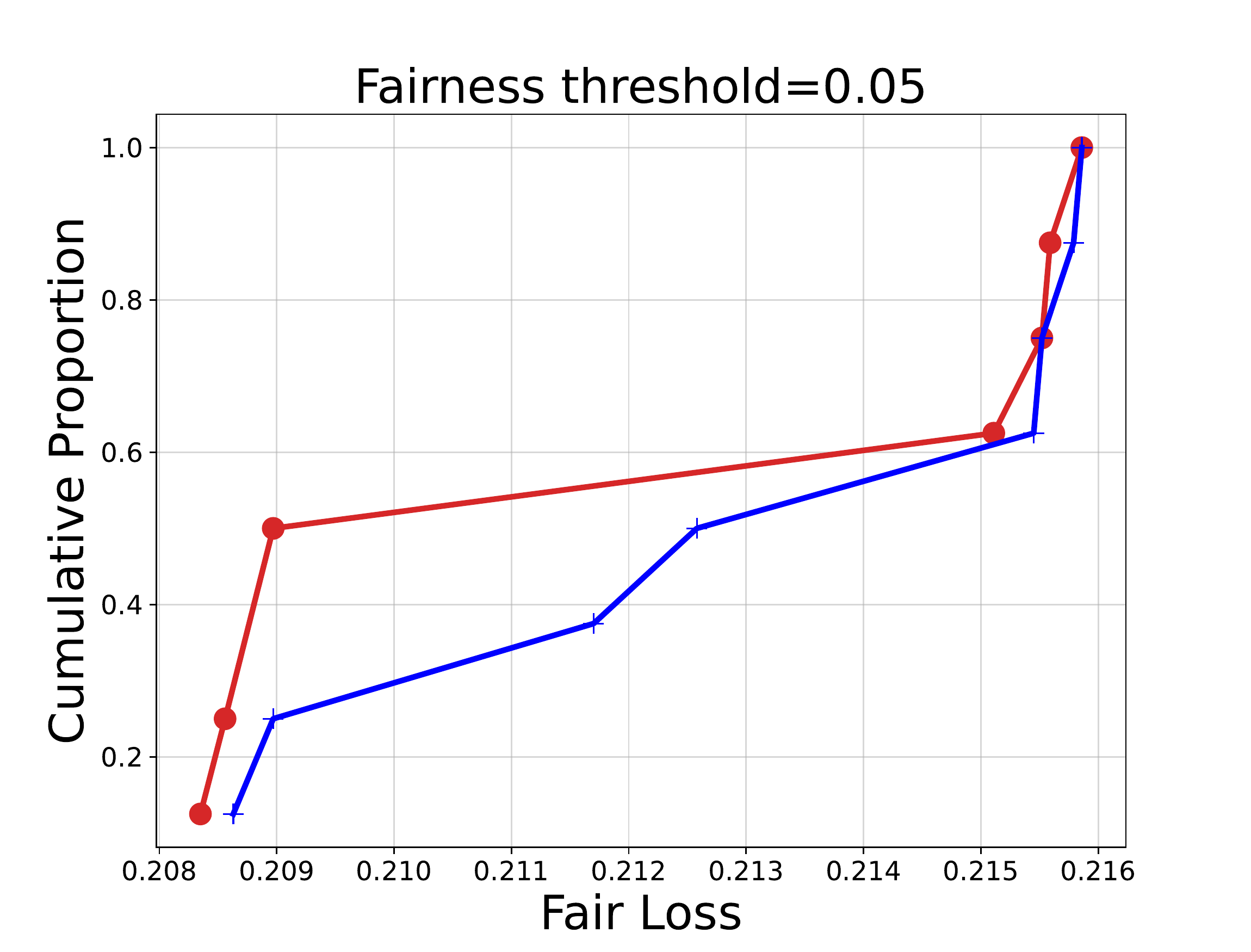}%
    \includegraphics[width=0.25\textwidth]{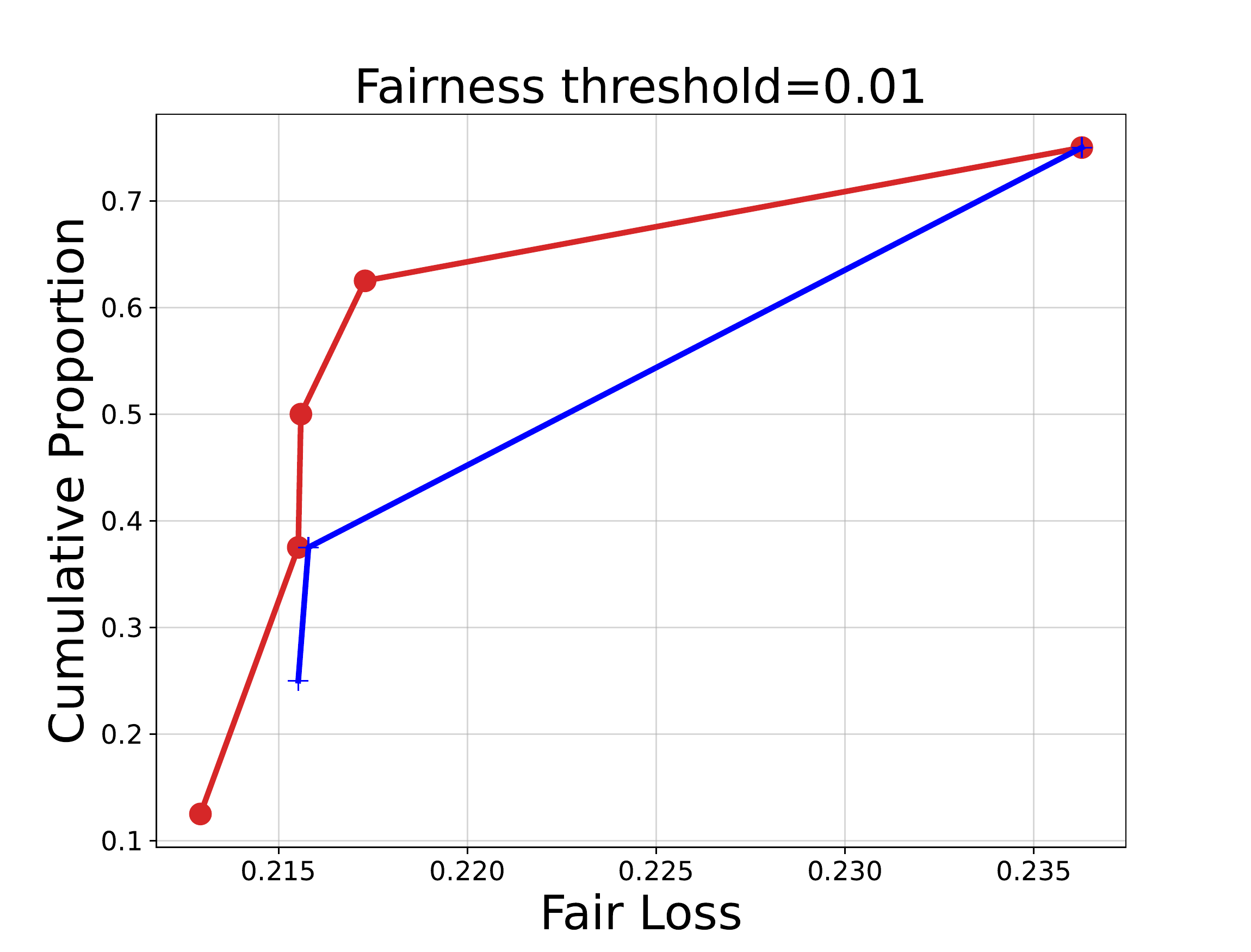}%
    }
    \caption{Comparisons of \fairFLAML with \PostFLAML on the \emph{Adult} dataset under different resource budgets when tuning XGBoost.}
    \label{fig:postflaml}
    \end{figure*}

However, there are pitfalls in this naive approach. (1) When the correlation between the loss before and after unfairness mitigation is weak, 
this approach may waste a lot of time doing unnecessary unfairness mitigation. Even if the correlation is strong, the relationship between the losses before and after mitigation is typically non-monotonic. 
(2) To remain an end-to-end solution, the system still needs to decide when to stop the original AutoML process and apply the mitigation operation, which is non-trivial especially if we want to apply the mitigation on multiple models. If this stopping time is not set properly, the system may not be able to produce a single fair model. To verify our argument, we constructed a version of this approach in the following way: given a particular resource budget, the method spends the first half of the budget doing hyperparameter search, and the second half of the budget doing unfairness mitigation. We name this method \PostFLAML, and compare it with \fairFLAML with different resource levels in Figure~\ref{fig:postflaml}. The results verified our arguments that the performance of \PostFLAML is sensitive to the resource budget. In general, \PostFLAML requires a sufficiently large resource budget. 

\subsection{Results on more unfairness mitigation methods}
\begin{figure*}
    \centering
    \subfigure[Fair loss on \emph{Adult}]{
  \includegraphics[width=0.25\textwidth]{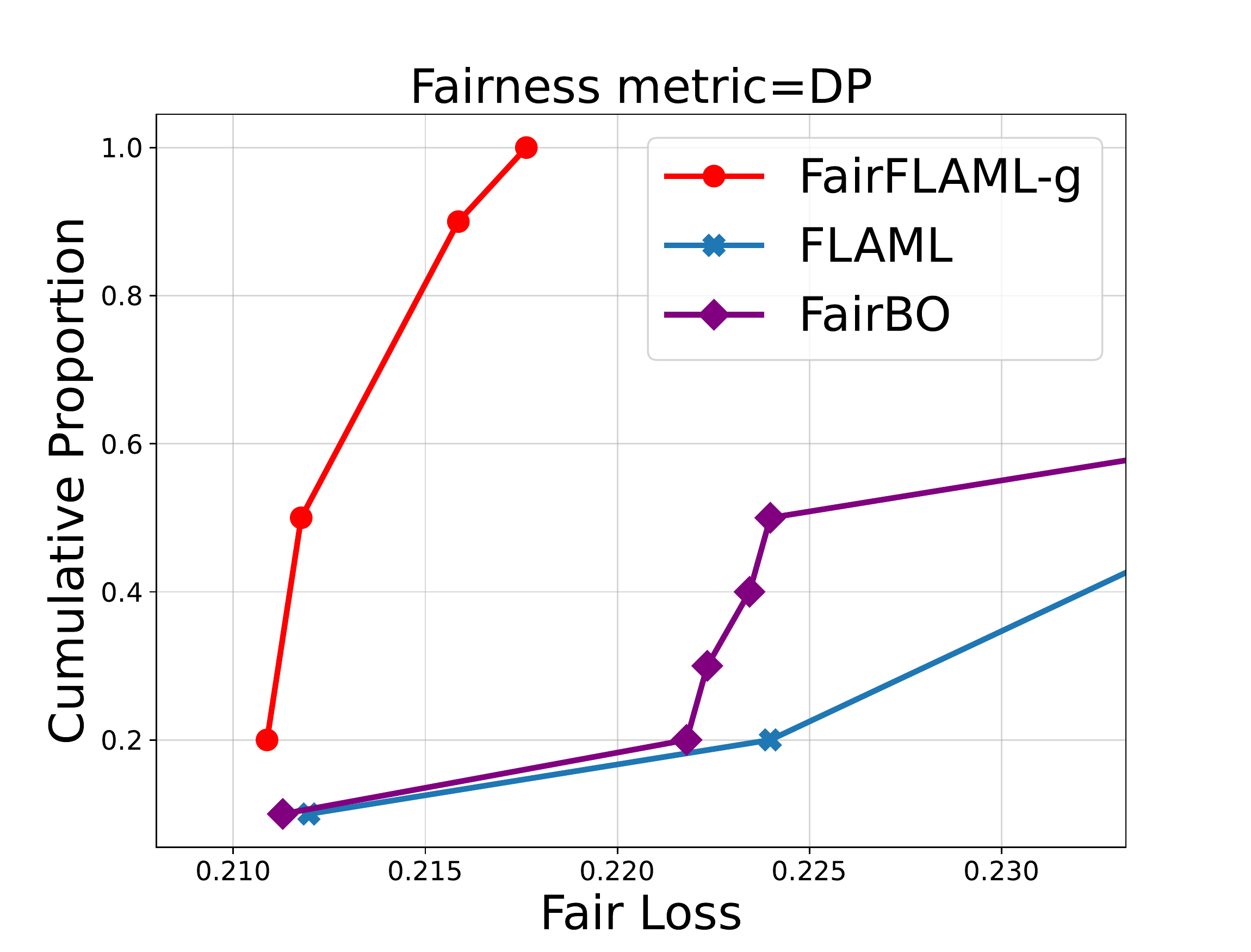}%
  \includegraphics[width=0.25\textwidth]{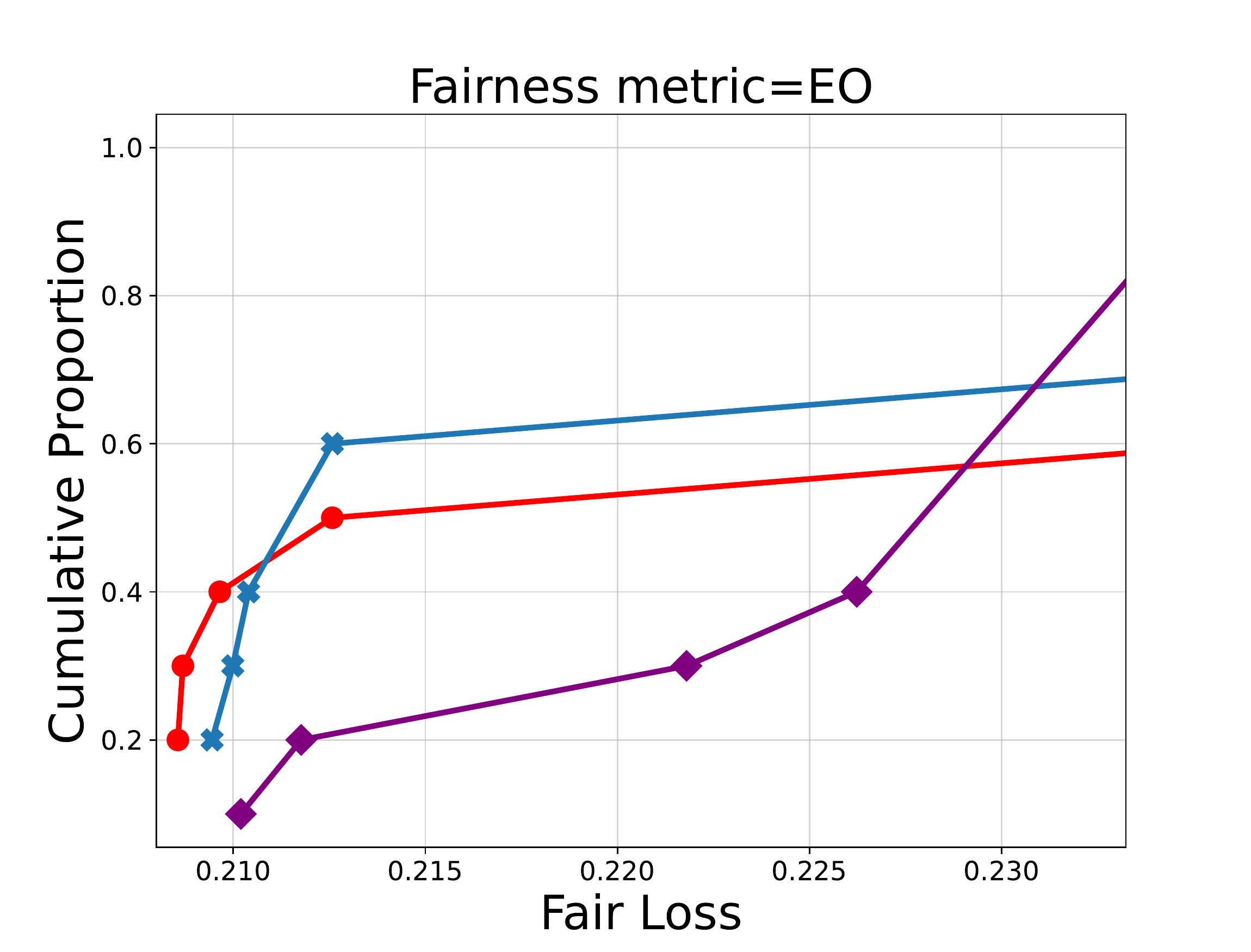}%
  \includegraphics[width=0.25\textwidth]{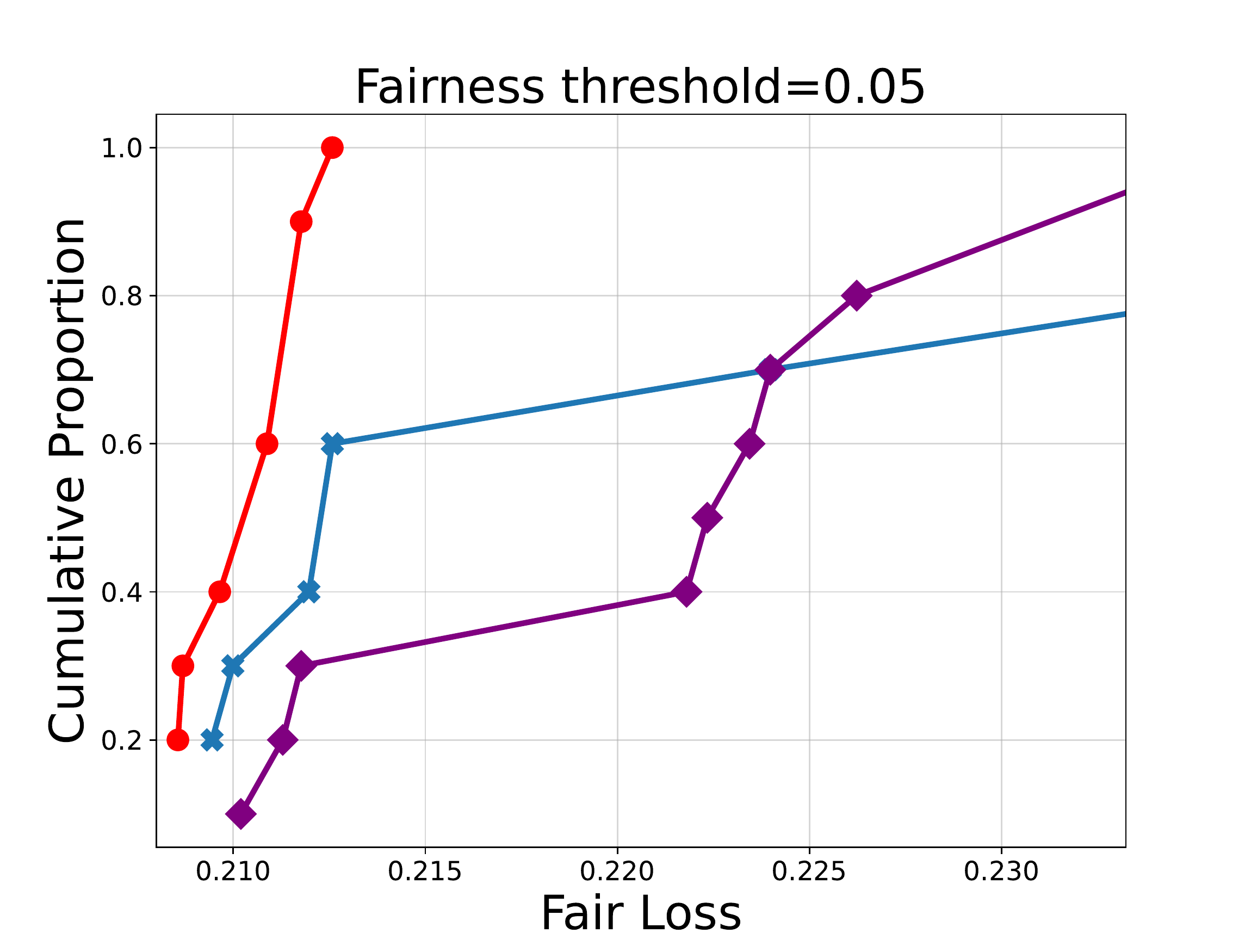}%
  \includegraphics[width=0.25\textwidth]{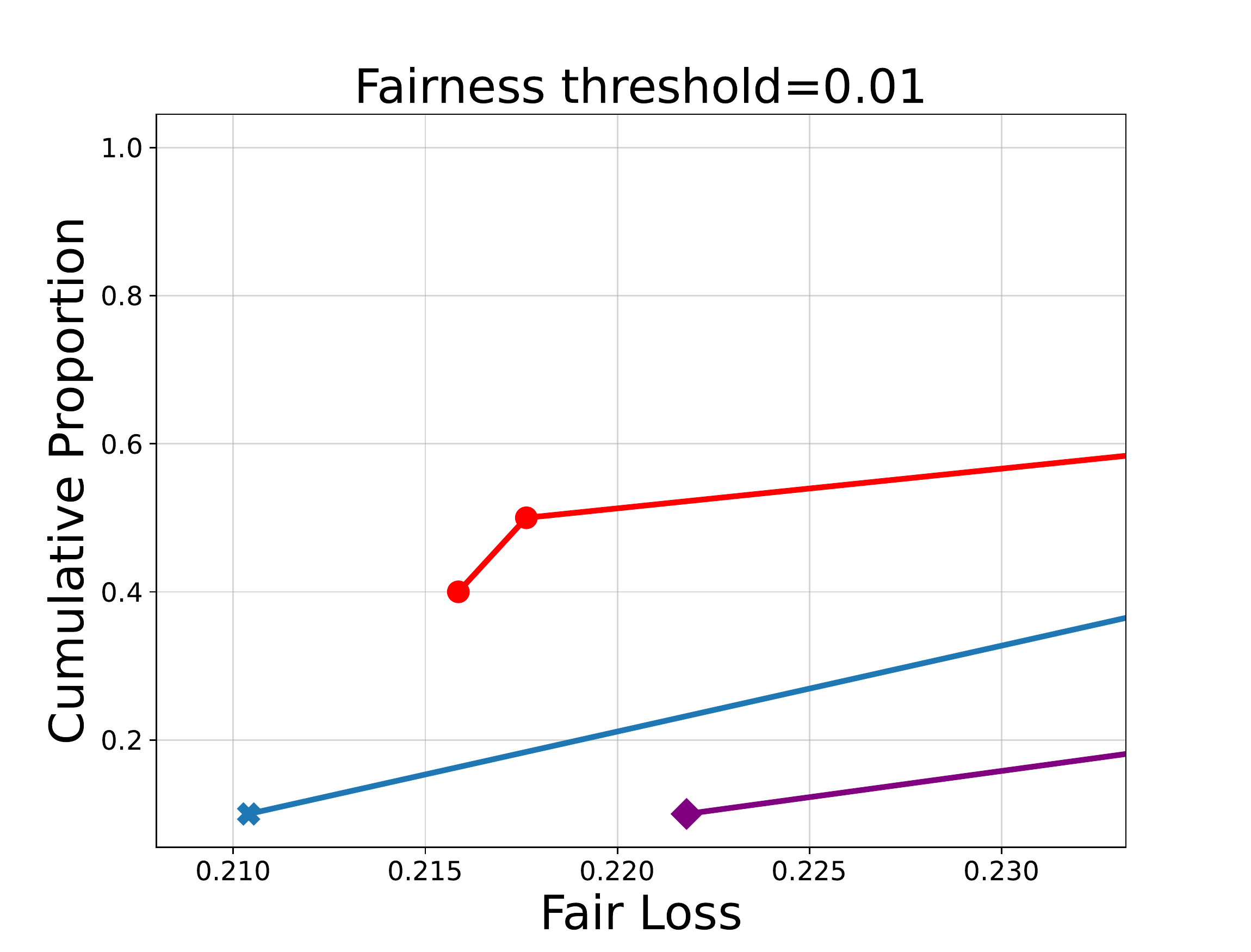}
  }
  \\
  \subfigure[Fair loss on \emph{Bank}]{
  \includegraphics[width=0.25\textwidth]{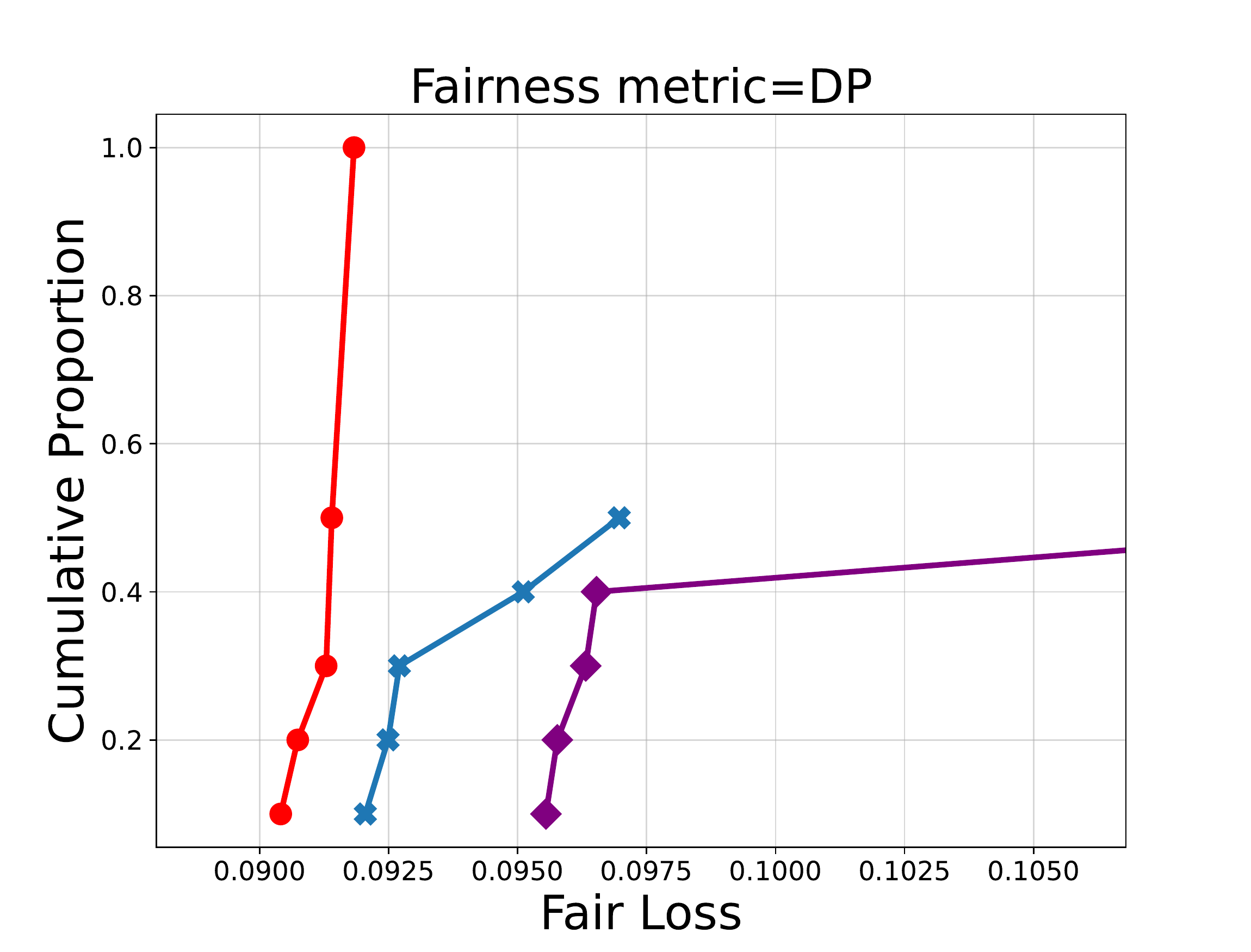}%
  \includegraphics[width=0.25\textwidth]{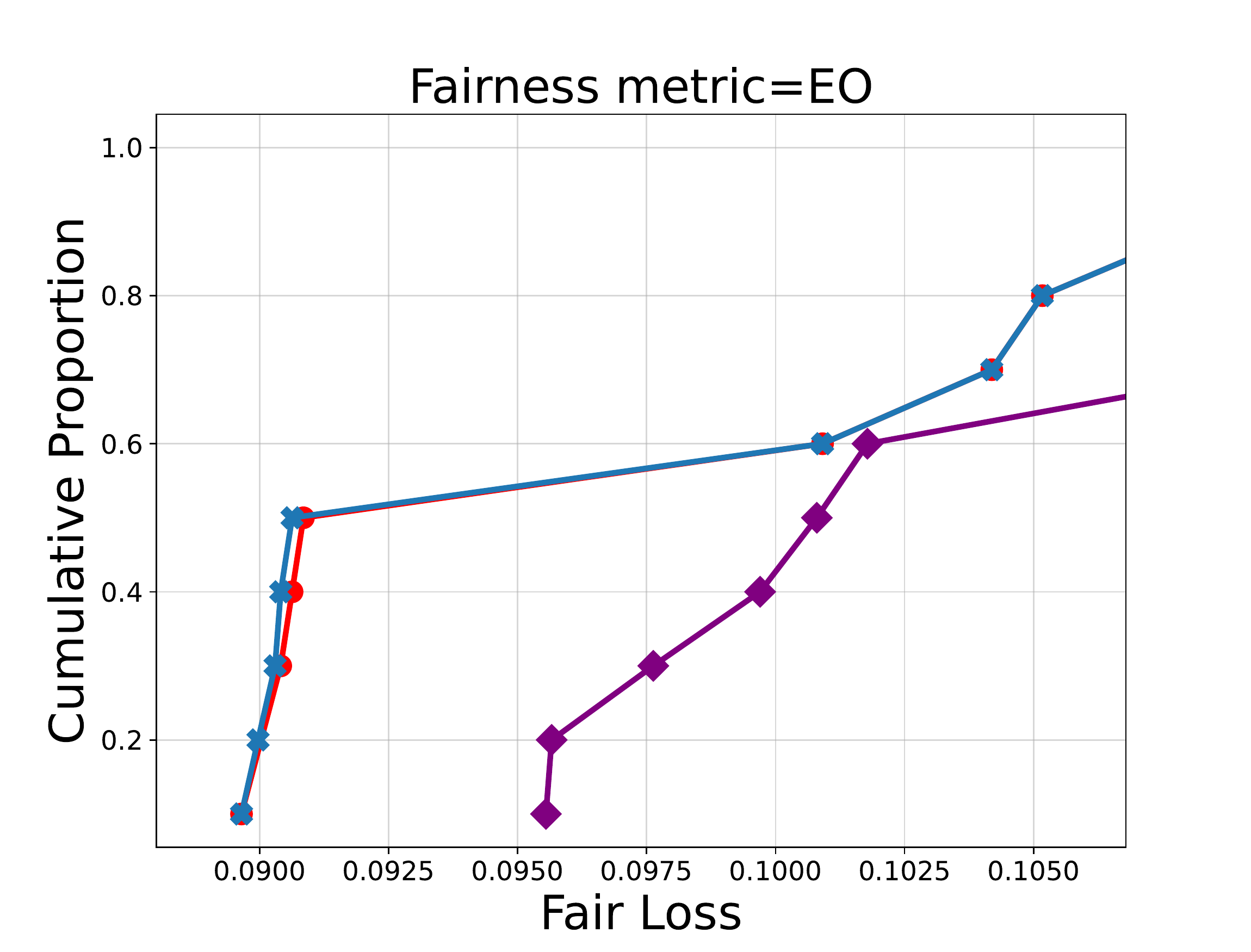}%
  \includegraphics[width=0.25\textwidth]{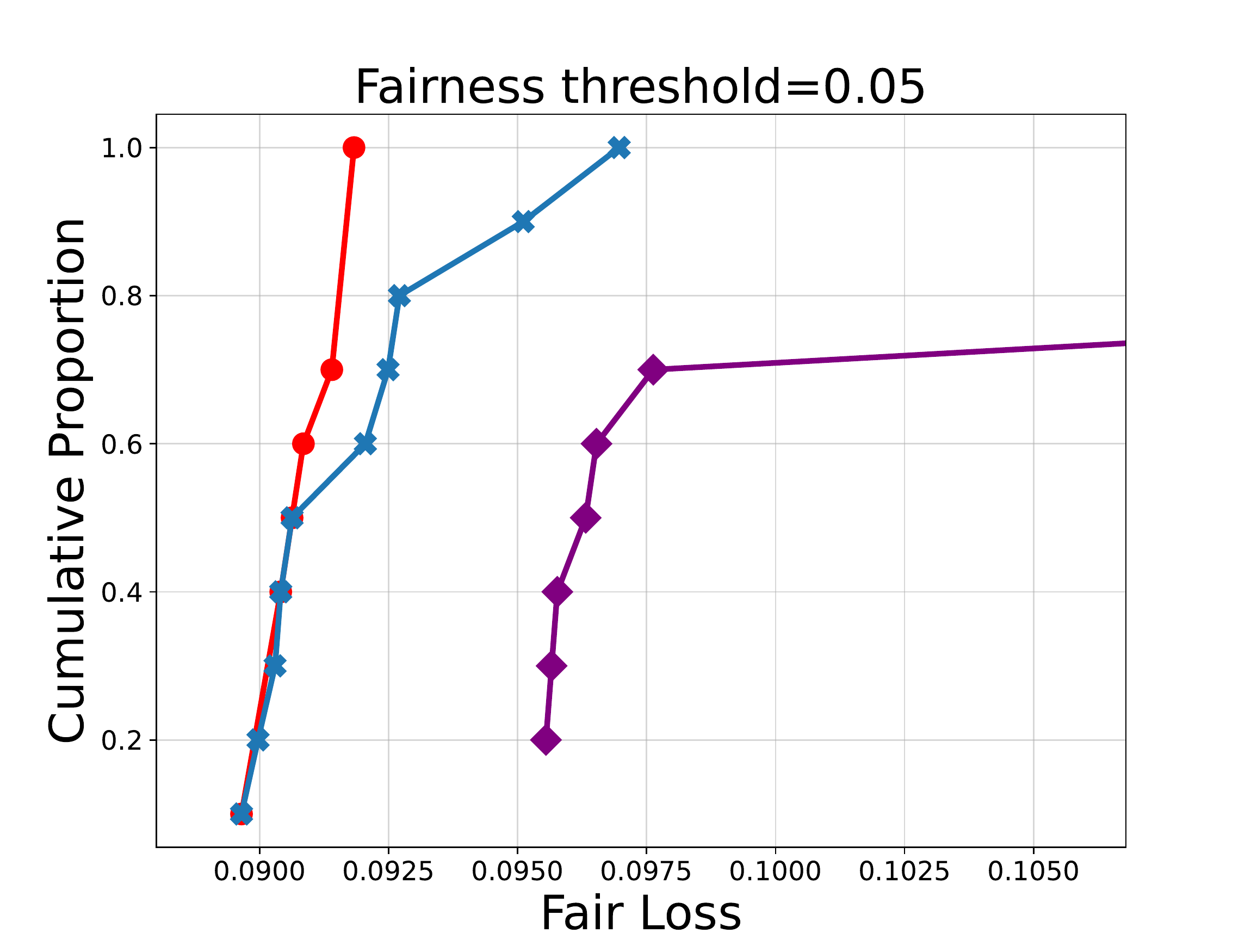}%
  \includegraphics[width=0.25\textwidth]{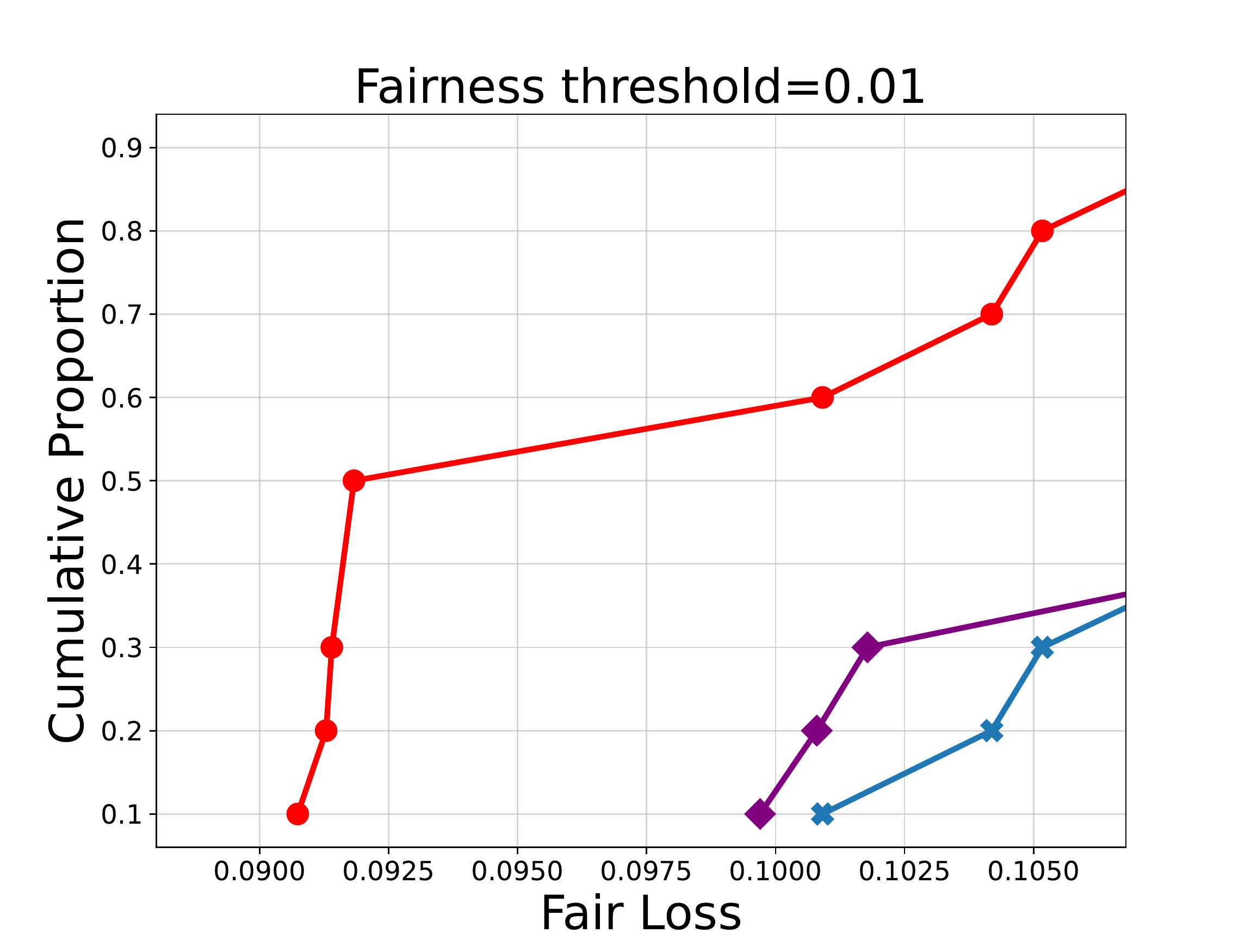}
  }
  \\
  \subfigure[Fair loss on \emph{Compas}]{
  \includegraphics[width=0.25\textwidth]{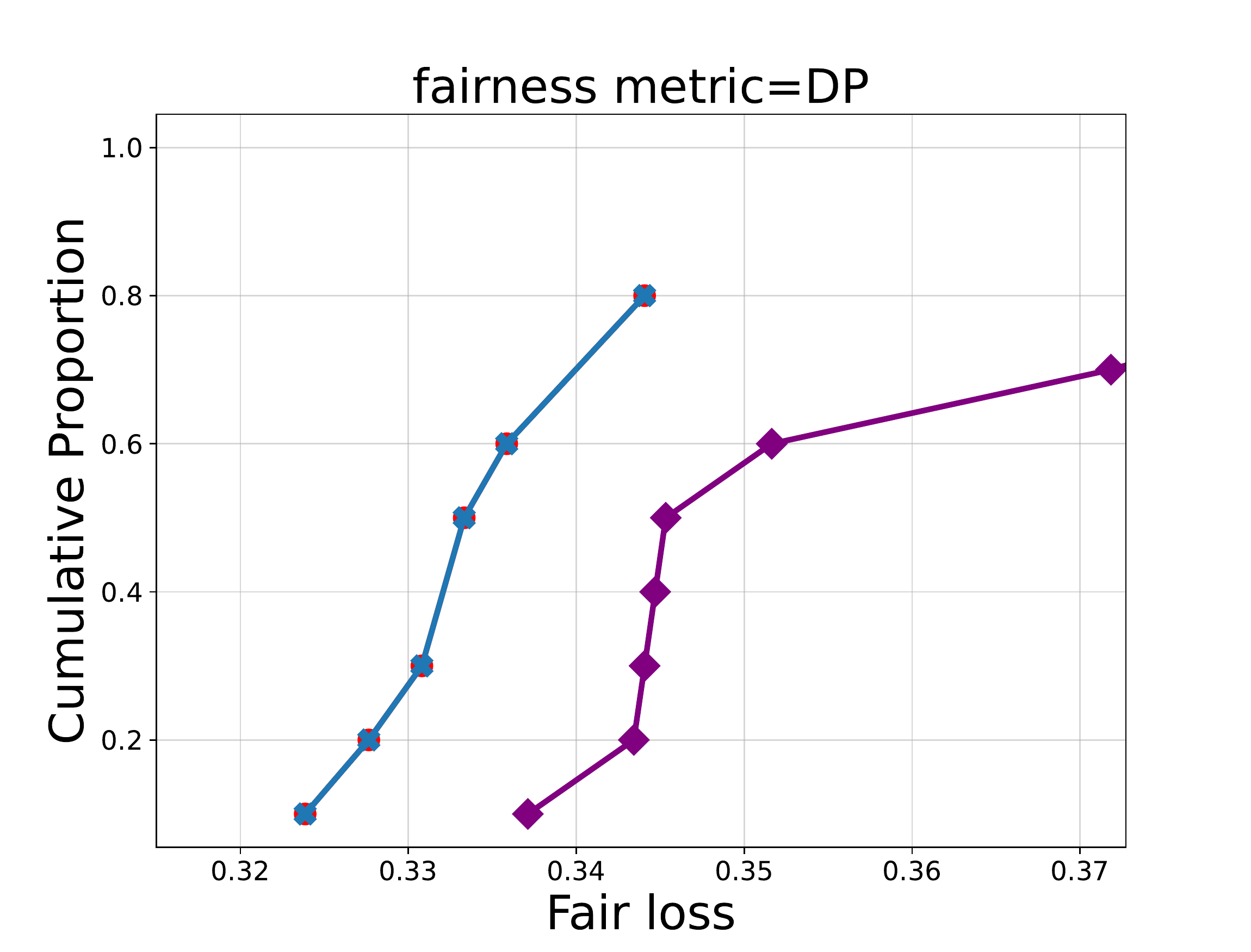}%
  \includegraphics[width=0.25\textwidth]{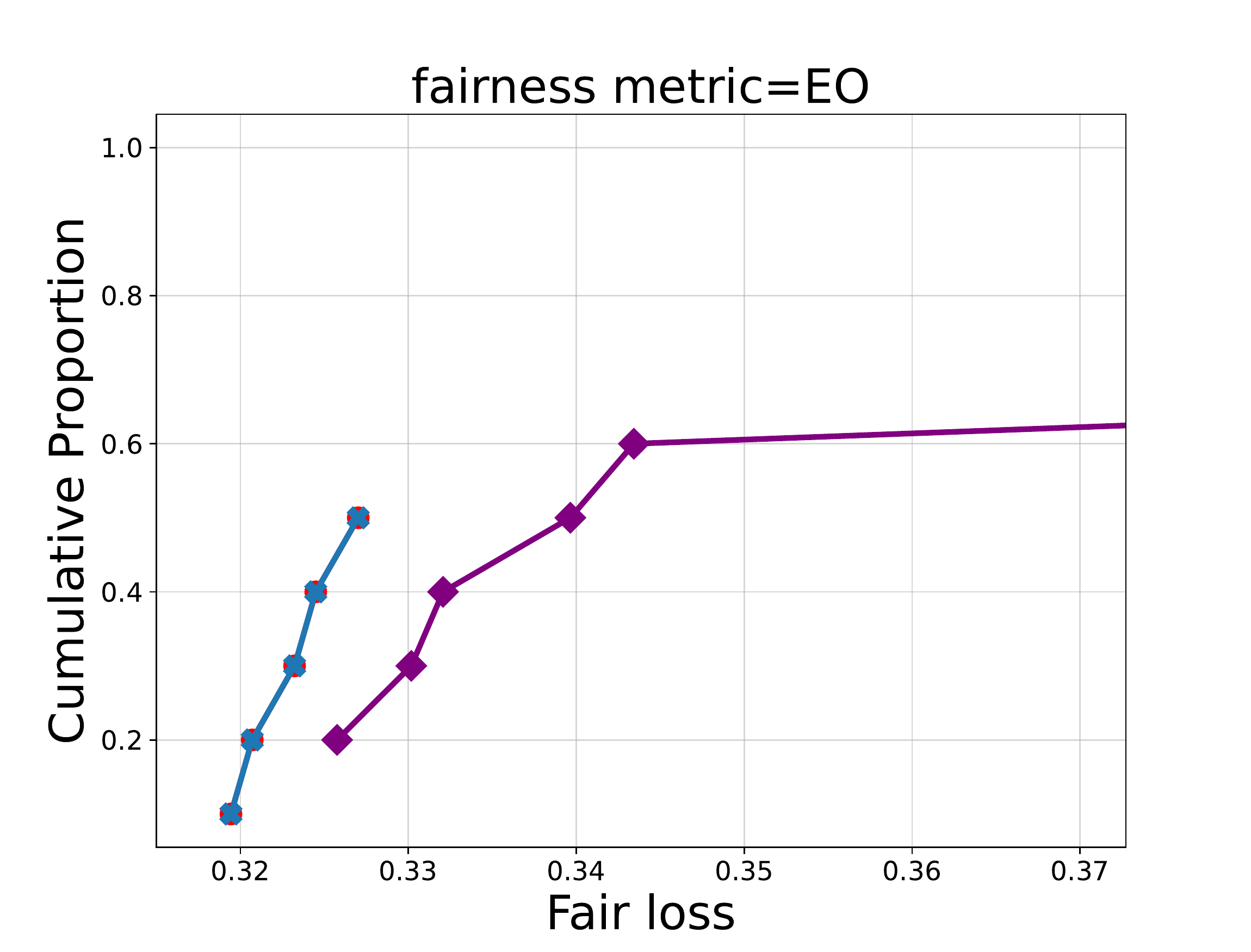}%
  \includegraphics[width=0.25\textwidth]{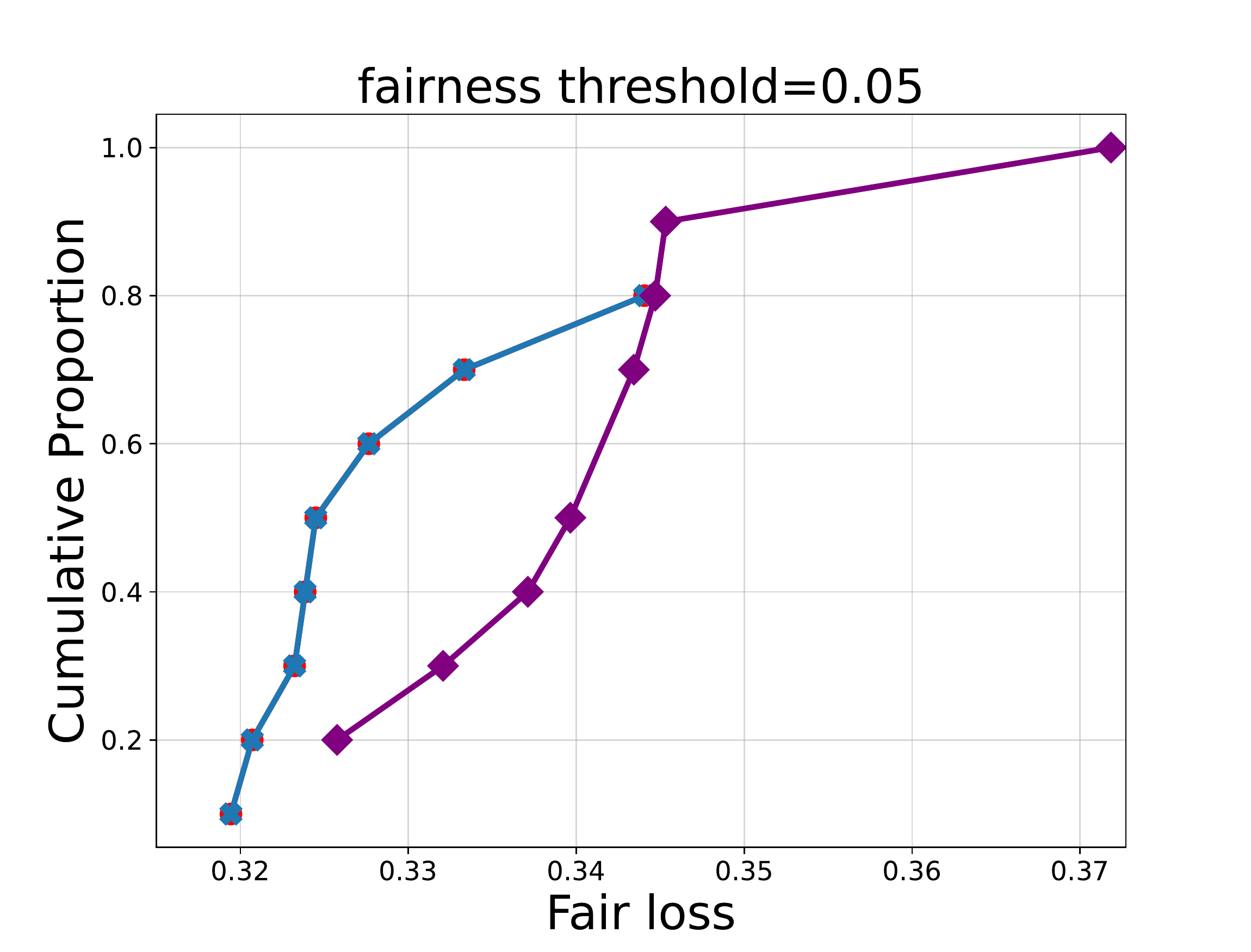}%
  \includegraphics[width=0.25\textwidth]{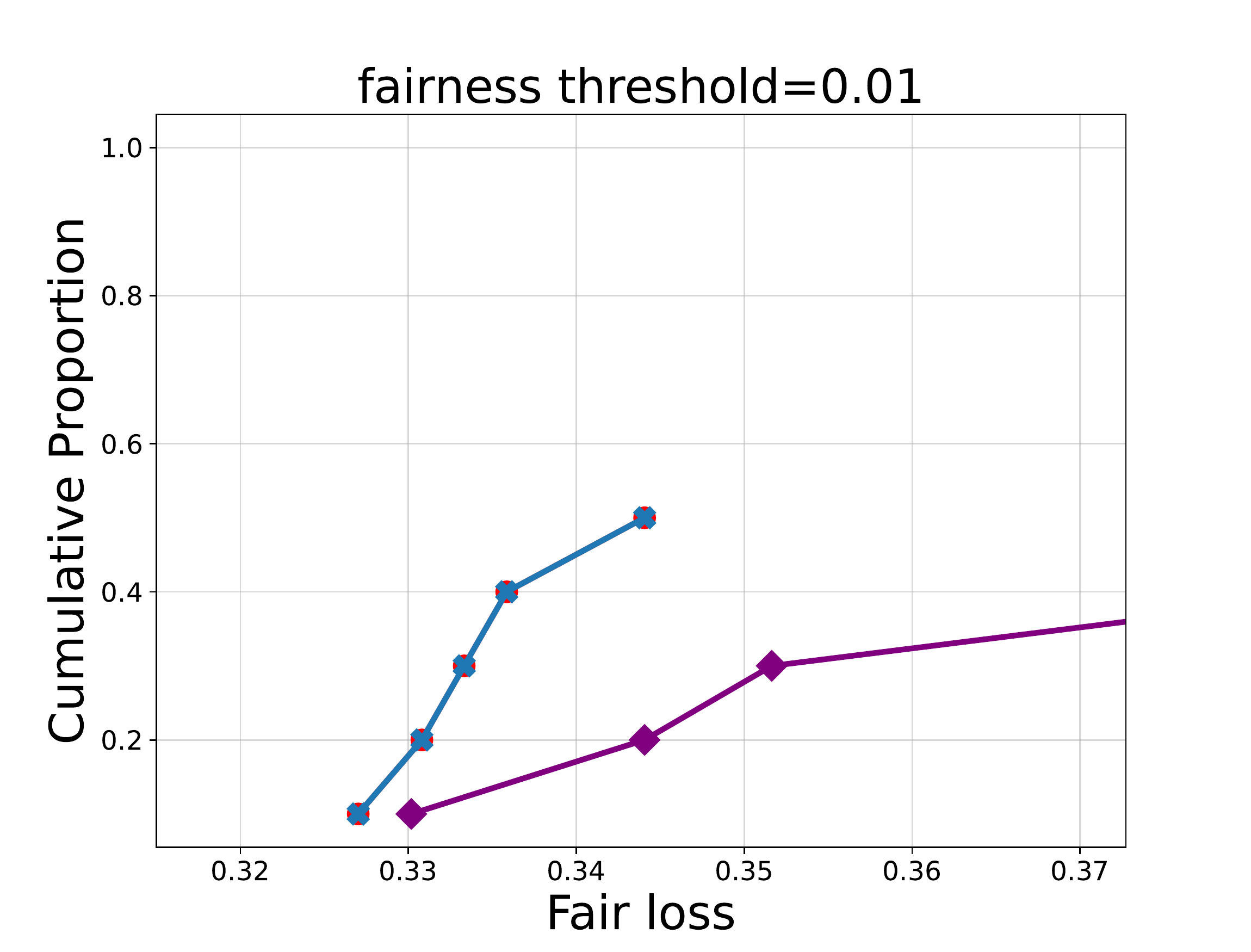}
  }
  \\
  \subfigure[Fair loss on \emph{MEPS}]{
  \includegraphics[width=0.25\textwidth]{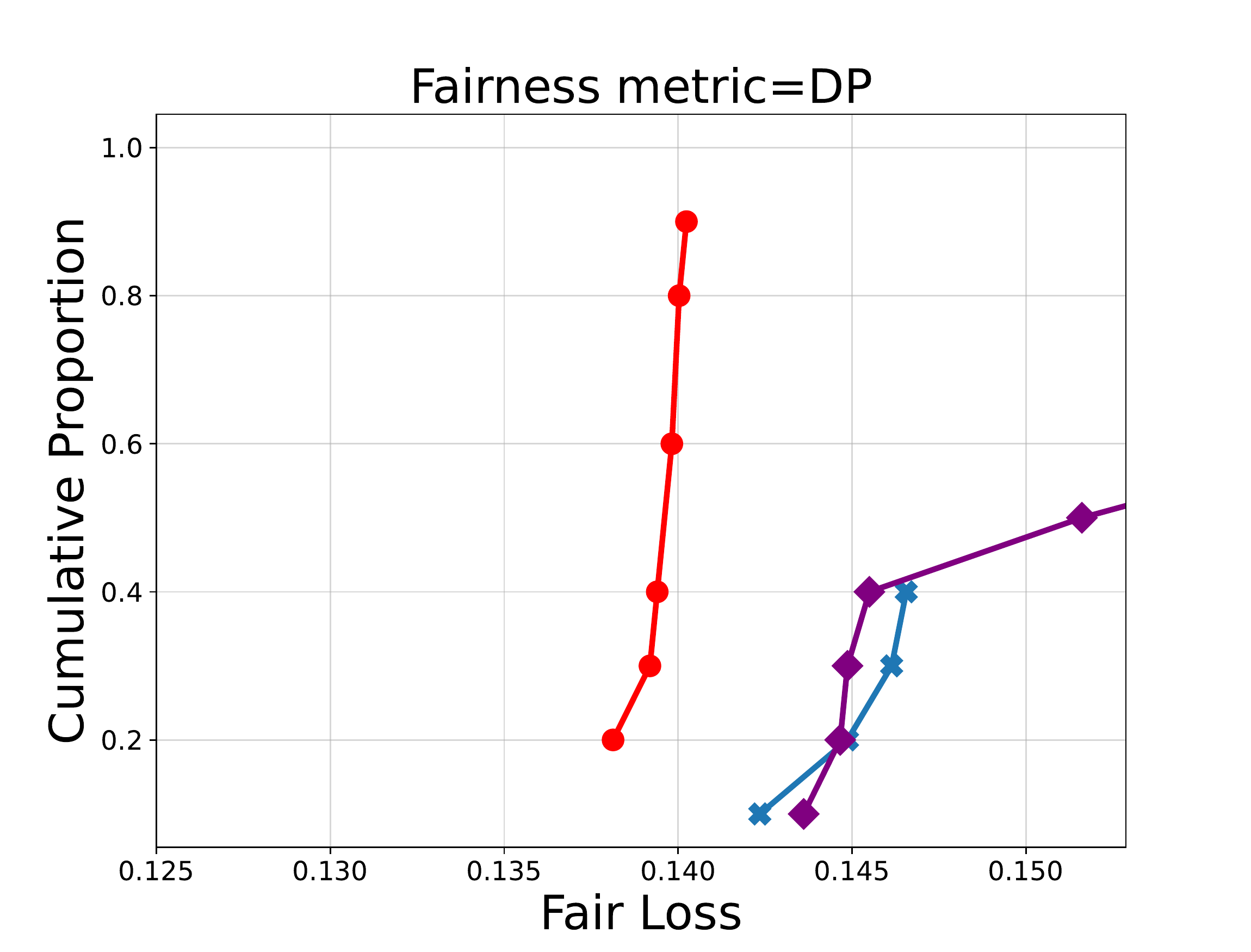}%
  \includegraphics[width=0.25\textwidth]{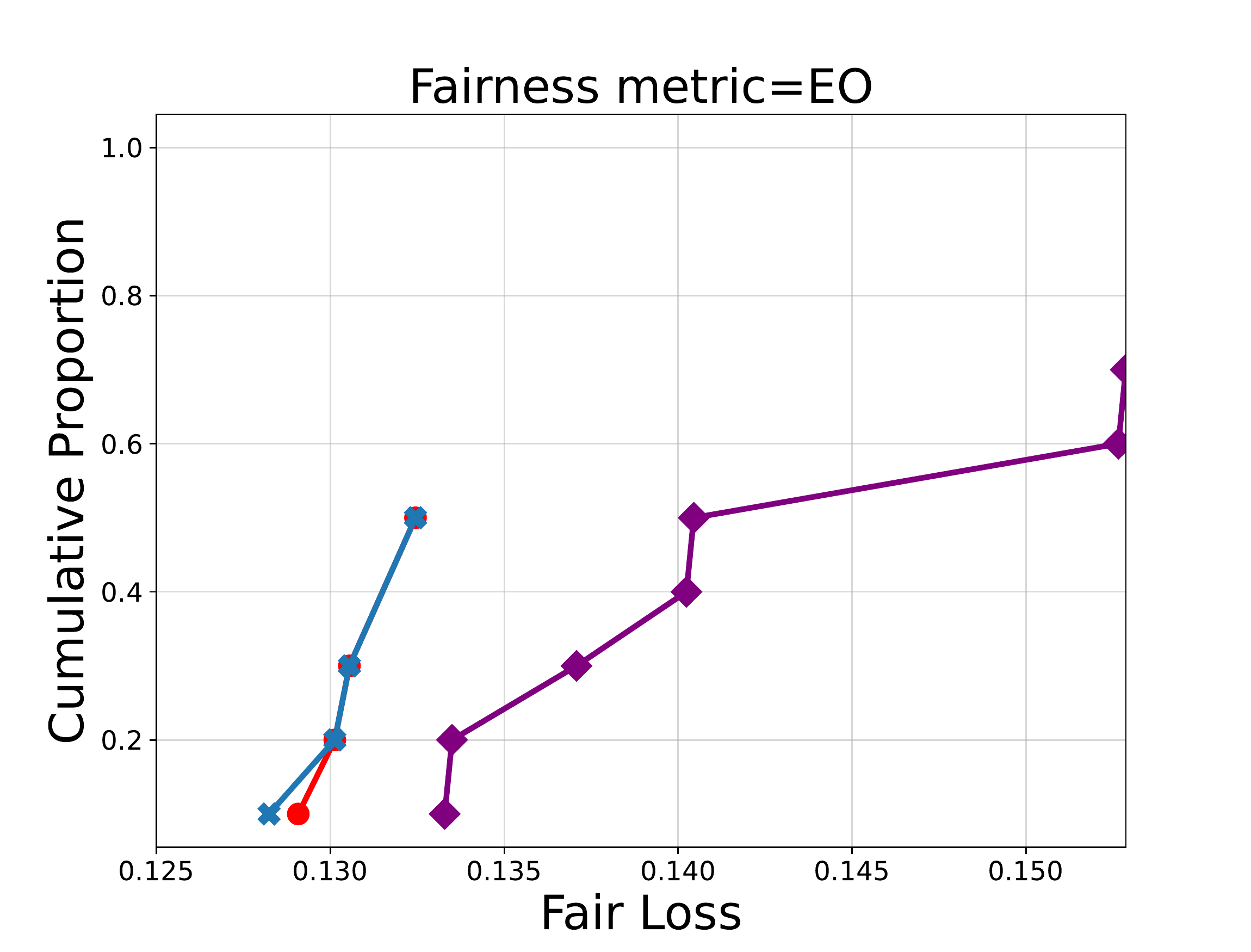}%
  \includegraphics[width=0.25\textwidth]{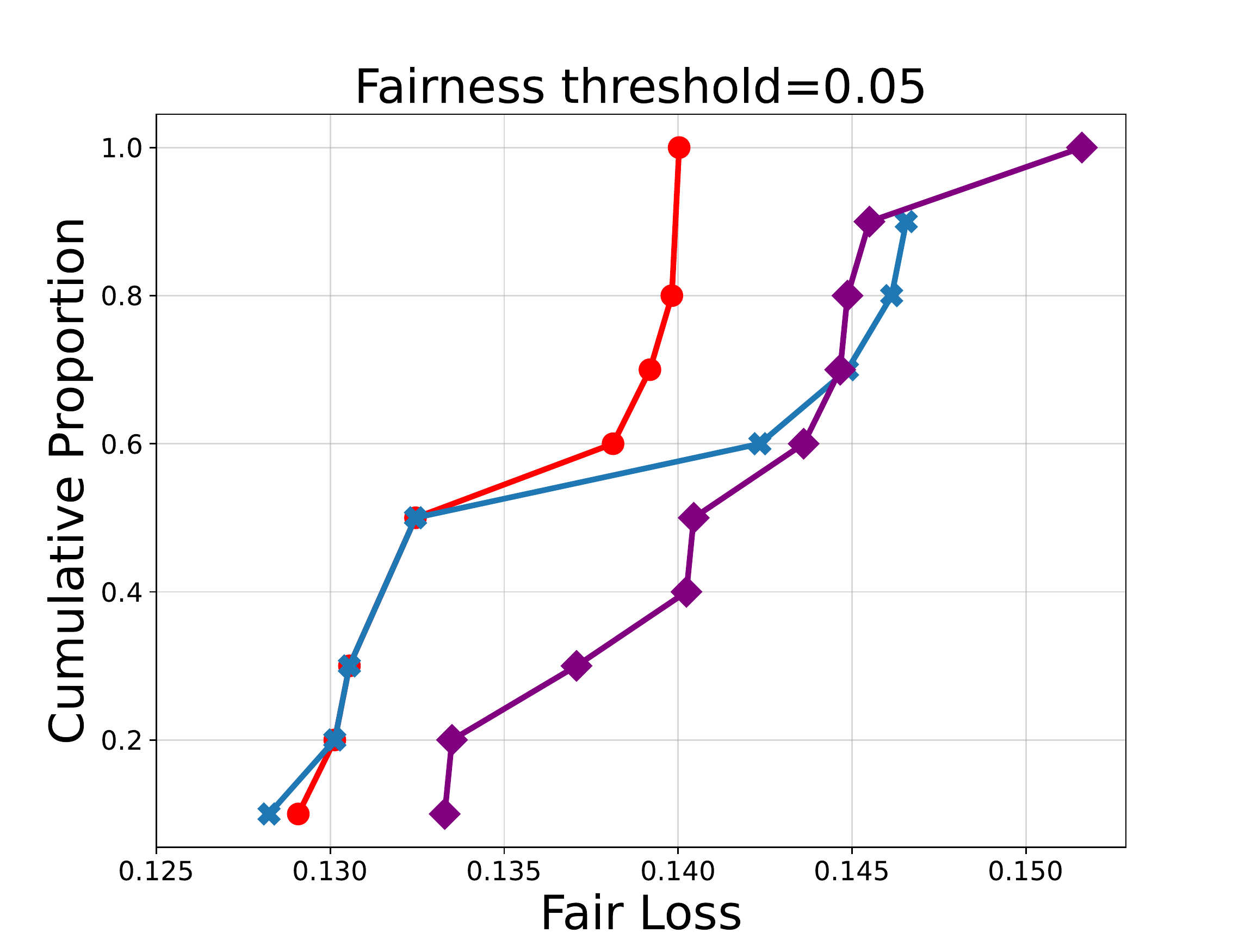}%
  \includegraphics[width=0.25\textwidth]{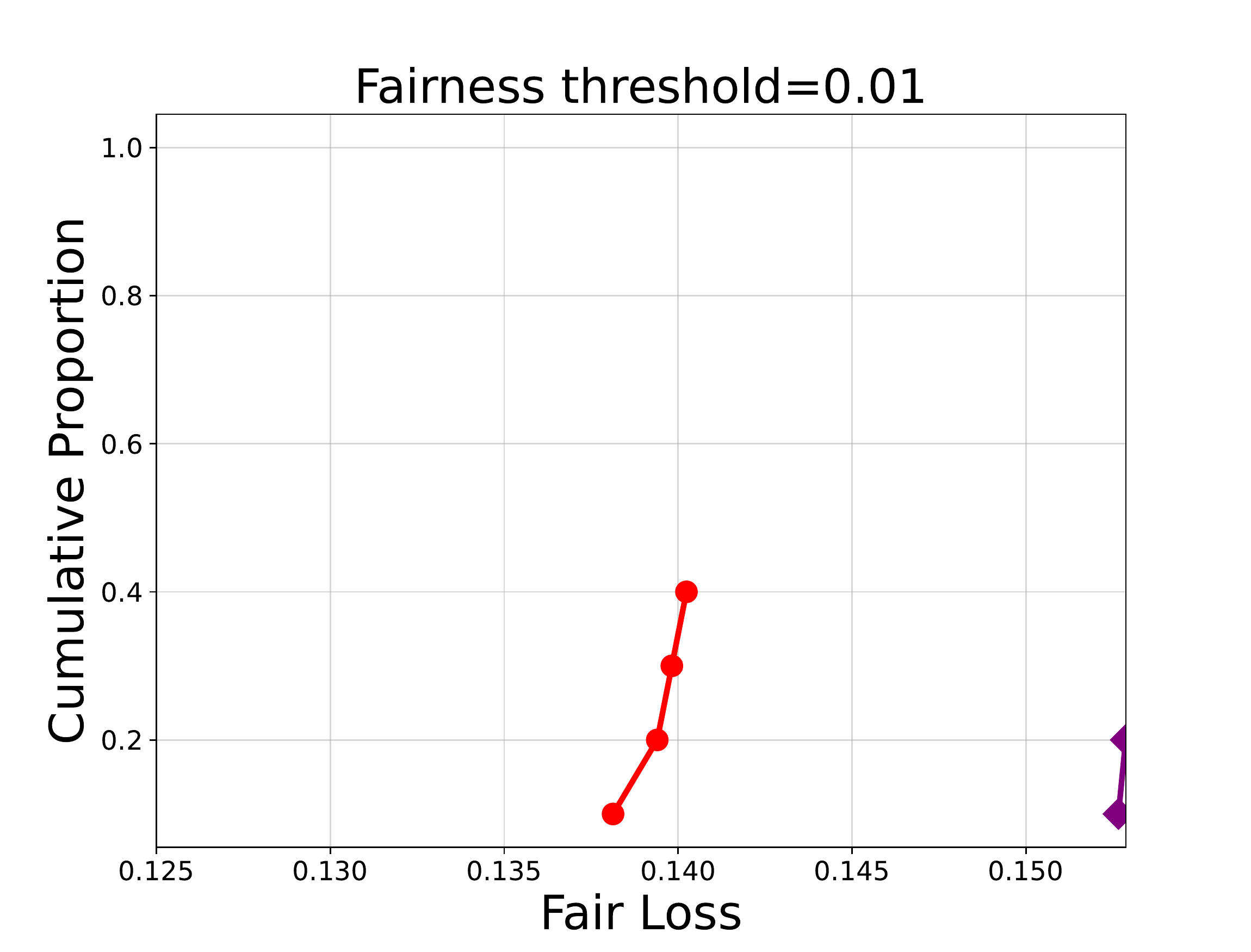}
  }
  \\
  \caption{Fair loss with grid search reduction as the unfairness mitigation method when tuning XGBoost.}
   \label{fig:exp_res_boxplot_grid}
  \end{figure*}

\begin{figure*}
    \centering
    \subfigure[Fair loss on \emph{Adult}]{
  \includegraphics[width=0.25\textwidth]{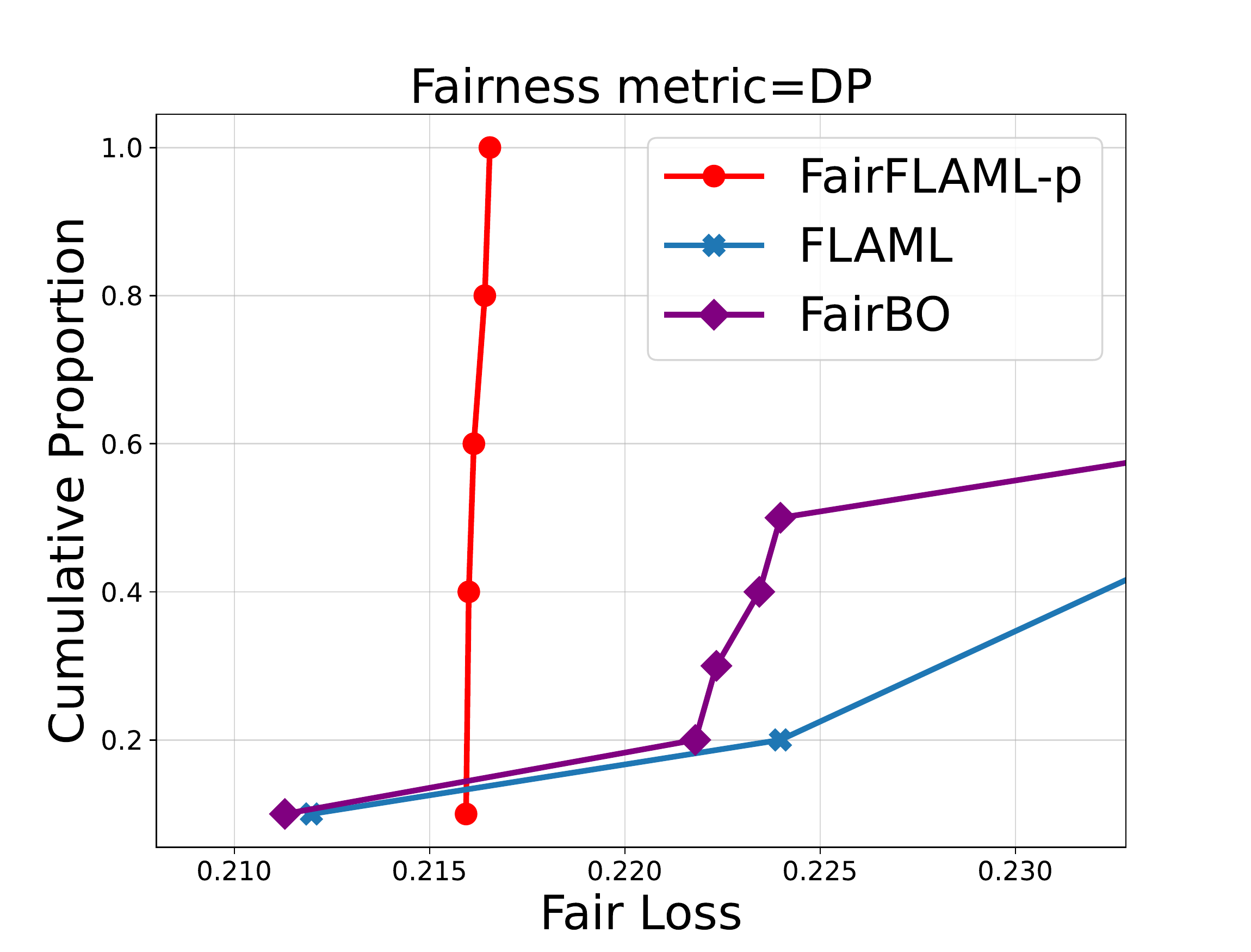}%
  \includegraphics[width=0.25\textwidth]{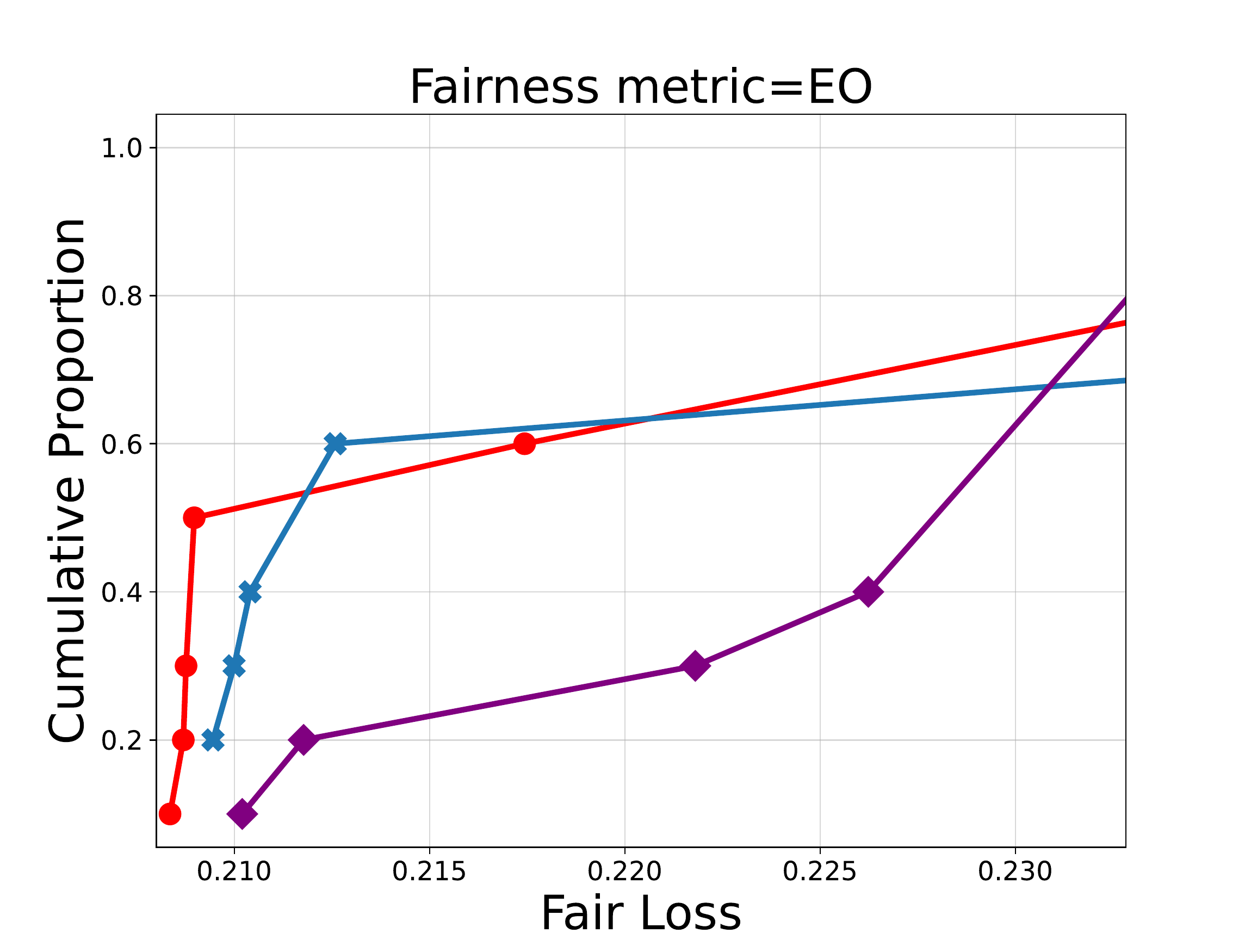}%
  \includegraphics[width=0.25\textwidth]{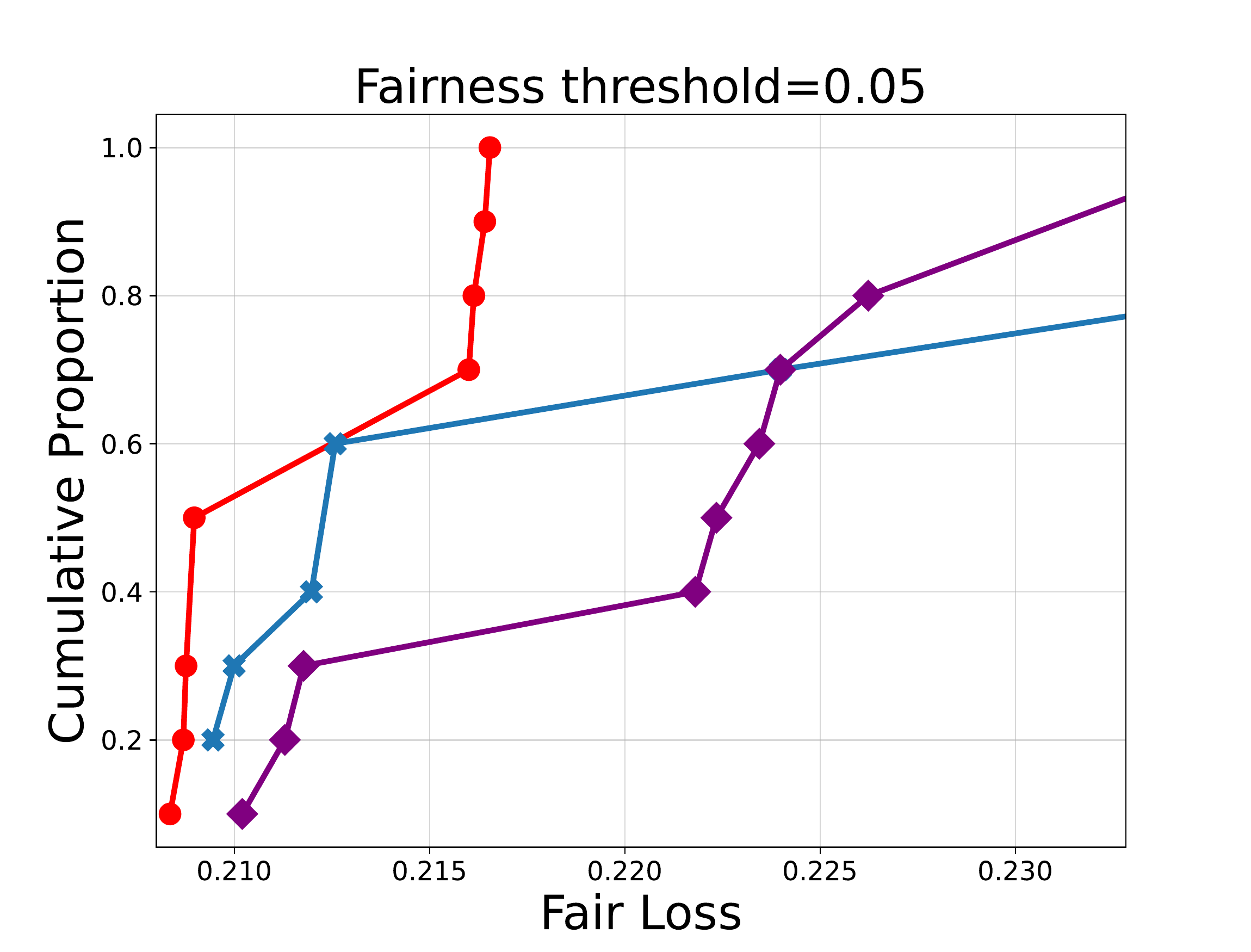}%
  \includegraphics[width=0.25\textwidth]{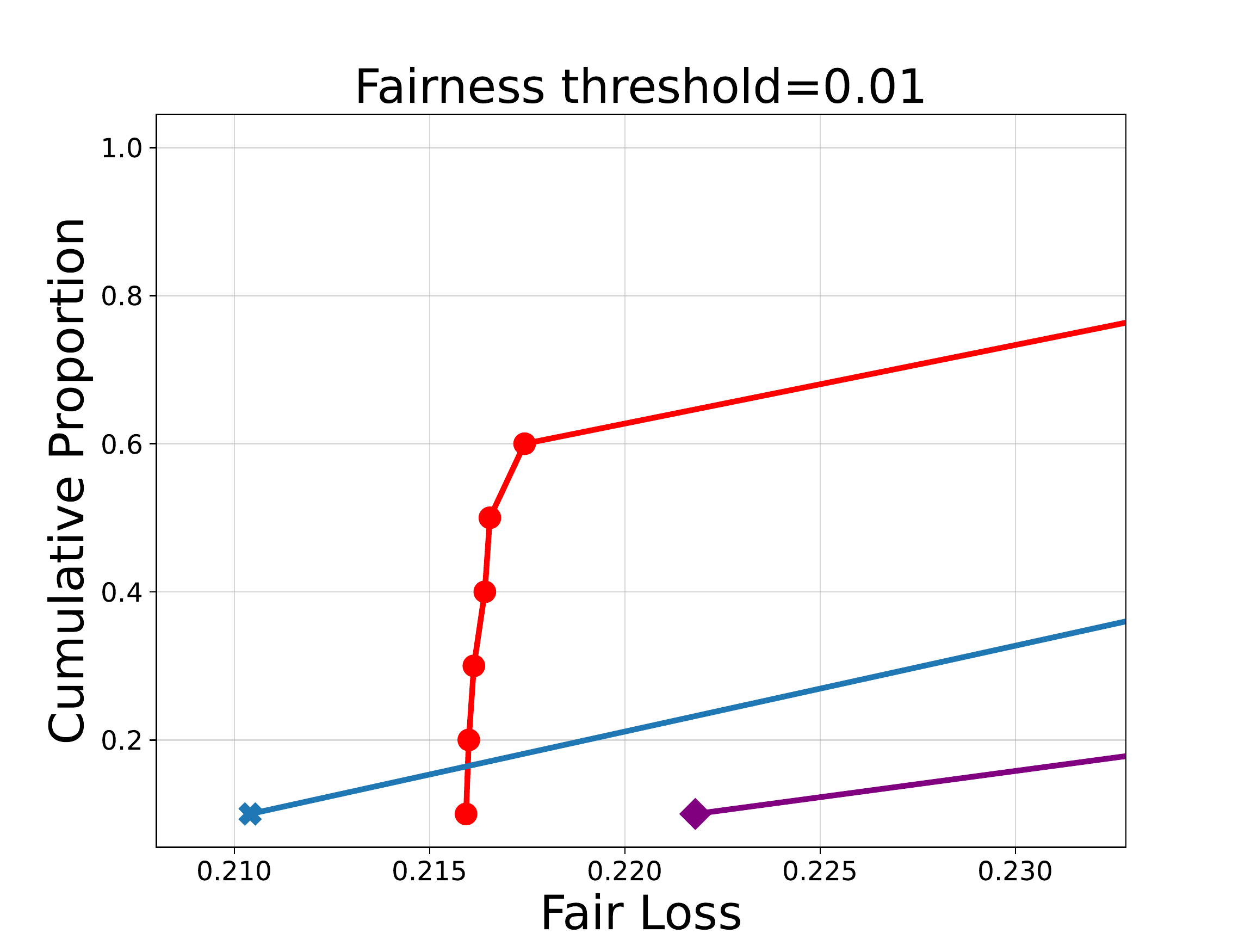}
  }
  \\
  \subfigure[Fair loss on \emph{Bank}]{
  \includegraphics[width=0.25\textwidth]{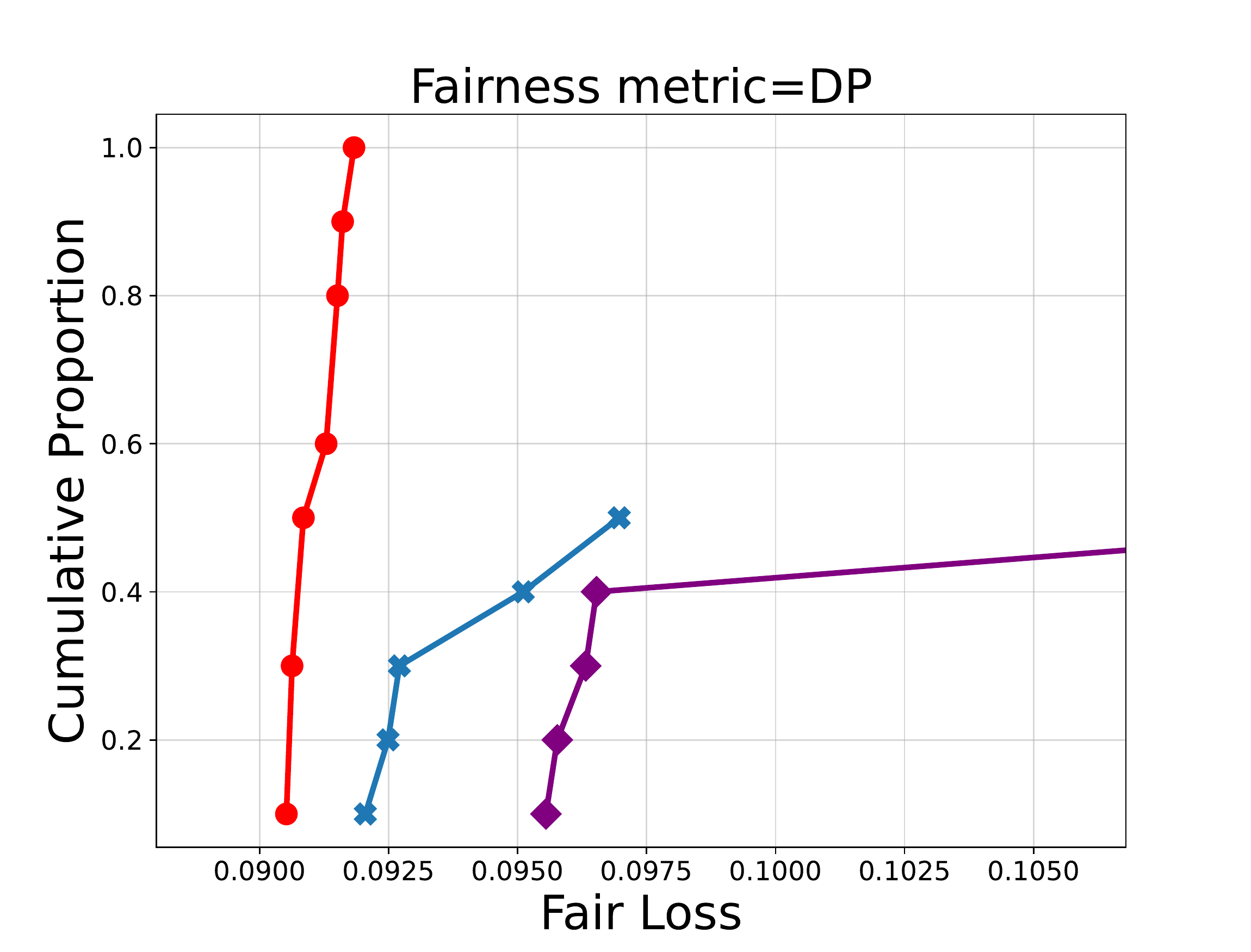}%
  \includegraphics[width=0.25\textwidth]{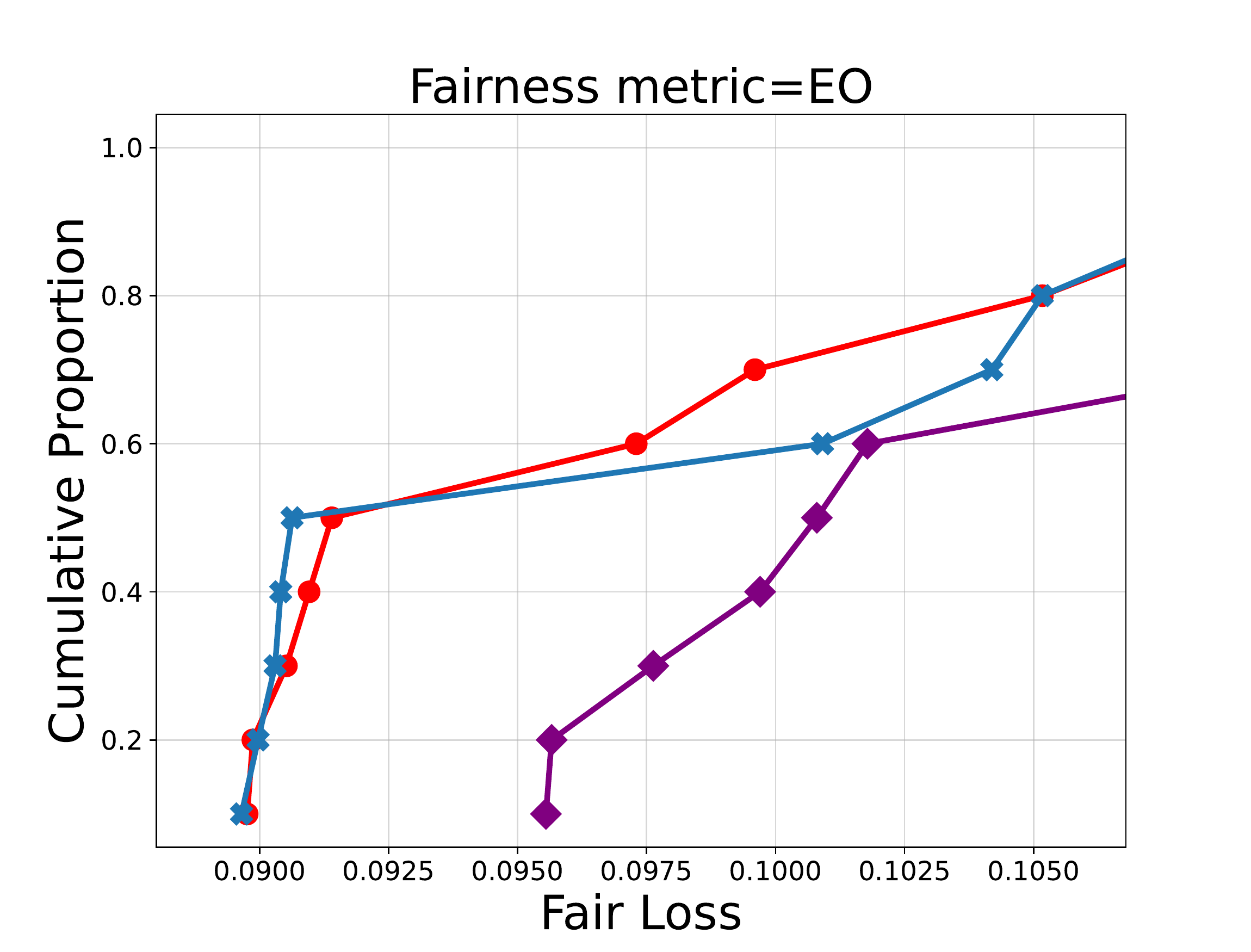}%
  \includegraphics[width=0.25\textwidth]{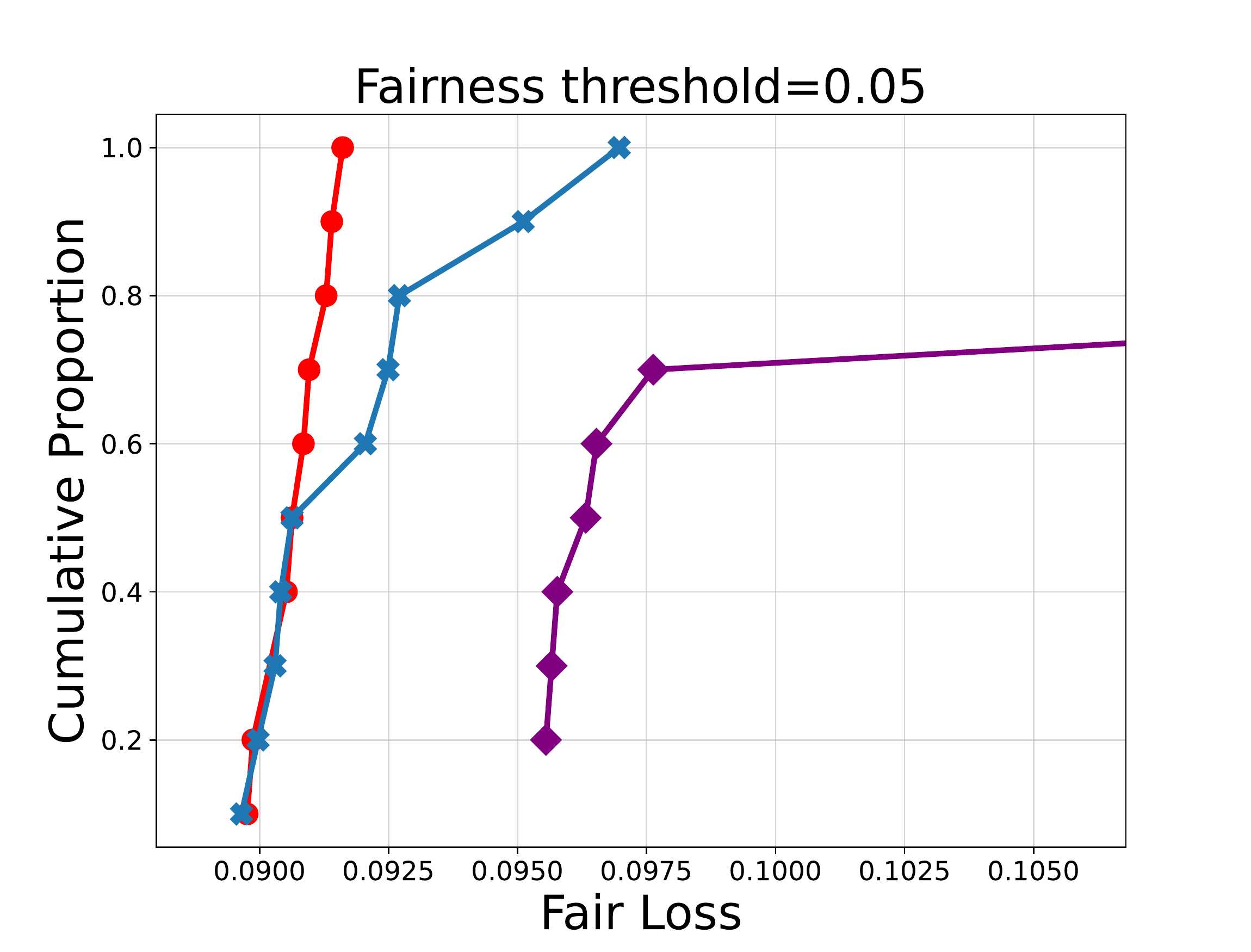}%
  \includegraphics[width=0.25\textwidth]{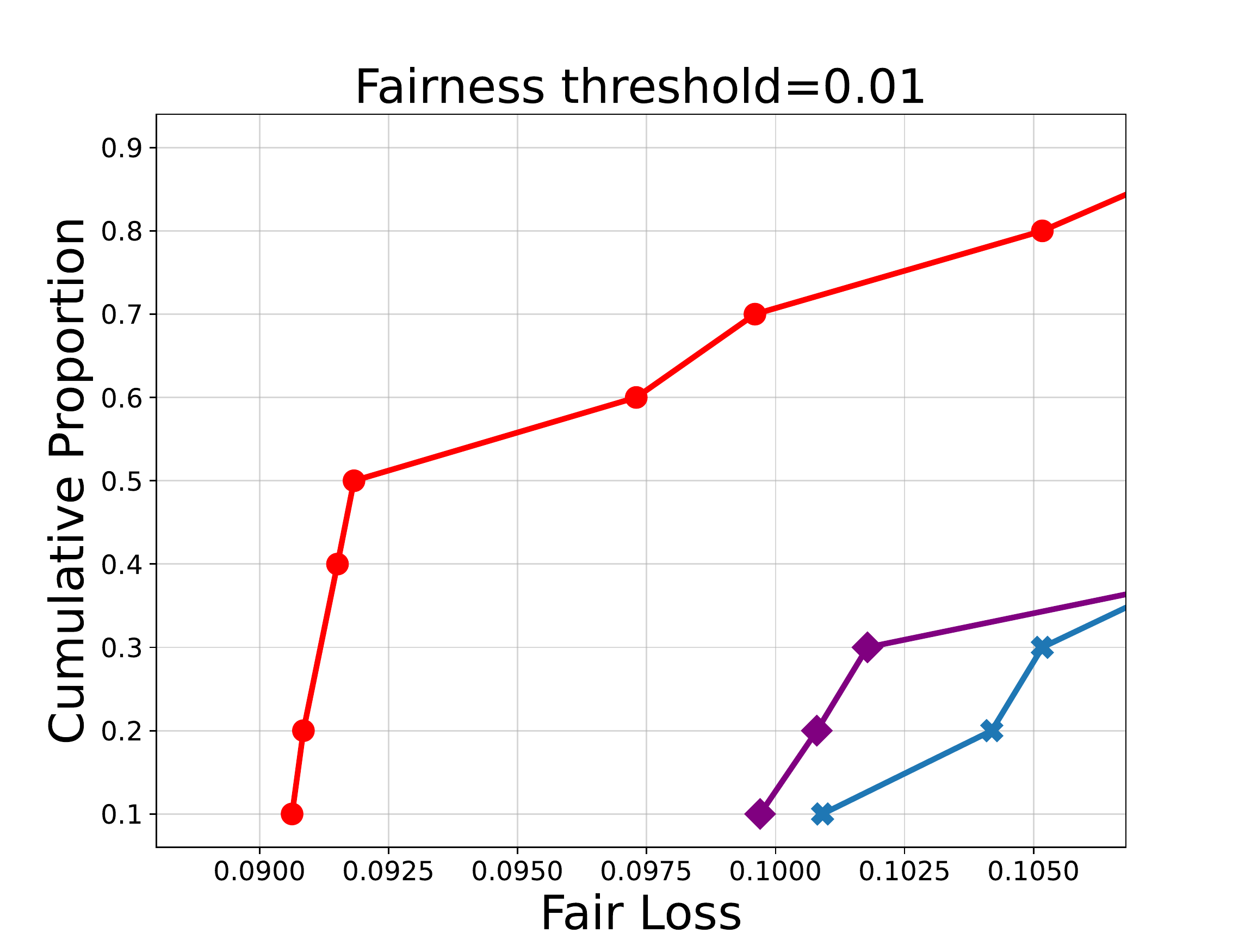}
  }
  \\
  \subfigure[Fair loss on \emph{Compas}]{
  \includegraphics[width=0.25\textwidth]{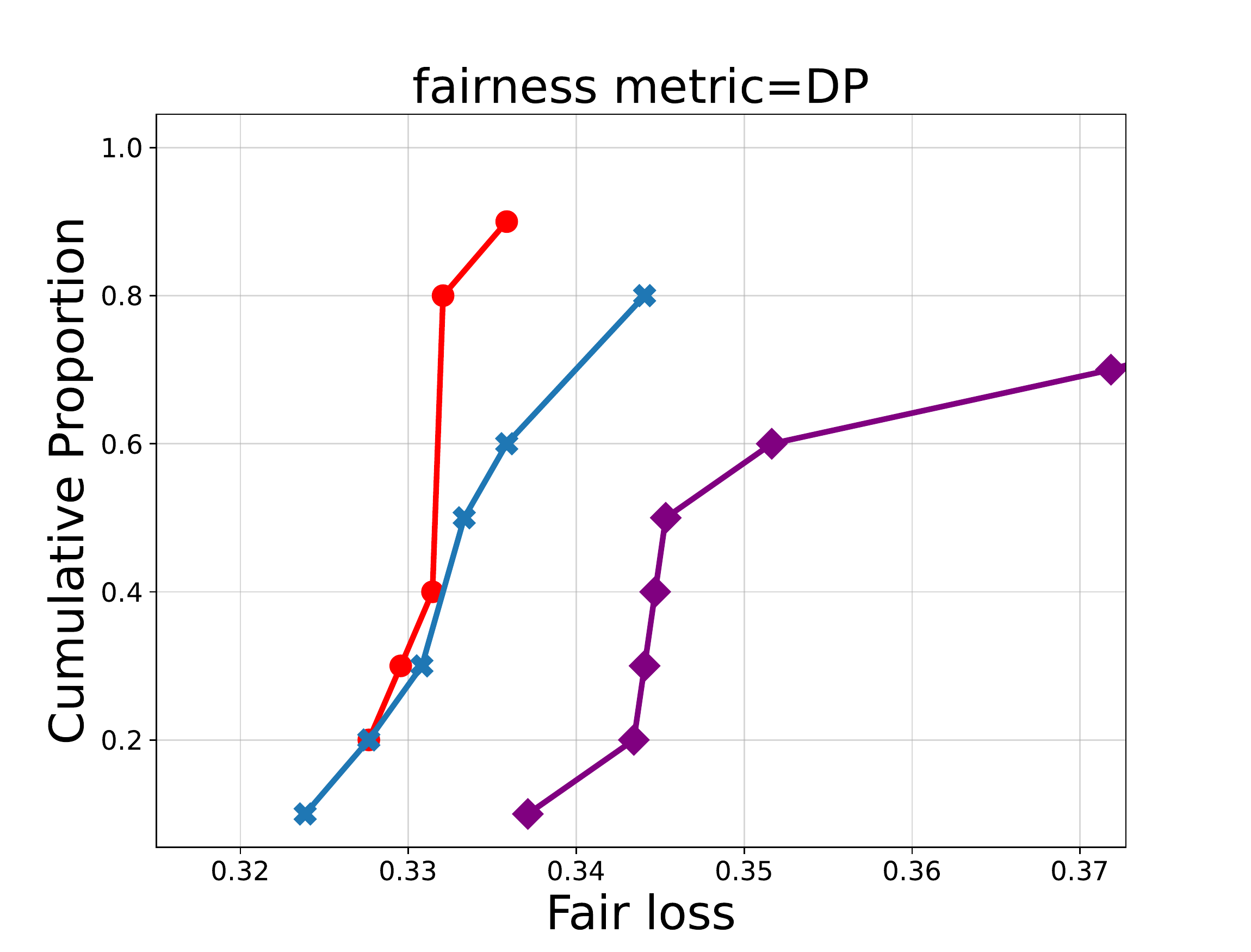}%
  \includegraphics[width=0.25\textwidth]{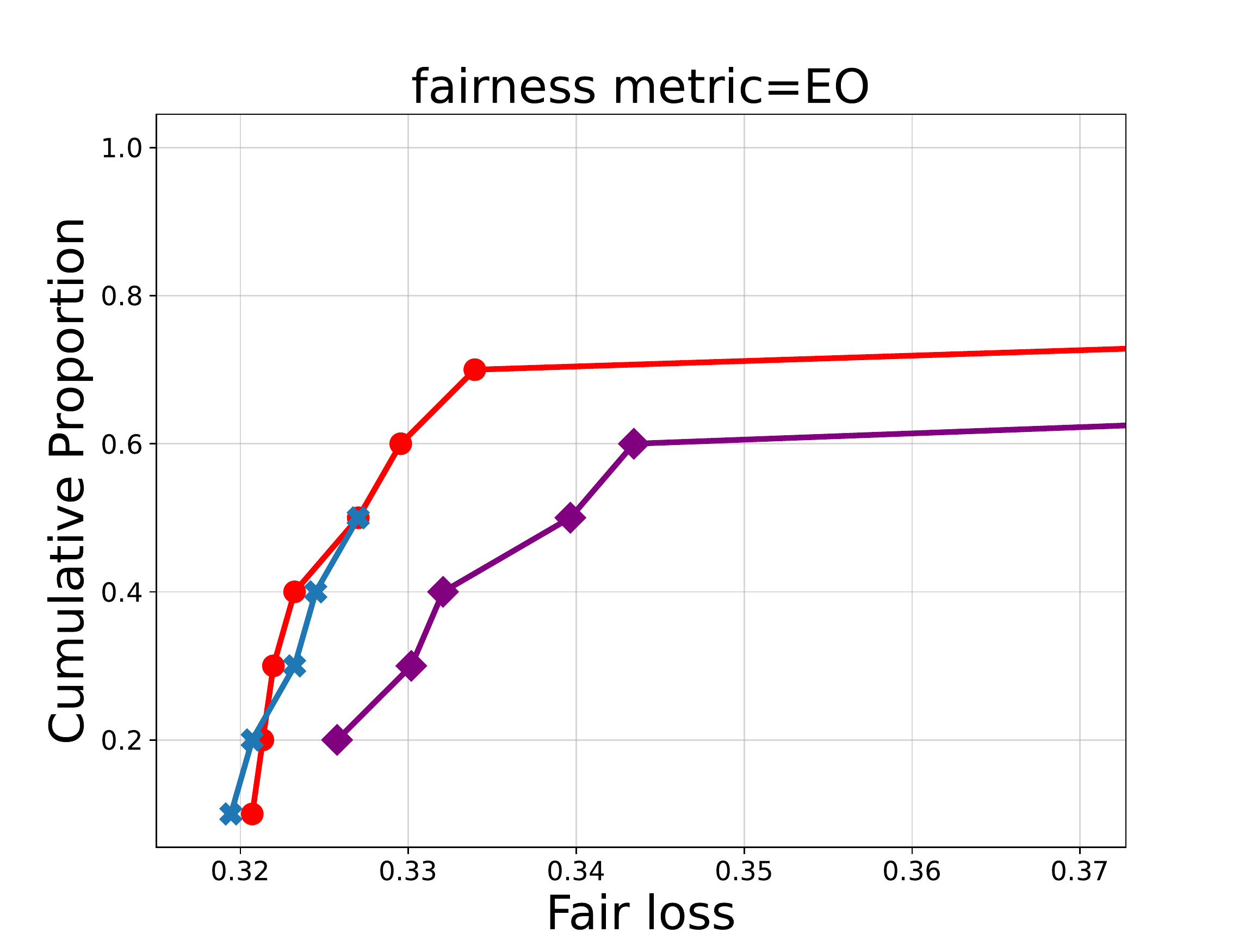}%
  \includegraphics[width=0.25\textwidth]{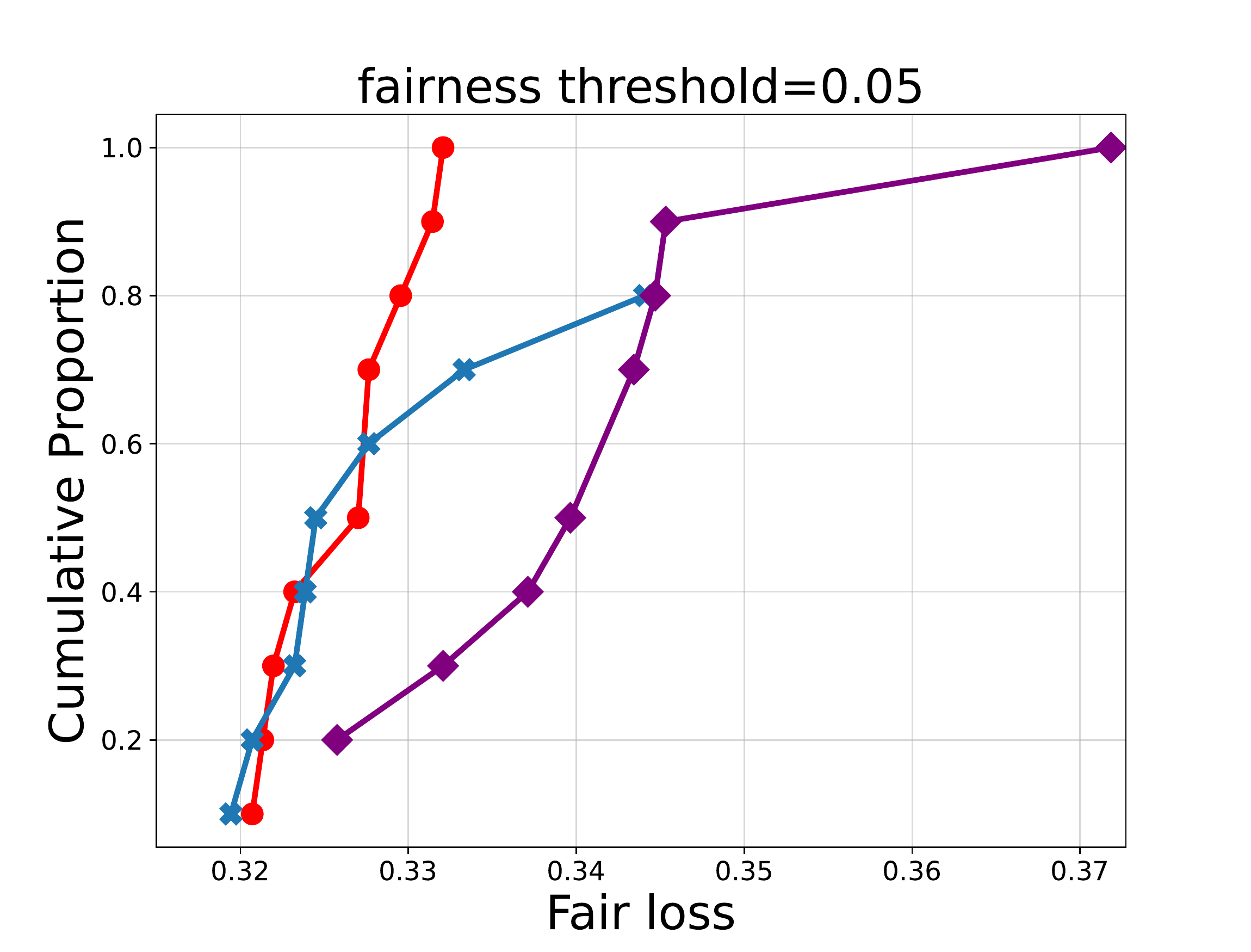}%
  \includegraphics[width=0.25\textwidth]{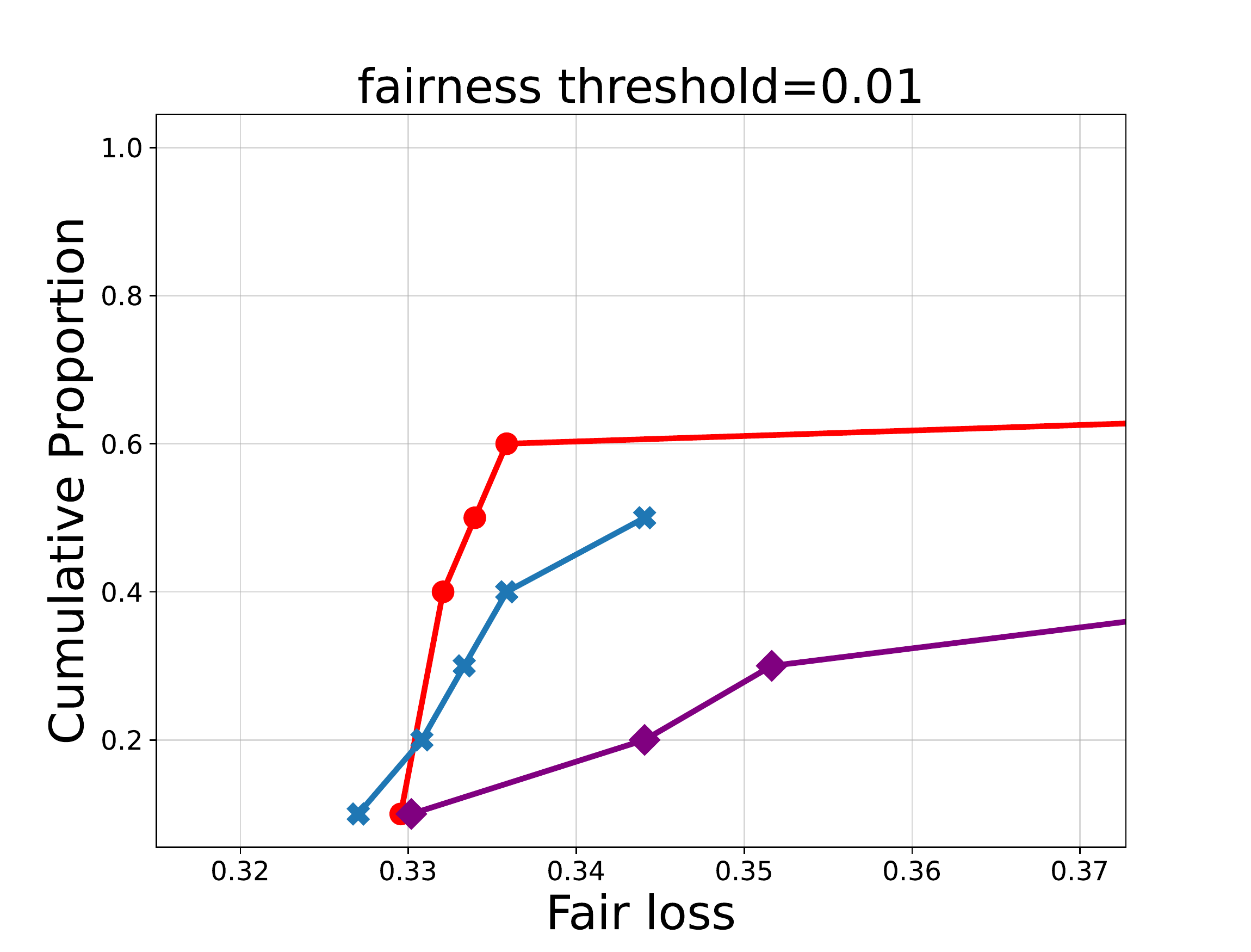}
  }
  \\
  \subfigure[Fair loss on \emph{MEPS}]{
  \includegraphics[width=0.25\textwidth]{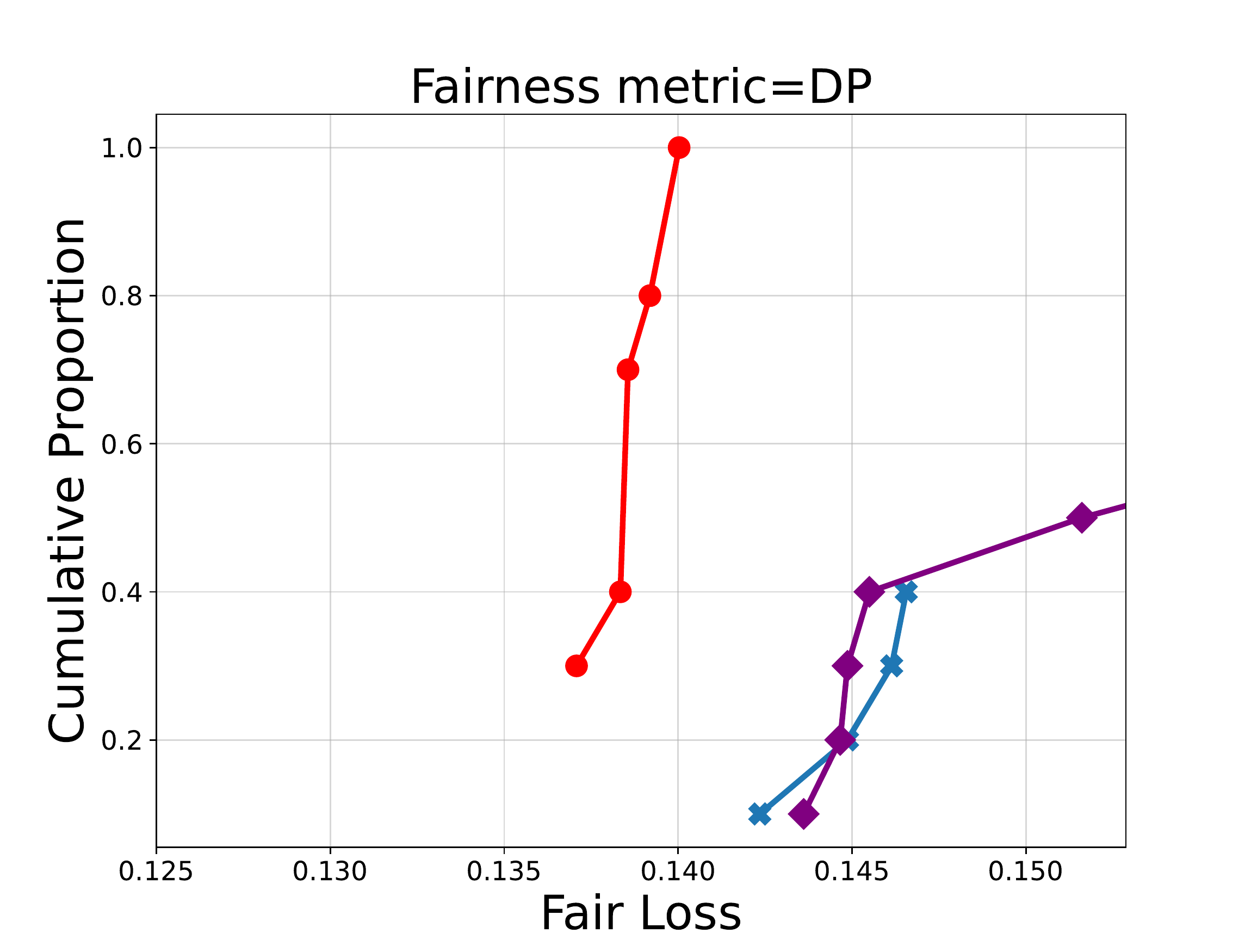}%
  \includegraphics[width=0.25\textwidth]{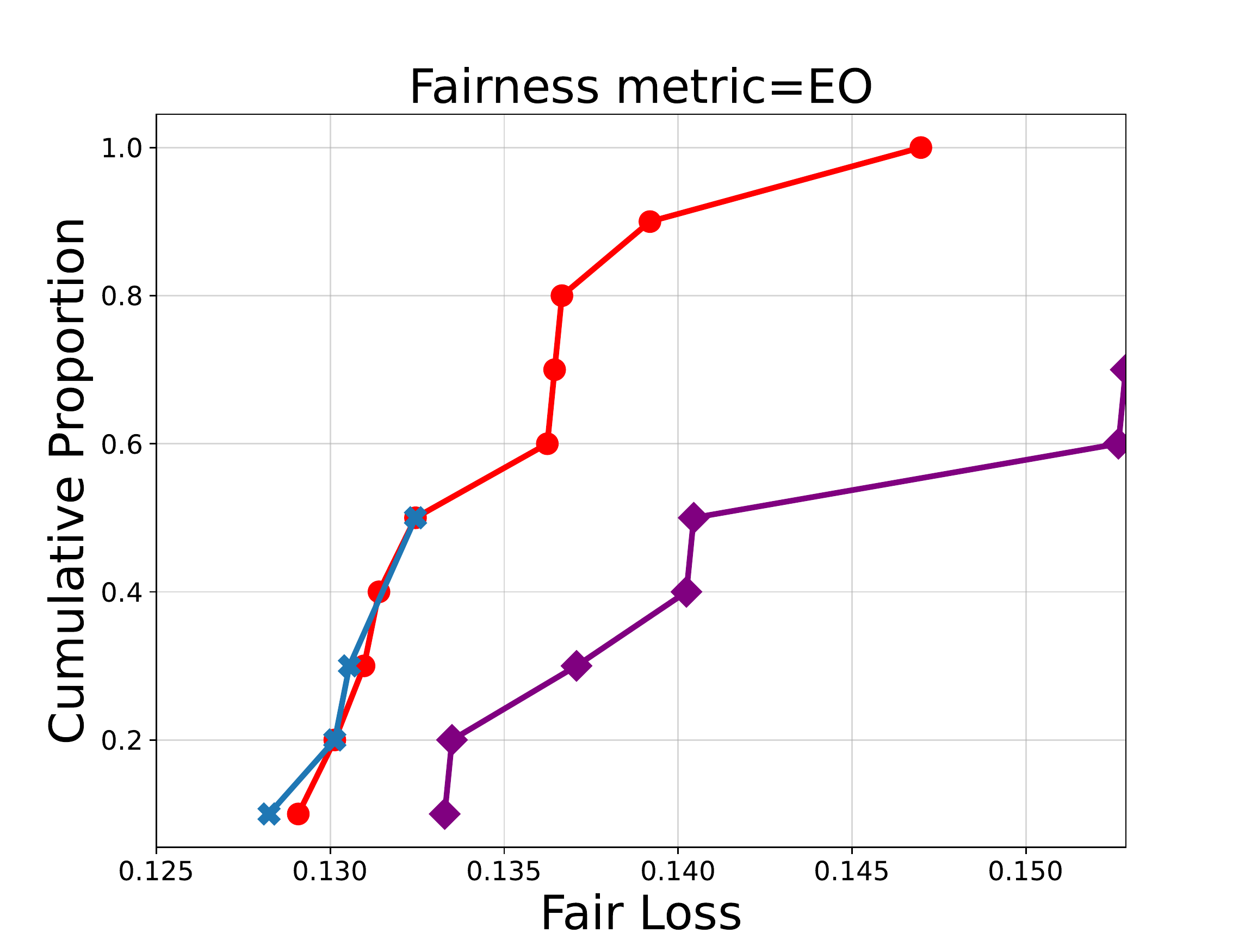}%
  \includegraphics[width=0.25\textwidth]{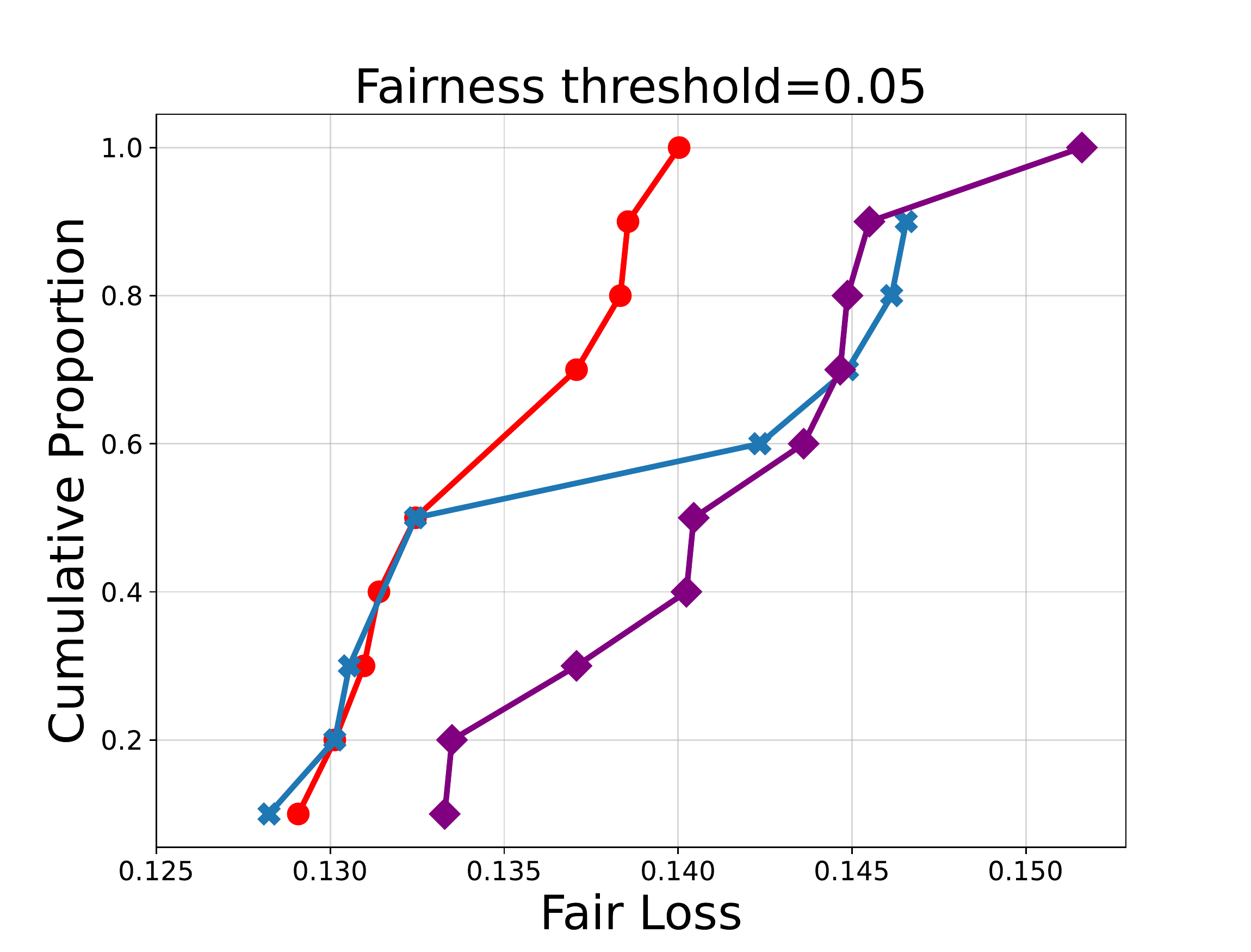}%
  \includegraphics[width=0.25\textwidth]{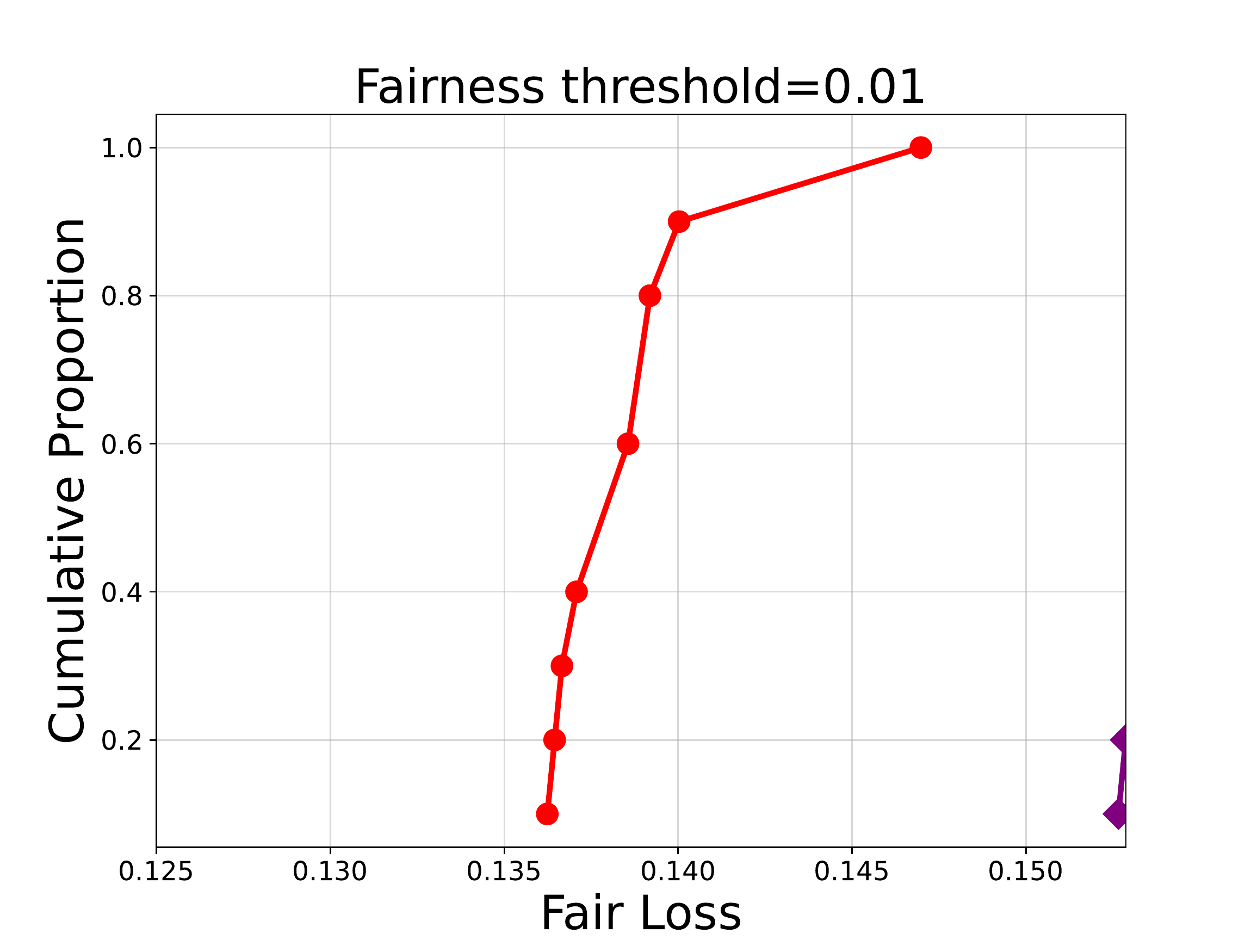}
  }
  \\
  \caption{Fair loss with threshold-based post-processing as the  unfairness mitigation method when tuning XGBoost.}
   \label{fig:exp_res_boxplot_post}
  \end{figure*}

We include the results obtained with two additional unfairness mitigation methods, including Grid Search reduction and threshold-based post-processing, in Figure~\ref{fig:exp_res_boxplot_grid} and Figure~\ref{fig:exp_res_boxplot_post}, where we use \fairFLAMLg and \fairFLAMLp to denote Grid Search reduction and the post-processing method respectively. 
From these results, we observe consistently good performance from \fairFLAML similar to the case with Exponentiated Gradient as reported in Figure~\ref{fig:exp_res_cdf} in the main paper. By comparing \fairFLAML with different unfairness mitigation methods, we find Exponentiated Gradient works the best overall and thus used it as the default unfairness mitigation method in \fairFLAML if not otherwise specified.

\subsection{Extensibility}\label{appendix:extension}
Our proposed framework and system are highly extensible and are compatible with a wide range of hyperparameter searchers, fairness definitions, and unfairness mitigation methods. We include code examples to verify the wide compatibility in our codebase (instructions are provided in the `README.md' file of the submitted code base).
\paragraph{Compatibility with pre-processing mitigation methods.} In addition to the two in-processing mitigation methods, and one post-processing mitigation method evaluated, our system is also compatible with pre-processing mitigation methods. We confirmed the compatibility with one pre-processing method~\cite{feldman2015certifying}.

\paragraph{Compatibility with other hyperparameter searchers.} Our framework is compatible with any hyperparameter searcher that satisfies the abstraction developed in Section~\ref{sec:framework}. We confirmed our framework's and system's compatibility with another two state-of-the-art searchers: Optuna~\cite{akiba2019optuna}, and BlendSearch~\cite{wang2021economical}. The former is a Bayesian optimization hyperparameter searcher, and the latter is a hyperparameter searcher which combines global search and local search. 

\paragraph{Compatibility with fairness definitions and mitigation methods for regression tasks.} All the empirical evaluations reported in this work are from binary classification tasks. For regression tasks, different fairness definitions and mitigation methods are needed. We confirmed FairFLAML's compatibility with one quantitative fairness definition, bounded group loss~\cite{agarwal2019fair}, and a reduction-based unfairness mitigation method~\cite{agarwal2019fair} for regression tasks.

\section{Social Impact and Limitations}

\paragraph{Social Impact.} 
In many real-world applications where the decisions to be made have a direct impact on the well-being of human beings, it is not sufficient to only have high prediction accuracy. We also expect the ML-based decisions to be ethical that do not put certain unprivileged groups or individuals at systematic disadvantages. 
One example of such human-centered applications where machine learning is heavily used is modern financial services, including financial organizations' activities such as credit scoring, lending and etc. 
According to a survey of UK regulators in 2019 \cite{2019uk_financial_sruvey},  two-thirds of UK financial industry participants rely on AI today to make decisions. Moreover, an Economist Intelligence Unit research report \cite{road_ahead_ai_and_financial_services} found that 86\% of financial services executives plan on increasing their AI-related investments through 2025. There has been increasing evidence showing various unfairness issues associated with machine-made decisions or predictions~\cite{o2016weapons,barocas2016big,2016machinebias}. 
Laws and regulations, for example the U.S. Fair Credit Reporting Act (FCRA) and Equal Credit Opportunity Act (ECOA), have been enforced to prohibit unfairness and discrimination in related financial activities. 

At the same time, AutoML is playing an increasingly impactful role in modern machine learning lifecycles. According to a recent report~\cite{AutoML_Market_report} from ReportLinker, the AutoML market is predicted to reach $\$14,830.8$ million by 2030, demonstrating a compound annual growth rate of 45.6$\%$ from 2020 to 2030. Despite the growing importance of AutoML in modern ML lifecycles, the mitigation of unfairness in AutoML is under-explored. Our work makes the first attempt to introduce unfairness mitigation into the AutoML pipeline. The flexible framework allows integration of most existing unfairness mitigation techniques. We believe this can greatly facilitate consideration of fairness issues in AutoML practices.  

\paragraph{Limitations and future work.}  There are currently three major limitations of our work and we plan to address some of them in our future work. 
\begin{enumerate}[leftmargin=*]
    \vspace{-2mm}
    \setlength\itemsep{-0.2em}
    \item The proposed fair AutoML system FairFLAML is advantageous to an alternative (e.g., MAlways) when the computation overhead of the unfairness mitigation procedure is high. When this overhead is marginal compared to the model training time, the carefully designed resource allocation strategy in FairFLAML won't make much difference (mainly because the room for resource-saving is small). We emphasize that this limitation does not fundamentally undermine the contribution of this work: (a) in such an `easy' case, our system is no worse than MAlways; (b) the `hard' cases investigated in this paper are non-negligible because of the in-processing mitigation methods' good empirical performance and theoretical guarantee~\cite{agarwal2018reductions}. 
    \item In this work, we are not providing theoretical guarantees on the benefit of FairFLAML in terms of the resource-saving for reaching a particular target fair loss or the best fair loss can be achieved given any resource budget. Such theoretical guarantees can potentially be developed using competitive analysis. We plan to investigate the theoretical properties of our proposed FairFLAML system in future work. 
    \item Although the proposed framework and system are presumably compatible with any unfairness mitigation satisfying the developed abstractions, we only evaluated a few state-of-the-art ones in this work. We plan to keep track of the cutting-edge research development in fairness definition and unfairness mitigation and potentially explore more variants of them. We also plan to further expand the framework such that multiple different types of unfairness mitigation methods can be automatically selected and used in combination when necessary.
\end{enumerate}

\end{document}